%% file: acl_latex.tex
\pdfoutput=1

\documentclass[11pt]{article}

\usepackage[preprint]{acl}

\usepackage{times}
\usepackage{latexsym}

\usepackage[T1]{fontenc}

\usepackage[utf8]{inputenc}

\usepackage{microtype}

\usepackage{inconsolata}

\usepackage{graphicx}

\input{libs}

\input{macros}

\title{\dataset{}: Generating News Events for Out-of-training Question Answering}
\title{\dataset{}: Evidence-based Question Answering with Generated News Events}

\author{
 Max Glockner$^1$\Thanks{Work was done while MG was an intern at Amazon AGI.},
 Xiang Jiang$^2$,
 Leonardo F. R. Ribeiro$^2$, 
 Iryna Gurevych$^1$,
 Markus Dreyer$^2$
 \vspace{0.25em}\\
  $^1$UKP Lab, TU Darmstadt and Hessian Center for AI (hessian.AI)
  $^2$Amazon AGI 
   \vspace{0.25em}\\
    \small \texttt{\url{www.ukp.tu-darmstadt.de}} \\
    \small \texttt{\{maxg216\}@gmail.com} \\
    \small\texttt{\{jxiang,leonribe,mddreyer\}@amazon.com}
}

\begin{document}
\maketitle
\begin{abstract}
Evaluating Retrieval-Augmented Generation (RAG) in large language models (LLMs) is challenging because benchmarks can quickly become stale. Questions initially requiring retrieval may become answerable from pretraining knowledge as newer models incorporate more recent information during pretraining, making it difficult to distinguish evidence-based reasoning from recall.
We introduce \dataset{} (\underline{N}ews \underline{E}vents for \underline{O}ut-of-training \underline{Q}uestion \underline{A}nswering), a benchmark designed to address this issue. To construct \dataset{}, we generated timelines and knowledge bases of fictional news events and entities along with news articles and Q\&A pairs to prevent LLMs from leveraging pretraining knowledge, ensuring that no prior evidence exists in their training data.
We propose our dataset as a new platform for evaluating evidence-based question answering, as it requires LLMs to generate responses exclusively from retrieved evidence and only when sufficient evidence is available.
\dataset{} enables controlled evaluation across various evidence scenarios, including cases with missing or misleading details.
Our findings indicate that LLMs struggle to distinguish subtle mismatches between questions and evidence, and suffer from short-cut reasoning when key information required to answer a question is missing from the evidence, underscoring key limitations in evidence-based reasoning.\footnote{\url{https://github.com/amazon-science/neoqa}}

\end{abstract}

\input{latex/content/00__intro}

\input{latex/content/02__parametric_vs_evidence}

\input{latex/content/03__our_dataset_construction}

\input{latex/content/04__our_dataset_stats_and_eval}
\input{latex/content/05__experiments}

\input{latex/content/06__analysis}

\input{latex/content/01__related_work}

\input{latex/content/07__conclusion}

\section*{Limitations}
GPT-4-generated event timelines may reflect the LLM's social biases \citep{shin-etal-2024-ask} and prompt-induced biases, making them unrepresentative of all real-world events. The challenges across question types depend on the LLM and prompts used. While suitable for zero-shot experiments, \dataset{} is not appropriate for fine-tuning, as models could overfit to the generated question characteristics. LLMs may also introduce numerous, often intractable inconsistencies within timelines \citep{yang-etal-2022-re3}. Our experiments do not consider such possible inconsistencies. Our experiments are limited to the reported Phi3 and Qwen2.5
models due to licensing and legal restrictions, institutional policies, and the requirements of long-context LLMs. Due to these restrictions, we have not experimented with larger models and our findings are restricted to the evaluated models and sizes.
Our work penalizes shortcut reasoning, which undermines model trustworthiness. However, some view it as beneficial for efficiency by skipping reasoning steps in CoT prompting \citep{ding2024break}. \dataset{} is limited to English.
Measuring the ``naturalness'' of generated text in a human study is challenging because the news articles produced by GPT-4o feature fictional entities, making it impossible to objectively compare them with real news. While we instructed GPT-4o to generate news articles in natural language and various styles, our manual evaluation focuses on the validity of the generated content rather than its naturalness.

\section*{Ethics Statement}
All timelines and named entities are entirely fictional as approximated via Wikipedia, yet may include real-world entities if the LLM failed to detect named entities as such. The generated dataset may exhibit social biases, influenced by underlying social biases of LLMs. While our work is focused on creating fictional timelines, some events may unintentionally resemble real-world occurrences or entities. We emphasize that this data is fictional, and any similarities to real events or entities are purely coincidental and should be interpreted as such.
Our paper passed an extensive multi-phase in-house review that took legal and ethical considerations into account.
The human annotations in this paper are provided by qualified Mechanical Turk workers.
We provided fair pay to our annotators.
For the multi-hop answer evaluation, the workers take on average five minutes to complete one evaluation.
We pay the workers \$0.35 per question and \$1.35 bonus, which leads to a pay of \$20.4 per hour.

\section*{Acknowledgments}
The authors wish to thank Mengwen Liu for her insightful discussions and Andrei Coman, Adrià de Gispert, Weiwei Cheng and Saar Kuzi for their valuable feedback on an earlier version of this work. Max Glockner has been supported by the German Federal Ministry of Education and Research and the Hessian Ministry of Higher Education, Research, Science and the Arts within their joint support of the National Research Center for Applied Cybersecurity ATHENE.
\newpage

\bibliography{custom}

\newpage
\appendix
\label{sec:appendix}
\input{latex/appendix/realtimeqa-experiments}

\input{latex/appendix/timeline-example}

\input{latex/appendix/story-generation-pipeline}

\input{latex/appendix/experiments}
\input{latex/appendix/prompts}

\end{document}

%% file: libs.tex
\usepackage{lipsum}  
\usepackage{booktabs}
\usepackage{adjustbox}
\usepackage{url}
\usepackage{amssymb}
\usepackage{amsmath}
\usepackage{multirow}
\usepackage{tabularx}
\usepackage{graphicx}
\usepackage{enumitem}
\usepackage{multirow}
\usepackage{mathtools}
\usepackage{siunitx}

\usepackage{times}
\usepackage{latexsym}
\usepackage{xcolor} %
\usepackage[most]{tcolorbox}
\usepackage[colorinlistoftodos]{todonotes}

\usepackage{subcaption} %

%% file: macros.tex
\newcommand{\dataset}[0]{\textsc{NeoQA}}

%% file: latex/content/00__intro.tex
\section{Introduction}
\begin{figure}[t]
    \centering
    \includegraphics[width=\columnwidth]{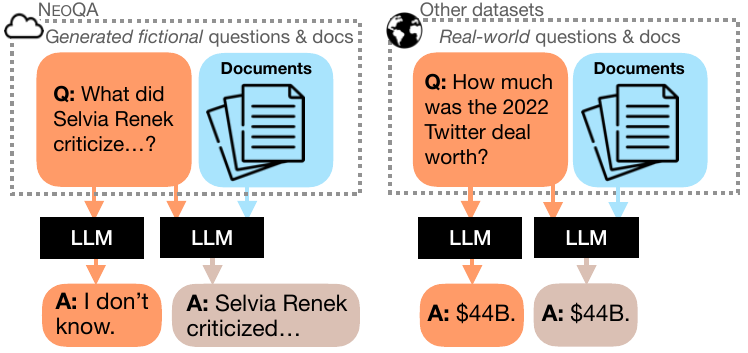}
    \caption{\textbf{Left:} \dataset{} features LLM-generated questions and documents about events from a fictional timeline, ensuring that LLMs can only answer by reasoning over the documents. \textbf{Right:} Real-world RAG datasets become ineffective for newer LLMs that have internalized knowledge of recent events, rendering the provided evidence documents redundant.}
    \label{fig:page1_fig}
\end{figure}
\input{figures/figure-timeline}

Retrieval-Augmented Generation (RAG) equips LLMs with external information to complement their parametric knowledge \citep{lewis2020retrieval, yu2024evaluation} and enables them to answer questions that involve information beyond their pretraining data, such as recent events or rare entities. For trustworthy applications, the ability to reason over multiple evidence documents is critical to producing verifiable answers grounded in these documents~\citep{liu-etal-2023-evaluating,yue-etal-2023-automatic,li-etal-2024-attributionbench}. 
 LLMs must not only be able to answer a question correctly when the evidence is sufficient, but also be able to deflect from answering if the question cannot be answered given the evidence \cite{cao-2024-learn}. 
However, benchmarks for evidence-based reasoning on real-world data lose value over time as LLMs increasingly can rely on updated parametric knowledge from pretraining rather than external information (Figure~\ref{fig:page1_fig}, \emph{right}), as analyzed empirically in Section~\ref{sec:impact-of-parametric-knowledge}.
Constructing datasets with recent data \citep{chen2024benchmarking, tang2024multihop, karpinska-etal-2024-one} only postpones the issue until LLMs are retrained, while frequent dataset updates \citep{vu2023freshllms, kasai2024realtime}
mitigate it but 
make consistent progress tracking difficult.

To address these challenges, we introduce \dataset{}, a fully LLM-generated dataset of fictional events with associated news articles, as well as question-answer pairs. Organized into \emph{timelines}, each with ten sequential \emph{events} (Figure~\ref{fig:timeline}), \dataset{} mimics how events unfold over time in reality. All events and named entities are fictional to avoid interference from LLMs with updated parametric knowledge (Figure~\ref{fig:page1_fig}, \emph{left}).
Each event includes resolved named entities, and a corresponding knowledge base (KB) entry is created and continuously updated as the events progress.
The events adhere to real-world physical laws and common sense, allowing models to leverage their commonsense reasoning \citep{choi2022curious}.
For each event, news articles and multiple-choice questions are independently generated and grounded in identifiable atomic information. This allows to pair a question with any set of news articles and clearly distinguish between sufficient or insufficient evidence, and unrelated documents. \emph{News articles} serve as evidence and focus on different information of a single event, \emph{questions} require reasoning over information from up to two previous events. For example, answering the question in Figure~\ref{fig:timeline} requires combining the red and purple facts. Any set of news articles with both facts provides sufficient evidence, while any with fewer is insufficient. 
This allows us to test models under different evidence conditions, requiring them to answer correctly when sufficient evidence is available and to deflect when it is not.
Overall, by grounding \dataset{}'s news articles---used as inference-time evidence---and questions in fictional timelines that are independent from real-world news cycles, 
\dataset{} 
remains a reliable benchmark for evaluating future LLMs, free from the risk of pretraining data contamination or knowledge conflicts about real-world events.
Following the recommendations of \citet{jacovi-etal-2023-stop}, we release \dataset{}  under a no-derivatives license (CC-BY-ND-4.0) and with public key encryption.
\input{figures/figure-realtimeqa-results-aggregated}

Because parametric knowledge cannot replace external evidence in \dataset{}, a question can only be answered correctly when sufficient external evidence is provided. When evidence is insufficient, the model can only \emph{guess} the correct answer using shortcut reasoning \citep{jiang-bansal-2019-avoiding, chen-durrett-2019-understanding, trivedi-etal-2022-musique}, undermining their trustworthiness. 
Determining whether a model is using shortcut reasoning to guess an answer or genuinely completing it with its learned (parametric) knowledge is difficult in real-world datasets, where models can often justifiably fill in knowledge gaps. However, this is not the case in \dataset{}, where answers require sufficient external evidence. This setup makes it possible to identify and penalize shortcut reasoning during evaluation. %
For example, in HotpotQA \citep{yang2018hotpotqa}, answering “Shenley Hall is a house in a parish how far from central London?” requires identifying the house’s village and its distance from London. If only the latter is provided, it is unclear whether the model reasons correctly or uses shortcuts.
\dataset{} enables controlled experiments on evidence-based reasoning that account for shortcut reasoning, requiring models to compare questions with evidence and answer only when justified, deflecting otherwise. It includes answerable and unanswerable questions (with unverifiable or incorrect assumptions \citep{kim-etal-2021-linguist, hu-etal-2023-wont}), combined with evidence that is sufficient, insufficient, and/or distracting. Our experiments show that models struggle to distinguish sufficient from insufficient evidence, frequently relying on shortcuts and failing to detect subtle mismatches.
In summary, our contributions are:
\begin{enumerate}[noitemsep]
    \item A \textbf{novel methodology} for automatically generating an evidence-based question-answering dataset grounded in fictional timelines.
    \item The \textbf{\dataset{} dataset} with diverse question types and evidence configurations for evaluating evidence-based reasoning.
    \item \textbf{Controlled experiments} reveal the challenges posed by shortcut reasoning.
\end{enumerate}

%% file: figures/figure-timeline.tex
\definecolor{timelineev1}{RGB}{255,0,0} 
\definecolor{timelineev2}{RGB}{153,0,153} 
\definecolor{timelineans}{RGB}{0,153,0} 
\definecolor{timelineevent}{RGB}{0,29,188}

\begin{figure*}[h!]
\centering
  \includegraphics[width=\textwidth]{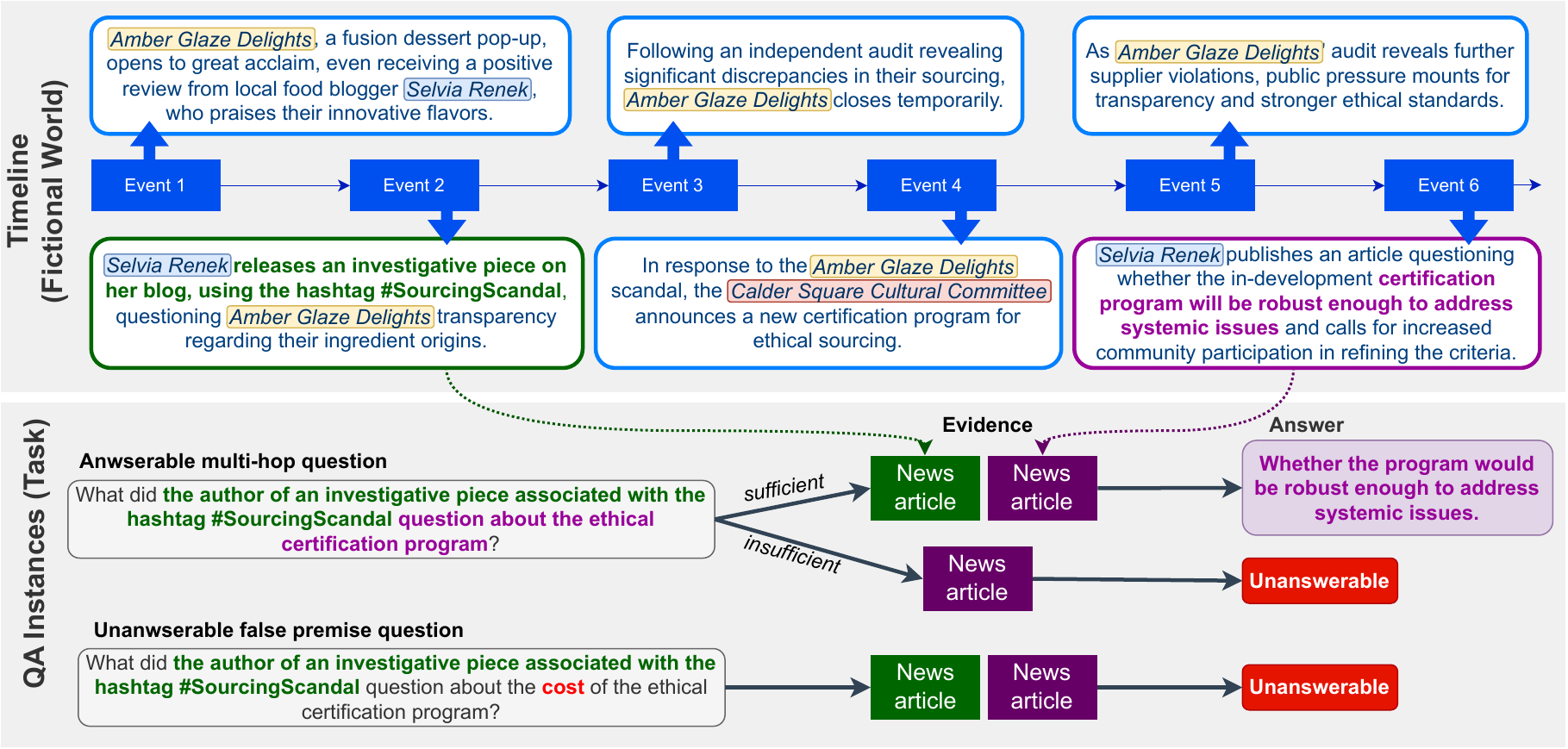}
  \caption{An extract of a timeline from \dataset{} with six out of ten events (summarized for visualization) with highlighted fictional named entities. Answering a multi-hop question requires combining information from two events. The model should deflect when only partial (\emph{insufficient}) information is available or when subtle permutations make the question unanswerable (e.g., false premise questions).}
  \label{fig:timeline}
\end{figure*}

%% file: figures/figure-realtimeqa-results-aggregated.tex
\definecolor{cutoffblue}{HTML}{296399}
\definecolor{cutoffgreen}{HTML}{22723F}
\definecolor{cutoffbright}{HTML}{888888}
\definecolor{cutoffdark}{HTML}{111111}

\begin{figure}[t]
\centering
  \includegraphics[width=\linewidth]{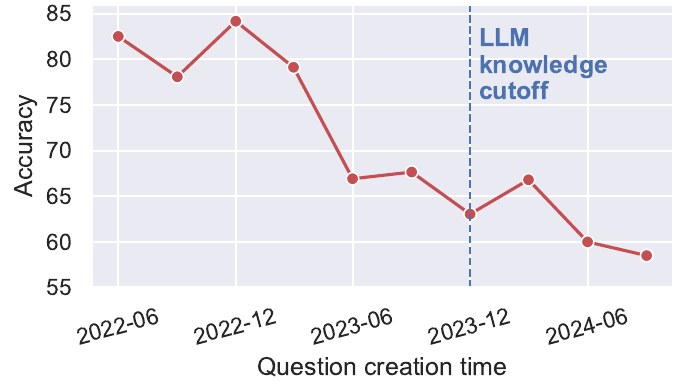}
  \caption{GPT-4 Turbo accuracy on RealTimeQA questions (no RAG evidence provided). It answers older questions more accurately from memory, suggesting that older RAG datasets can be solved  without RAG.}
  \label{fig:realtimeqa-results-aggregated}
\end{figure}

%% file: latex/content/02__parametric_vs_evidence.tex
\section{Parametric Knowledge Interference}
\label{sec:impact-of-parametric-knowledge}
Data contamination in LLM pretraining, where test data overlaps with training data, has compromised several benchmarks \citep{magar2022data,jacovi-etal-2023-stop,elazar2023s, sainz-etal-2024-data}. We test whether RAG benchmarks are similarly affected by events overlapping with the pretraining data. If LLMs acquire relevant knowledge during pretraining, such benchmarks lose their purpose. 
To quantify this, we evaluate GPT-4 Turbo (reported knowledge cutoff: December 2023\footnote{\url{https://platform.openai.com/docs/models}}) on RealTimeQA~\citep{kasai2024realtime}, a dataset of weekly multiple-choice news quizzes (June 2022–Jan 2024). We extend the dataset through September 2024 (see Appendix~\ref{appendix:realtimeqa-experiments}) and test the model's accuracy using only its parametric knowledge, without external evidence.
Figure~\ref{fig:realtimeqa-results-aggregated} shows higher accuracy on older questions, indicating that the LLM acquired much of the relevant information during pretraining. Performance drops sharply around March 2023, several months before the reported knowledge cutoff. We hypothesize that this discrepancy arises because reported knowledge cutoffs are conservative estimates, while the effective cutoff for different sources may be earlier \citep{cheng2024dated}.
Performance on unseen news remains above chance (25\% for selecting from four options), likely due to common-sense reasoning, which helps eliminate distractors.

%% file: latex/content/03__our_dataset_construction.tex
\section{Task Definition}
We introduce \dataset{}, a QA dataset agnostic to parametric knowledge by focusing on fictional events and named entities that do not exist in the real world.
The task is formulated as multiple-choice, where the model receives a question, a preselected set of news articles as evidence, and a set of seven candidate answers (a correct answer, a deflection option if unanswerable, and five distractor answers). The model must assess whether the evidence is sufficient to answer the question, select the correct answer if possible, and deflect if the evidence is insufficient, or if the question’s assumptions are unverifiable or incorrect. We always include an explicit ``unanswerable'' option, which has been shown to help models deflect when the answer is unknown \citep{slobodkin-etal-2023-curious}.
We do not include document retrieval in our task formulation, but instead control the preselected evidence to simulate realistic retrieval conditions with sufficient, insufficient, or irrelevant information. This allows for a fine-grained evaluation of how models reason over imperfect evidence, as real-world retrieval often includes noise or missing details.

The multiple-choice approach addresses two key challenges in LLM-generated answers: (1) avoiding the need for an expensive judge LLM, and (2) reducing false negatives caused by ambiguous questions, which can lead to multiple different but valid interpretations based on different evidence \citep{min-etal-2020-ambigqa, glockner-etal-2024-ambifc}. While we instructed the LLM to avoid such question ambiguity during dataset creation, this cannot be guaranteed, and exhaustive annotation across numerous evidence articles is impractical. 
Similar to challenge datasets \citep{mccoy-etal-2019-right, schuster2019towards, gardner-etal-2020-evaluating}, we do not fine-tune models on \dataset{} to prevent them from reverse-engineering its specific question generation strategies. Instead, we define the task as zero-shot. However, we provide a separate development set for prompt selection to avoid overfitting to the test set.

\section{\dataset{} Construction}
\label{sec:dataset-construction}
Our goal is to create a stable QA dataset based on fictional event timelines where LLMs with updated parametric knowledge have no unfair advantages.
Creating \dataset{} involves three steps:
(1) generating independent fictional timelines of ten events, (2) writing news articles about the events, and (3) creating questions about them. Questions and news articles are independently generated and linked to specific information sentences in the events, allowing the automatic generation of instances with questions with sufficient or insufficient evidence, and with distracting documents. We use GPT-4o to generate all parts of \dataset{}.

\subsection{Timeline Generation}
\label{sec:dataset-construction:timelines}
The fictional world of \dataset{} uses independent timelines that mirror real-world event progression \citep{shahaf2010connecting, pratapa-etal-2023-background}, following sequential events in narrative literature \citep{keith2023survey}. Unlike tree-like structures with divergent subplots \citep{liu2017growing}, this approach conditions each event on all previous events in the same storyline, reducing the risk of introducing inconsistencies.
Each event includes a \emph{date}, an \emph{outline} describing the plot, and a KB of fictional \emph{named entities}, which is updated after each event. Each outline contains 20-30 single-sentence \emph{outline items}, each providing distinct details about the event. Examples are provided in Appendix~\ref{appendix:timeline-example}.

\input{tables/tab_question_examples}

\input{figures/figure-data-storyline-generation}
The timeline generation adapts steps from story generation methods \citep{yang2022doc,lee2024navigating,zhu2023end}, but differs as it avoids classic narrative templates with protagonists and antagonists. Specifically, it begins with an LLM-generated seed summary and news genre, followed by three core steps using multiple %
prompts: (1) creating, checking, and refining the outline, (2) generating and updating fictional named entities, and (3) producing a new seed summary for the next event. 
We employ heuristics to detect and correct errors by critiquing the LLM outputs~\citep{gou2024critic}. 
Events are generated sequentially, conditioned on prior events, named entities, and the latest summary sentence (Figure~\ref{fig:figure-data-storyline-generation}).  To help the LLM maintain consistency with named entities, we provide all past event outlines with resolved entities and updated knowledge base entries, mirroring neural representation approaches \citep{clark-etal-2018-neural}.

The boundary between a ``realistic'' fictional world and reality is blurred due to inevitably overlapping concepts. For example, common sense (e.g., what ``rain'' is) and physical laws must still apply in the fictional world, while unrealistic elements, like dragons, must not exist. We define two practical criteria for fictional worlds to minimize the impact of updated parametric knowledge from real-world data while preserving common sense and physical laws: a) fictional named entities and b) mutually exclusive sampling for seed summaries of subsequent events.
First, \dataset{} distinguishes seven types of named entities from \citet{ling2012fine} with one extra type for ``miscellaneous''. We compare each named entity against Wikipedia to ensure it doesn't overlap with well-known real-world entities.  However, this alone is insufficient, as the LLM might generate timelines based on its parametric knowledge, aligning with real-world events but simply replacing the named entities. 
To prevent this, the LLM generates multiple mutually exclusive summaries for subsequent events, from which one is randomly selected. This way, the timeline follows irreversible paths and prevents the LLM from generating summaries that closely resemble real-world events from its parametric knowledge.
For example, a possible  (not chosen)  seed summary for the second event in Figure~\ref{fig:timeline} was, that after the initial accusations in the first event, Amber Glaze Delights suspended operations for an internal audit. 

News articles about the events in each timeline serve as evidence documents in \dataset{}.
The LLM generates news articles with four profiles  (``progressive'' ``conservative'', ``objective'', ``sensational'') to simulate how organizations emphasize different aspects in their reporting \citep{fan-etal-2019-plain}. For each profile, the model (1) selects three subsets of multiple outline items from each individual event outline, (2) drafts an article for each selection, and (3) verifies that all information from these outline items is included in the final article. The generation process is detailed in Appendix~\ref{appendix:dataset-construction}.

\subsection{Question and Answer Generation}
\label{sec:dataset-construction:questions}
\dataset{} includes four question types. These questions require distinct evidence-based reasoning skills. The models must provide accurate answers when possible and detect when the answer cannot be determined due to unverifiable, contradictory, or missing evidence.
(1) \textbf{Time-span} questions involve temporal reasoning to calculate the duration between outline items across up to two events (Example: \emph{``How many days passed between the announcement of the public forum by the Calder Square Cultural Committee and the day the forum was held?''}). Both can be used to form answerable instances (with sufficient evidence) or unanswerable (with insufficient evidence).
(2) \textbf{Multi-hop} questions use a fictional named entity as a bridge entity \citep{yang2018hotpotqa,tang2024multihop} to link information from two sentences (see example in Figure~\ref{fig:timeline}). 
From the multi-hop questions, we create two types of questions that are always unanswerable, regardless of the evidence: (3) \textbf{False premise} questions have incorrect assumptions that directly contradict the evidence. This differs from false premise  questions in other works~\citep{yang2024crag} where the assumptions contradict general world knowledge. 
(4) \textbf{Uncertain specificity} questions ask for details that are too specific to be answered by the available evidence in the fictional timeline. Examples are shown in Table~\ref{tab:question-type-examples}.

To generate answerable questions, the LLM selects two outline items from up to two distinct events. Based on these outline items it then generates the question and correct answer, as well as five distractors, framing the question as if asked after the most recent event.  We instruct the LLM to ensure (1) no other outline item can answer the question, and (2) both selected outline items are essential for a definite answer, with additional outline items added if needed. We provide all past event outlines as context to help the LLM avoid drafting ambiguous questions that could be answered differently using other information. For each multi-hop question, we instruct the LLM to generate multiple false premise and uncertain specificity questions using the same answer options, adding subtle contradictions or unverifiable details. These unanswerable questions share the same answer options as the original multi-hop question.
Knowing what information is needed to answer a question and which news article contains it, allows us to combine questions with news articles to create scenarios where evidence is sufficient for an answer or insufficient, requiring the model to deflect.
See Appendix~\ref{appendix:dataset-construction:questions} for details on the generation process, and Figure~\ref{fig:instance-example} for a complete instance with question, answer options and news articles as evidence.

\input{tables/tab_fitiqa_stats}
\subsection{Quality Filtering and Assessment}
To automatically link questions with news articles and determine their answerability, two key requirements must be met:
\begin{itemize}[noitemsep]
    \item \textbf{Requirement 1:} The selected outline items for creating each answerable question must be both fully sufficient and necessary to answer the question with certainty.
    \item \textbf{Requirement 2:} News articles must convey all factual information from the selected outline items and exclude any information from the non-selected outline items.
\end{itemize}
Requirement 1 can be violated when questions depend on information beyond the selected outline items, as the model had access to all events during generation. To mitigate this, we remove all 1,122 questions (42.5\%) that the LLM, which generated the question, cannot answer correctly itself using only the selected outline items as evidence (see Appendix~\ref{appendix:quality:question-filtering}).
For Requirement 2, we use a pre-trained T5 NLI model \citep{honovich-etal-2022-true-evaluating} to verify that selected sentences are entailed by the news article, while unselected ones are not. The model agreed with the assumed entailment labels in 98.1\% of selected sentences and 92.2\% of unselected sentences. 7.3\% of the remaining unselected sentences did not receive any label instead of a label disagreement (see Appendix~\ref{appendix:quality:nli}).
Lastly, we conduct human annotation on 350 instances to verify the correctness of the reference answer. Majority voting from three annotators agreed with our labels 94\% of the time. Fleiss's kappa of 0.516 indicates moderate agreement, underscoring the task's difficulty even for humans (see Appendix~\ref{appendix:quality:annotation} for details).
Table~\ref{tab:fitiqa-stats} summarizes the statistics of \dataset{} dataset after quality filtering.
We use three timelines as the development set for prompt selection, and the rest twelve timelines as the test set.

%% file: tables/tab_question_examples.tex
\definecolor{darkgreen}{rgb}{0,0.5,0} %
\definecolor{darkred}{rgb}{0.5,0,0}   %

\begin{table*}[]
\small
\centering
\begin{tabularx}{\textwidth}{l X}
    \toprule
    \textbf{Type} & \textbf{Example Text} \\
    \midrule
    \textbf{Evidence 1} & Selvia Renek released an investigative piece on her blog questioning Amber Glaze Delights’ transparency, which gained traction on social media under the hashtag \#SourcingScandal.\\
    \addlinespace
    \textbf{Evidence 2} & Selvia Renek published an article questioning whether the certification program would be \textbf{\textcolor{red}{robust enough}} to address systemic issues and called for increased community participation in refining the criteria. \\
    \midrule
    \begin{tabular}[t]{@{}l@{}} 
        \textbf{Multi-Hop} \\ 
        \textbf{Question} (\textcolor{darkgreen}{\(\checkmark\)})
    \end{tabular} & What did the author of an investigative piece associated with the hashtag \#SourcingScandal question about the ethical certification program? \\
    \addlinespace
    \begin{tabular}[t]{@{}l@{}} 
        \textbf{False} \\ 
        \textbf{Premise} (\textcolor{darkred}{\(\times\)}) 
    \end{tabular}  & What did the author of an investigative piece associated with the hashtag \#SourcingScandal question     \textbf{\textcolor{red}{about the cost}} of the ethical certification program? \\
        \addlinespace
    \begin{tabular}[t]{@{}l@{}} 
        \textbf{Uncertain} \\ 
        \textbf{Specificity} (\textcolor{darkred}{\(\times\)}) 
    \end{tabular}   & What did the author of an investigative piece associated with the hashtag \#SourcingScandal question about the certification program’s \textbf{\textcolor{blue}{ability to address exploitative labor practices}}? \\
    \bottomrule
\end{tabularx}
\caption{An answerable multi-hop question (parent question) and unanswerable questions. The false premise question contradicts the evidence (\textcolor{red}{\textbf{red}}). The uncertain specificity question introduces unverifiable details (\textcolor{blue}{\textbf{blue}}).}
\label{tab:question-type-examples}
\end{table*}

%% file: figures/figure-data-storyline-generation.tex
\begin{figure}[t]
\centering
  \includegraphics[width=\columnwidth]{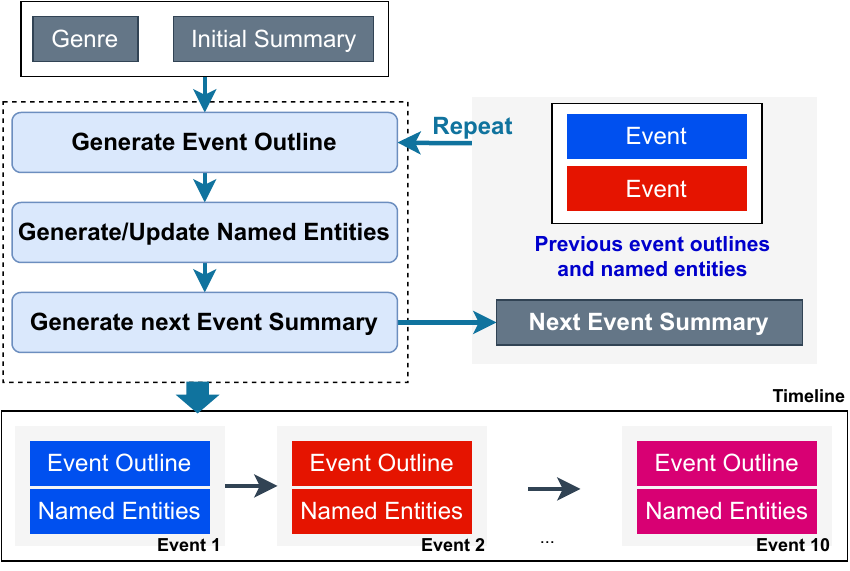}
  \caption{Events are generated sequentially based on a summary sentence and the previously generated events.}
  \label{fig:figure-data-storyline-generation}
\end{figure}

%% file: tables/tab_fitiqa_stats.tex
\begin{table}[t]
\small
\centering
\begin{tabular}{llrr} %
\toprule
& & \textbf{Overall} & \textbf{Per Timeline} \\
\midrule
\multirow{4}{*}{\rotatebox[origin=c]{90}{\textsc{World}}} & \textbf{Timelines} & 15 & 1 \\
& \textbf{Events} & 150 & 10 \\
& \textbf{Outline Sentences} & 3,174 & 211.6 \\
& \textbf{Named Entities} & 393 & 26.2 \\
\midrule
\multirow{5}{*}{\rotatebox[origin=c]{90}{\textsc{Task}}} & \textbf{Multi-hop} & 839 & 55.9\\
& \textbf{Time-span } & 678 & 45.2\\
& \textbf{False premise} & 2,879 & 191.9\\
& \textbf{Uncertain specificity} & 2,952 & 196.8\\
\cmidrule(lr){2-4} %
& \textbf{News articles} & 1,800 & 150.0 \\
\bottomrule
\end{tabular}
\caption{Summary statistics of elements in \dataset{}.}
\label{tab:fitiqa-stats}
\end{table}

%% file: latex/content/05__experiments.tex
\section{Main Experiments}
\label{sec:benchmark-experiments}

\input{tables/tab_dataset_stats}

\input{tables/tab_main_results}

In our main experiments, we form multiple-choice instances by combining each question with all news articles available up to the question date. 
For multi-hop and time-span questions, we create one instance with complete evidence—where the model has access to all relevant information—and several instances with systematically reduced evidence. These \emph{insufficient-evidence} instances are generated by omitting specific news articles, each corresponding to a different outline sentence necessary to answer the question. When the evidence is complete, the model is expected to select the correct answer; when it is insufficient, the correct response is to choose the deflection option.
Following \citet{schuster-etal-2021-get}, this approach presents the same question with varying evidence, leading to different correct answers and requiring the model to use the evidence to perform well.
Additionally, we include 2,500 instances with false premise and uncertain specificity questions, each paired with the same full set of news articles as the answerable instances. Details on instance creation are provided in Appendix~\ref{appendix:combining:benchmark}, with overall statistics in Table~\ref{tab:dataset-statistics-fitibench}.
Instances can include up to 120 documents, with a total of 1,349 to 45,484 tokens across both the question and evidence documents. This requires LLMs with sufficiently large context windows.
 We evaluate several open-source consumer-sized LLMs with up to 32B parameters: Qwen2.5~\citep{yang2024qwen2} (7B, 14B, 32B), Phi3~\citep{abdin2024phi} (mini, small, medium), and Phi3.5 MoE. All these models support context sizes of at least 128k tokens. 
Prompts were selected per LLM using ADTScore (see below) on the development set (see Appendix~\ref{appendix:experiments:prompt-tuning}). Following \citet{levy-etal-2024-task}, we also test Chain-of-Thought (CoT) prompting \citep{wei2022chain} with the elicitation string \citep{zhou2023large} to evaluate its effect on LLM performance.
As our primary metric, we introduce ADTScore (\underline{A}nswer \underline{D}eflection \underline{T}radeoff Score), defined as the harmonic mean of accuracy for answerable instances
($acc_{a}$) 
and unanswerable instances where the model must select the ``unanswerable'' option to deflect
($acc_{u}$):
\[
\mathrm{ADTScore} = \frac{2 \times acc_a \times acc_u}{acc_a + acc_u}
\]
ADTScore is robust to class imbalance of \textit{answerable} and \textit{non-answerable} instances and achieves the maximum performance when model accuracy is balanced across both subsets.

\paragraph{Results}
Table~\ref{tab:main_results} shows that while all LLMs handle multi-hop questions well, performance on time-span questions improves with model size. Unanswerable questions where the model must deflect are most challenging, especially for Phi3-based models. Qwen2.5 with 32B performs best overall with and ADTScore of 53.2 but still struggles to detect the subtle inconsistencies with false premise and uncertain specificity questions. Similar to prior work~\citep{levy-etal-2024-task} we observe mixed effects on long contexts with CoT prompting, which mostly improves the performance of Phi3-based models by increasing their deflections, which boosts the overall score but harms multi-hop performance. The high answer parsing rate (97.9\% with CoT, 99.2\% without) suggests that mistakes arise from reasoning errors rather than poor instruction-following (see Appendix~\ref{appendix:parsing}).

\paragraph{Insufficient Evidence}
\input{figures/figure-analysis-insufficient-evidence-benchmark}
Multi-hop questions can lack sufficient evidence in three ways: (a) missing answer information (purple in Figure~\ref{fig:timeline}), (b) missing bridge entity information (green), or (c) missing both. The third case (c) occurred in 124 instances where both required information pieces were in the same article and removed together. When the answer itself was missing (cases a \& c), the task resembled the IDK task by \citet{vodrahalli2024michelangelo}.
However, missing only the bridge entity (b) was the most challenging. In these cases, models often inferred the correct answer through shortcuts rather than recognizing the evidence as incomplete and deflecting appropriately. Figure~\ref{fig:accuracy-per-insufficient-category} shows performance by evidence type (in all cases the distracting documents up to the question date remained). Models found it easiest to deflect when all relevant information was absent but struggled most when the answer was present while the bridge entity was missing.
The errors followed a consistent pattern across models. When the answer itself was missing, they were more likely to select a misleading option (52.9\%-77.9\%). When the bridge entity was missing, models frequently answered as if nothing were missing (69.7\%-90.7\% of errors).
In general, we observed that LLMs tend to overlook subtle differences between questions and evidence. Except for Phi3 (mini \& small), which struggled with deflection, we found a significant negative association between accuracy on multi-hop questions with sufficient evidence and questions where bridge entity information was omitted or where questions were manipulated into false premise or uncertain specificity (phi coefficient: $\phi = -0.114$ to $-0.374$, \textit{p} < 0.001). For details, see Appendix~\ref{appendix:experiments:error-analysis-insufficient-multi-hop} and \ref{appendix:experiments:error-analysis-timespan}.

\paragraph{GPT-4 Turbo}
To compare with our RealTimeQA experiments in Section~\ref{sec:impact-of-parametric-knowledge}, we tasked GPT-4 Turbo with answering questions without evidence. We randomly sampled 250 multi-hop and time-span questions and converted them into four-way multiple choice format (excluding the deflection option). Accuracy was near random for time-span questions (24.4\%) but higher for multi-hop questions (53.6\%). Upon manual inspection of correctly predicted questions without evidence, we found no obvious give-away information in the question or answer options, and hypothesize that this is due to the synthetic data generation (see Appendix~\ref{appendix:experiments:gpt4turbo} for examples). A model may benefit from learned token probabilities during inference because the dataset was sampled from the same token distribution. As discussed in Section~\ref{sec:dataset-construction:timelines}, a clean separation between fictional and real-world knowledge is unrealistic. This highlights the importance of controlled experiments where reliance on parametric knowledge is penalized.
We did not evaluate GPT-4 Turbo on the full dataset due to high computational costs and its large context requirements, especially since the data was generated by GPT-4o. For informativeness, we estimated its performance on 499 randomly sampled instances matching the question type distribution in Table~\ref{tab:main_results}. GPT-4 Turbo achieved an ADTScore of 42.4. It performed well on answerable questions (88\% for multi-hop, 84\% for time-span) but struggled with unanswerable ones—scoring just 25.3\% on multi-hop with insufficient evidence, and 15\% on both false premise and uncertain specificity questions. An exception was time-span questions with insufficient evidence, where it reached 58\% accuracy.

%% file: tables/tab_dataset_stats.tex
\begin{table}[t]
\small
    \centering
    \begin{tabular}{rcc}
    \toprule
    & \textbf{Answerable}  & \textbf{Unanswerable} \\
    \midrule
\textbf{Time-span} & 532 & 1,063 \\
\textbf{Multi-hop} & 625 & 1,239  \\
\textbf{False premise} &  -- & 1,250   \\
\textbf{Uncertain specificity} & -- & 1,250  \\
\midrule
\textbf{All instances} & \textbf{1,157} & \textbf{4,802} \\
      \bottomrule
    \end{tabular}
    \caption{Instances used for benchmarking experiments.}
    \label{tab:dataset-statistics-fitibench}
\end{table}

%% file: tables/tab_main_results.tex
\begin{table*}[h!]
\small
    \centering
    \begin{tabular}{lcccccccc}
    \toprule
    \multicolumn{2}{c}{\textbf{}} & \multicolumn{2}{c}{\emph{Answerable}} & \multicolumn{4}{c}{\emph{Unanswerable}} \\
    \textbf{Model} & \textbf{ADTScore} & \textbf{Multi H.} & \textbf{Time S.} & \textbf{Multi H.} & \textbf{Time S.} & \textbf{False P.} & \textbf{Uncertain S.} \\
    \cmidrule(lr){1-1} \cmidrule(lr){2-2} \cmidrule(lr){3-4} \cmidrule(lr){5-8}
    \emph{Random} & 14.3 & 14.3 & 14.3 & 14.3 & 14.3 & 14.3 & 14.3 \\
    \midrule
    Phi3 mini (3.8B) & 12.9 & 79.8 & 21.0 & 3.1 & 27.4 & 0.6 & 1.2 \\
    Phi3 small (7B) & 23.5 & 80.5 & 34.8 & 10.2 & 39.5 & 8.2 & 4.5 \\
    Phi3 medium (14B) & 19.9 & 79.7 & 53.8 & 16.6 & 17.1 & 8.3 & 5.5 \\
    Phi3.5 MoE (16$\times$3.8B) & 32.4 & \textbf{82.9} & 44.7 & 16.5 & 54.3 & 13.4 & 6.7 \\
    Qwen2.5 (7B) & 31.5 & 67.5 & 42.9 & 26.7 & 23.1 & 16.2 & 21.8 \\
    Qwen2.5 (14B) & 51.6 & 76.3 & 66.0 & \textbf{44.9} & 41.0 & \textbf{42.8} & \textbf{32.7} \\
    Qwen2.5 (32B) & \textbf{53.2} & 79.4 & 67.3 & 41.7 & \textbf{62.1} & 38.6 & 26.7 \\

    \midrule
    Phi3 mini (3.8B) + CoT & 41.9 & 53.8 & 29.1 & 42.7 & 43.8 & 41.9 & 37.7 \\
    Phi3 small (7B) + CoT& 26.6 & 81.6 & 40.8 & 11.4 & 47.3 & 8.4 & 5.0 \\
    Phi3 medium (14B) + CoT & 24.6 & 72.9 & 56.0 & 18.7 & 15.9 & 14.1 & 12.2 \\
    Phi3.5 MoE  (16$\times$3.8B) + CoT& 38.0 & 78.1 & 46.1 & 23.3 & 54.4 & 19.4 & 15.5 \\
    Qwen2.5 (7B) + CoT & 31.3 & 63.4 & 38.9 & 24.9 & 24.3 & 21.8 & 18.7 \\
    Qwen2.5 (14B) + CoT & 49.6 & 75.8 & 66.2 & 42.0 & 41.9 & 39.3 & 29.3 \\
    Qwen2.5 (32B) + CoT & 48.2 & 82.7 & \textbf{69.5} & 36.8 & 55.7 & 31.8 & 19.4 \\
    \bottomrule
    \end{tabular}
    
    \caption{\dataset{} evaluation results for multi-hop, time-span, false premise, and uncertain specificity questions with all evidence up to the question date with/without CoT prompts. Metrics: ADTScore and accuracy by question type. %
    }
    \label{tab:main_results}
\end{table*}

%% file: figures/figure-analysis-insufficient-evidence-benchmark.tex
\begin{figure}[t]
\centering
  \includegraphics[width=\columnwidth]{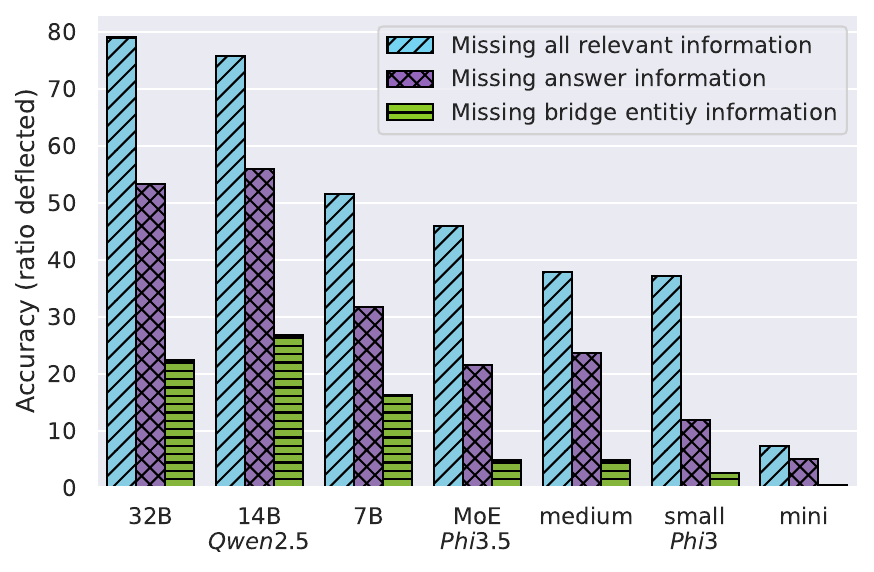}
  \caption{Model deflection ratio in multi-hop questions with varying evidence gaps.}
  \label{fig:accuracy-per-insufficient-category}
\end{figure}

%% file: latex/content/06__analysis.tex
\section{Impact of Irrelevant Documents}
\label{sec:irrelevant-documents-ablation}
\input{figures/figure-context-ablation-agg}

To evaluate the effect of irrelevant documents,
we reuse the same questions---with sufficient and insufficient evidence, and unanswerable cases---and vary the number of irrelevant news articles from 0 to 80 in increments of 20. This results in 10,210 instances (Appendix~\ref{appendix:combining:context-ablation}).
Figure \ref{fig:context-length-ablation-all} shows the overall ADTScore and the aggregated accuracy for answerable and unanswerable instances. Across all models and configurations, performance is best when no irrelevant documents are present and declines as irrelevant documents are added. For answerable questions with only relevant evidence, all models perform in similar ranges. The smallest models in our experiments (Phi3 mini and Qwen2.5 7B) experience the sharpest decline, while the larger models are more robust. %
For each model, the major drop in performance occurs within the first 20 added irrelevant documents and then stabilizes, with the two larger Qwen2.5 models performing best. See Appendix~\ref{appendix:experiments:error-analysis-irrelevant-doc-numbers} for visualizations per model.

\input{figures/figure-sankey-sufficient-insufficient}
\paragraph{Prediction changes for insufficient evidence}
Figure~\ref{fig:sankey-multi-hop-insufficient} shows how multi-hop question predictions change when either answer or bridge entity information is omitted. The ideal model always predicts the correct answer (left , blue), when all evidence is available, and deflects (right, orange), when not. Without distracting news articles (top), Qwen2.5 14B mostly deflects with insufficient evidence, while Phi3 (medium) often selects distractors when the answer is missing or maintains its original prediction when bridge entity information is absent. Adding 80 irrelevant articles (bottom) decreases performance, but trends remain similar when answer information is omitted. When bridge entity information is missing, Qwen2.5 14B also predicts as though evidence were sufficient, highlighting the challenge of detecting insufficient evidence when distracting documents are present. Appendix~\ref{appendix:experiments:error-analysis-prediction-change-fp-us} shows the prediction changes for false premise and uncertain specificity questions.

%% file: figures/figure-context-ablation-agg.tex
\begin{figure}[t]
\centering
  \includegraphics[width=1\linewidth]{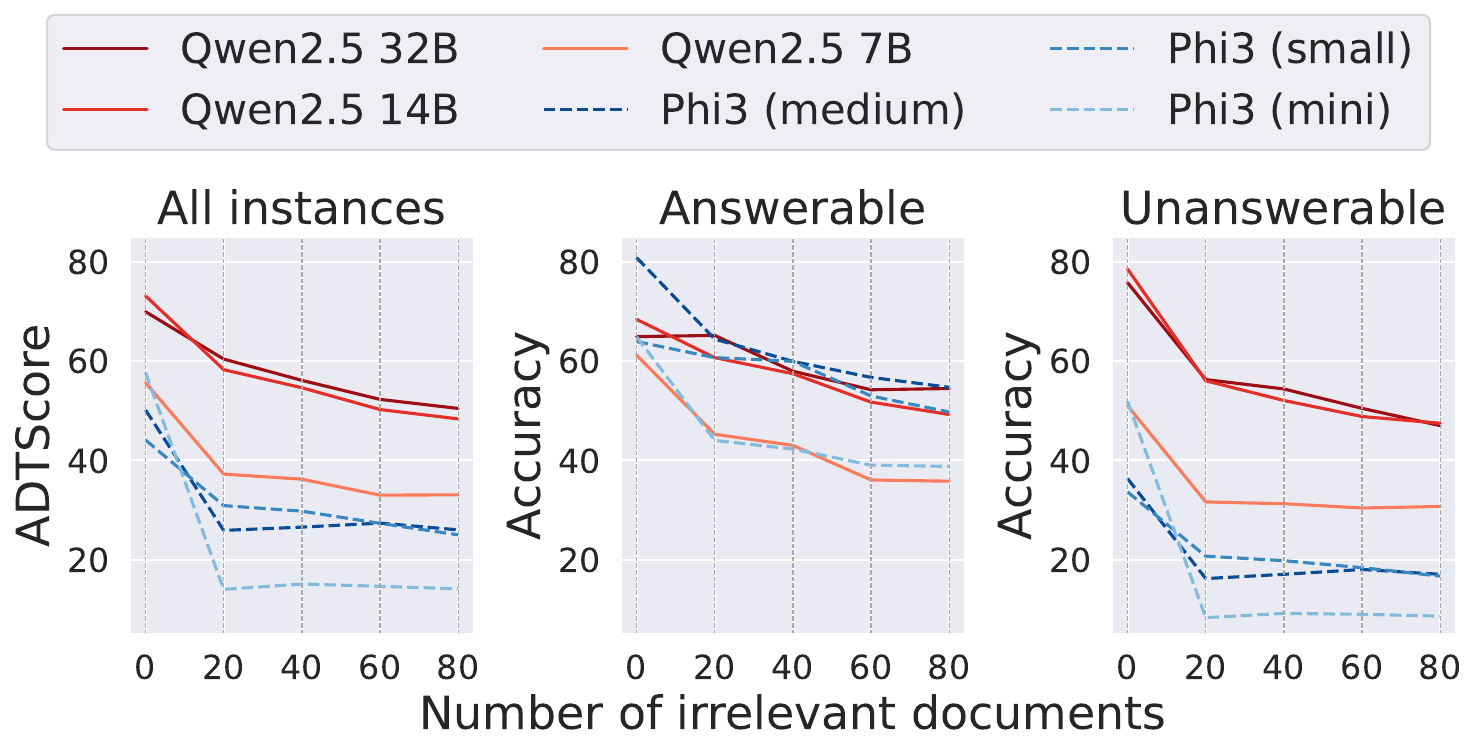}
  \caption{Performance over all instances (left), answerable (center), and unanswerable (right) instances with increasing number of irrelevant documents.}
  \label{fig:context-length-ablation-all}
\end{figure}

%% file: figures/figure-sankey-sufficient-insufficient.tex
\definecolor{sankey_correct}{RGB}{31,119,180}
\definecolor{sankey_deflect}{RGB}{255, 127, 14}

\begin{figure*}[htbp]
    \centering
    \begin{subfigure}[t]{0.6\linewidth}  %
        \centering
        \includegraphics[width=\linewidth]{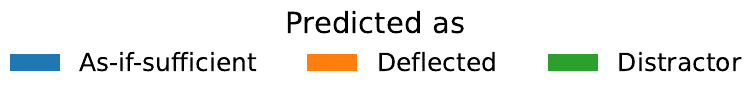} %
    \end{subfigure}

    \begin{subfigure}[t]{0.25\linewidth}
        \centering
        \includegraphics[width=\linewidth]{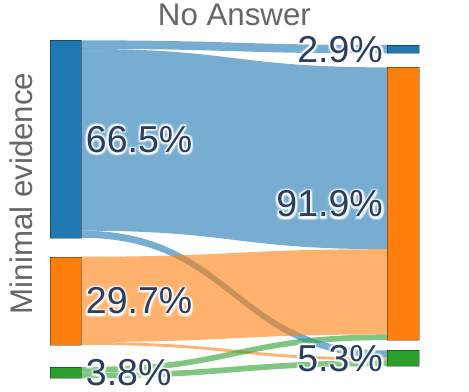}
    \end{subfigure}
    \hspace{-4.6mm}
    \begin{subfigure}[t]{0.25\linewidth}
        \centering
        \includegraphics[width=\linewidth]{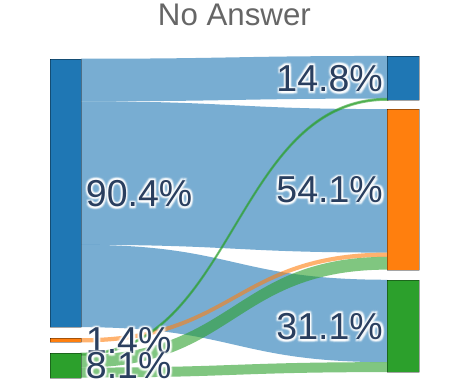}
    \end{subfigure}
    \hspace{-4.6mm}
    \begin{subfigure}[t]{0.25\linewidth}
        \centering
        \includegraphics[width=\linewidth]{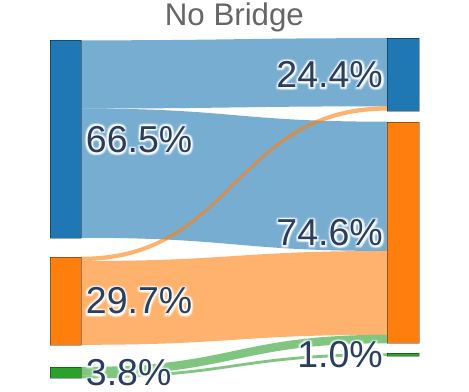}
    \end{subfigure}
    \hspace{-4.6mm}
    \begin{subfigure}[t]{0.25\linewidth}
        \centering
        \includegraphics[width=\linewidth]{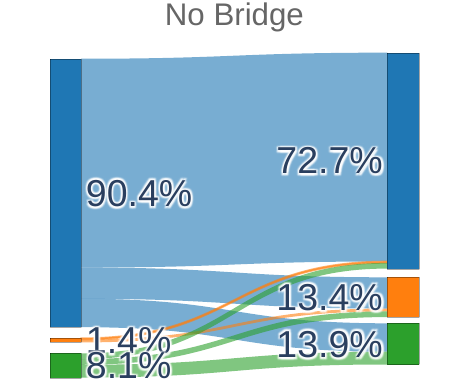}
    \end{subfigure}

    \begin{subfigure}[t]{0.25\linewidth}
        \centering
        \includegraphics[width=\linewidth]{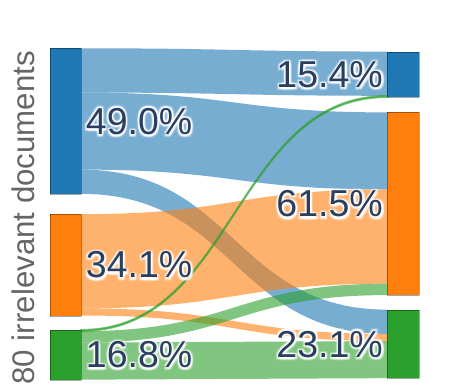}
        \caption{\textbf{Qwen2.5 14B}}
    \end{subfigure}
    \hspace{-4.6mm}
    \begin{subfigure}[t]{0.25\linewidth}
        \centering
        \includegraphics[width=\linewidth]{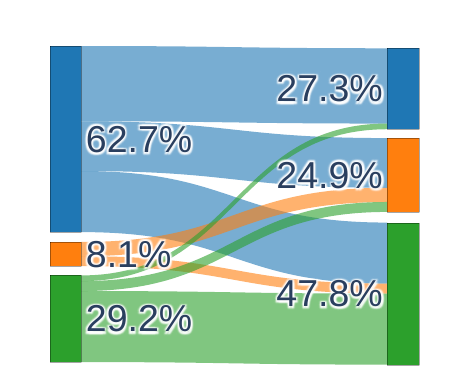}
        \caption{\textbf{Phi3 (med.)}}
    \end{subfigure}
    \hspace{-4.6mm}
    \begin{subfigure}[t]{0.25\linewidth}
        \centering
        \includegraphics[width=\linewidth]{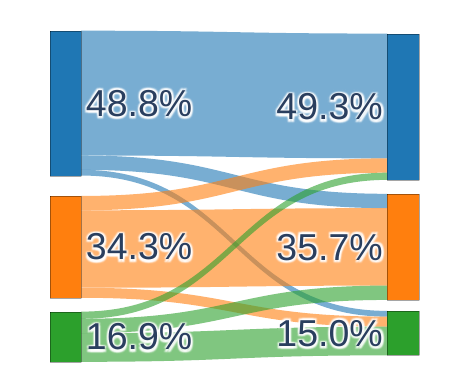}
        \caption{\textbf{Qwen2.5 14B}}
    \end{subfigure}
    \hspace{-4.6mm}
    \begin{subfigure}[t]{0.25\linewidth}
        \centering
        \includegraphics[width=\linewidth]{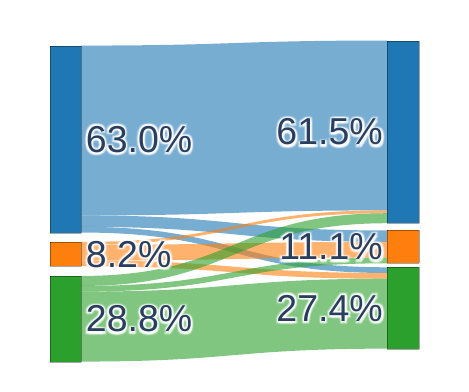}
        \caption{\textbf{Phi3 (med.)}}
    \end{subfigure}

    \caption{Multi-hop question predictions change after removing key information containing the answer (\emph{``No Answer''}) or the bridge entity (\emph{``No Bridge''}). When evidence is sufficient (\textbf{left} in each diagram), the model must  always predict the \textcolor{sankey_correct}{\textbf{correct answer}}. When evidence is insufficient (\textbf{right} in each diagram), the model must \textcolor{sankey_deflect}{\textbf{deflect}}.}
    \label{fig:sankey-multi-hop-insufficient}
\end{figure*}

%% file: latex/content/01__related_work.tex
\section{Related Work}

Several recent studies addressed parametric knowledge interference through time-sensitive questions~\citep{vu2023freshllms,yang2024crag}, dataset updates~\citep{kasai2024realtime}, or automatic dataset generation from (recent) real-world data~\citep{liska2022streamingqa,tang2024multihop,guinet2024automated}. 
However, updated datasets introduce different instances, making direct performance comparisons over time unreliable, as fluctuations may result from dataset variations from different times \citep{luu-etal-2022-time}.
Other approaches focus on time-invariant questions~\citep{wei2024measuring} or conflicts between external and parametric knowledge \citep{longpre-etal-2021-entity, neeman-etal-2023-disentqa, tan-etal-2024-blinded, xu-etal-2024-knowledge-conflicts}. In contrast, \dataset{} generates a self-contained world, independent of real-world events and entities, to provide a stable test-bed despite parametric knowledge updates.
Our work extends needle-in-the-haystack tasks, where models must locate and reason over information in long text~\citep{Kamradt2023,levy-etal-2024-task,kuratov2024babilong}. These benchmarks are often based on real-world information or literature~\citep{shaham-etal-2022-scrolls,shaham-etal-2023-zeroscrolls,bai-etal-2024-longbench,an-etal-2024-l,liu-etal-2024-lost,wang2024novelqa,hilgert-etal-2024-evaluating}, where parametric knowledge can interfere. Some mitigate this by constructing datasets from recent information~\citep{li2023loogle,karpinska-etal-2024-one}. Apart from RGB~\citep{chen2024benchmarking}, which heuristically determines relevant evidence, these datasets focus solely on answerable questions.
Closest to our work is Michelangelo~\citep{vodrahalli2024michelangelo}, which evaluates LLMs’ long-context abilities using synthetic data outside their pretraining set and includes IDK questions where the answer is not in the text. \dataset{} goes further by generating parallel worlds with recurring named entities and unanswerable questions with subtle mismatches with the grounding, rather than merely omitting explicit answers.
Unanswerable questions 
have been studied in the context of adversarial manipulation \citep{rajpurkar-etal-2018-know,sulem-etal-2021-know-dont,gautam-etal-2023-lightweight}, missing information in multi-hop reasoning \citep{trivedi-etal-2020-multihop,trivedi-etal-2022-musique,atanasova-2022-insufficient}, and false premises based on incorrect assumptions \citep{kim-etal-2021-linguist,yu-etal-2023-crepe,hu-etal-2023-wont,yang2024crag}. Similarly, we explore these challenges but focus on external evidence in a parallel world, where parametric knowledge can not detect flawed assumptions nor compensate for imperfect evidence.

%% file: latex/content/07__conclusion.tex
\section{Conclusion and Future Work}
We introduce \dataset{}, a novel dataset featuring out-of-training event timelines and question-answer pairs that are independent of real-world events.
\dataset{} serves as a robust platform for evaluating evidence-based question answering, as it requires models to answer questions exclusively from evidence and only when sufficient evidence is available.
By automatically pairing questions and news articles,  \dataset{} simulates various retrieval conditions, ranging from scenarios with sufficient evidence to those with insufficient or irrelevant evidence.
Our experiments across seven models reveal significant challenges in evidence-based reasoning: when key evidence required to answer a question is missing, models frequently resort to shortcut reasoning, a critical shortcoming for trustworthy applications.
Future work may expand \dataset{} with new questions 
and develop trustworthy models that reliably perform evidence-based reasoning.

%% file: latex/appendix/realtimeqa-experiments.tex
\section{RealtimeQA Experiments}
\label{appendix:realtimeqa-experiments}
The RealtimeQA dataset spans weekly news quizzes from June 16, 2022, to January 12, 2024. We select 1,548 questions with four answer options where gold evidence is provided, enabling direct comparison of model performance using evidence versus parametric knowledge.
Our experiments use GPT-4 Turbo (``gpt-4-turbo-2024-04-09'') with the reported  knowledge cutoff in December, 2023. 
To include sufficient questions beyond the reported cutoffs, we collected 660 additional instances from January 18 to September 13, 2024, via the Wayback Machine\footnote{\url{https://web.archive.org/}} using the same sources as RealtimeQA, totaling 2,208 questions.
It is important to emphasize that this experiment cannot definitively attribute the observed decrease in performance to reduced parametric knowledge alone, as it may also result from variations in the instances themselves. Nonetheless, the consistent downward trend across all models over time strongly suggests that the decline is primarily due to less useful parametric knowledge.

%% file: latex/appendix/timeline-example.tex
\section{Timeline Example}
\label{appendix:timeline-example}
\begin{figure*}[h]
      \includegraphics[width=1\linewidth]{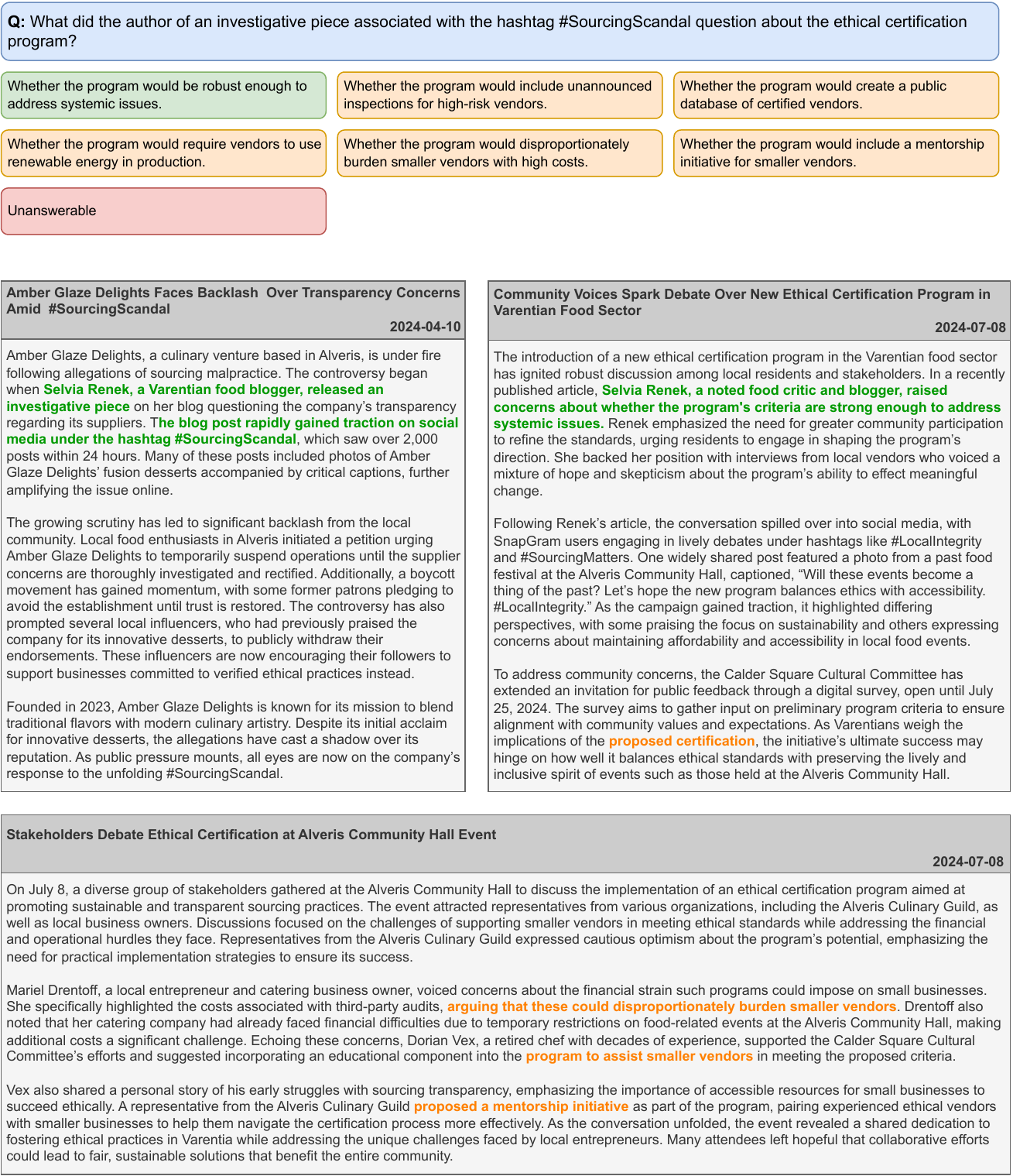}
    \caption{An example of a multi-hop question using three news articles as evidence: the first two articles must be combined (\emph{green}) to answer the question. Individually, they lack sufficient information. The third article is irrelevant and may mislead the LLM (\emph{orange}) toward an incorrect answer.}
    \label{fig:instance-example}
\end{figure*}

The complete outline of the first event from Figure~\ref{fig:timeline} is shown in Figure~\ref{fig:outline-example}. Each outline item has a unique ID within the timeline and conveys distinct, specific information. Table~\ref{tab:entity-example} shows the updated KB entry after the final event for the fictional person Selvia Renek. Figure~\ref{fig:news-example} shows a news article with resolved named entities. Figure~\ref{fig:instance-example} shows an complete instance with a question, answer options and news articles as evidence.

\begin{table*}[h]
\small
    \centering
    \begin{tabularx}{\linewidth}{l|X}
    \toprule
    \textbf{ID} & PERSON-3 \\
    \textbf{Name} & Selvia Renek \\
    \textbf{Entity Class} & Person \\
    \textbf{Description} & Selvia Renek is an enthusiastic and detail-oriented food critic who focuses on celebrating local culinary creativity. \\
    \textbf{Date of Birth} & 1992-03-09 \\
    \textbf{Gender} & Female \\
    \textbf{Profession} & Food Blogger \\
    \textbf{Nationality} & Varentian \\
    \textbf{Education} & Bachelors in Journalism, University of Alveris \\
    \textbf{Height} & 160 cm \\
    \textbf{Hair Color} & Auburn \\
    \textbf{Eye Color} & Blue \\
    \textbf{Affiliation} & Progressive \\
    \textbf{Marital Status} & Single \\
    \midrule
    & \textbf{History} \\
    \midrule
    \textbf{2024-03-15} &  Praised Amber Silk as “the most delicate balance of flavors” and contributed to the event's buzz on social media. \\

        \textbf{2024-04-10} & Published an investigative blog post highlighting unethical practices in Amber Glaze Delights' supply chain. \\

        \textbf{2024-07-03}  &  Published an op-ed discussing implications of Amber Glaze Delights' audit findings and emphasized systemic change in sourcing oversight. \\

        \textbf{2024-07-08}  &  Published an article questioning the robustness of ethical certification programs and called for increased community participation in refining criteria. \\

        \textbf{2024-07-09}  &  Published a blog spotlighting Amara Hearth Café's plans to adopt ethical sourcing practices, including an interview with Erena Treflin. \\

        \textbf{2024-07-10}  &  Published a SnapGram post summarizing the forum, highlighting tensions between ethical practices and accessibility for smaller businesses. \\

        \textbf{2024-07-21} &  Hosted a live SnapGram Q\&A session to address community concerns about the Calder Square Cultural Committee's ethical certification program. \\

        \textbf{2024-08-01} &  Published a blog detailing the rollout of the self-certification program and included interviews with representatives and vendors. \\

      \bottomrule
    \end{tabularx}
    \caption{The updated KB entry for the fictional person Selvia Renek after the last event of the timeline. For each event in which Selvia Renek participated, the history summarizes her role.}
    \label{tab:entity-example}
\end{table*}

\begin{figure*}[h]
    \centering
    \small
    \begin{tcolorbox}[colback=blue!10!white, colframe=blue!80!black, coltitle=black, 
                      boxrule=0.3mm, left=10mm, right=10mm] %
        \begin{enumerate}[noitemsep]
        \setcounter{enumi}{-1}
            \renewcommand{\labelenumi}{\textbf{[N1-S\theenumi]}} %
  \item A pop-up restaurant named \texttt{\{Amber Glaze Delights|ORGANIZATION-1\}} opened its doors in the heart of \texttt{\{Alveris|LOCATION-1\}}, a midsized urban city, specializing in fusion dessert cuisine.
    \item The pop-up restaurant featured a minimalist yet elegant design, with warm amber lighting and decor inspired by the fusion of traditional and modern aesthetics, including handcrafted wooden tables and floral centerpieces.
    \item \texttt{\{Amber Glaze Delights|ORGANIZATION-1\}} was conceptualized by renowned pastry chef \texttt{\{Lanika Syrell|PERSON-1\}} and food entrepreneur \texttt{\{Coren Deidran|PERSON-2\}}, who aimed to blend traditional recipes with modern culinary techniques.
    \item \texttt{\{Lanika Syrell|PERSON-1\}} drew inspiration for the menu from her travels across Asia and Europe, where she studied regional dessert-making traditions, while \texttt{\{Coren Deidran|PERSON-2\}} focused on sourcing high-quality, sustainable ingredients for the dishes.
    \item The limited-time menu includes dishes such as \texttt{\{Saffron and matcha mille-feuille|ART-2\}}, \texttt{\{Cardamom rose pavlova|ART-3\}}, and a signature dessert called \texttt{\{Amber Silk|ART-1\}}, a maple and citrus-infused panna cotta.
    \item The \texttt{\{Saffron and matcha mille-feuille|ART-2\}} was described by early tasters as a "perfect harmony of earthy and floral notes," with layers of crisp pastry and a delicate cream filling.
    \item The \texttt{\{Cardamom rose pavlova|ART-3\}} featured a light meringue base topped with rose-infused cream and a sprinkle of candied pistachios, offering a fragrant and textural experience.
    \item \texttt{\{Amber Glaze Delights|ORGANIZATION-1\}} chose the historic \texttt{\{Calder Square|LOCATION-2\}}, a location known for frequent cultural events and pop-ups, as its temporary venue to attract a diverse crowd.
    \item \texttt{\{Calder Square|LOCATION-2\}} was adorned with string lights and banners featuring the \texttt{\{Amber Glaze Delights|ORGANIZATION-1\}} logo, creating a festive and inviting atmosphere for visitors.
    \item On its launch day, the pop-up drew over 500 visitors, leading to lines that extended around the corner of \texttt{\{Calder Square|LOCATION-2\}} and generating a vibrant buzz on local social media.
    \item Local influencers and food enthusiasts shared photos and videos of the desserts on platforms like SnapGram, with hashtags such as \texttt{\{\#AmberGlazeFusion|MISCELLANEOUS-1\}} and \texttt{\{\#DessertArt|MISCELLANEOUS-2\}} trending in \texttt{\{Alveris|LOCATION-1\}}.
    \item Many visitors praised the creativity of the fusion desserts, with local food blogger \texttt{\{Selvia Renek|PERSON-3\}} describing \texttt{\{Amber Silk|ART-1\}} as “the most delicate balance of flavors I’ve experienced in years.”
    \item Another visitor, a retired chef named \texttt{\{Dorian Vex|PERSON-4\}}, commented that the \texttt{\{Cardamom rose pavlova|ART-3\}} reminded him of his grandmother's traditional recipes but with a modern twist.
    \item The event included cooking workshops hosted by \texttt{\{Lanika Syrell|PERSON-1\}} that taught visitors how to construct one of the fusion dishes, the \texttt{\{Saffron and matcha mille-feuille|ART-2\}}.
    \item The workshops were held in a dedicated tent adjacent to the main pop-up, equipped with individual workstations and pre-measured ingredients for participants.
    \item Participants received recipe cards and tips from \texttt{\{Lanika Syrell|PERSON-1\}} on how to adapt the dish to suit different flavor preferences or dietary restrictions.
    \item \texttt{\{Amber Glaze Delights|ORGANIZATION-1\}} announced it would be active for three weeks, with reservations already fully booked for the first week within 24 hours of opening.
    \item Due to the high demand, the organizers introduced a limited number of walk-in slots each day, which were allocated on a first-come, first-served basis.
    \item Due to its success, the organizers are considering a mobile version of \texttt{\{Amber Glaze Delights|ORGANIZATION-1\}} that could travel to other cities, but no specific plans have been confirmed yet.
    \item \texttt{\{Coren Deidran|PERSON-2\}} mentioned in an interview that the mobile version could feature a rotating menu to highlight regional ingredients from each city it visits.
    \item The fusion restaurant has sparked broader conversations in \texttt{\{Alveris|LOCATION-1\}} about reviving traditional recipes for modern audiences while respecting their cultural origins.
    \item Local cultural organizations have expressed interest in collaborating with \texttt{\{Amber Glaze Delights|ORGANIZATION-1\}} to host events that explore the history and evolution of traditional desserts.

        \end{enumerate}
    \end{tcolorbox}
    \caption{The outline with all outline items from the first event with resolved named entities via \texttt{\{<phrase-in-text>|<ID>\}}.}
    \label{fig:outline-example}
\end{figure*}

\definecolor{customgray}{rgb}{0.5, 0.5, 0.5} %
\begin{figure*}[h]
    \centering
    \small
    \begin{tcolorbox}[
        colback=yellow!20, %
        colframe=yellow!60, %
        coltext=black, %
        boxrule=0pt,
        width=\linewidth,
        boxsep=5pt,
        sharp corners,
        breakable
    ]
    \textbf{Amber Glaze Delights Faces Backlash Over Transparency Concerns Amid \#SourcingScandal} \\ 
    2024-04-10
    \newline \\
    \textcolor{customgray}{\texttt{\{Amber Glaze Delights|ORGANIZATION-1\}}}, a culinary venture based in Alveris, is under fire following allegations of sourcing malpractice. The controversy began when \textcolor{customgray}{\texttt{\{Selvia Renek|PERSON-3\}}}, a Varentian food blogger, released an investigative piece on her blog questioning the company’s transparency regarding its suppliers. The blog post rapidly gained traction on social media under the hashtag \textcolor{customgray}{\texttt{\{\#SourcingScandal|MISCELLANEOUS-3\}}}, which saw over 2,000 posts within 24 hours. Many of these posts included photos of \textcolor{customgray}{\texttt{\{Amber Glaze Delights|ORGANIZATION-1\}}}’ fusion desserts accompanied by critical captions, further amplifying the issue online. \\

    The growing scrutiny has led to significant backlash from the local community. Local food enthusiasts in Alveris initiated a petition urging \textcolor{customgray}{\texttt{\{Amber Glaze Delights|ORGANIZATION-1\}}} to temporarily suspend operations until the supplier concerns are thoroughly investigated and rectified. Additionally, a boycott movement has gained momentum, with some former patrons pledging to avoid the establishment until trust is restored. The controversy has also prompted several local influencers, who had previously praised the company for its innovative desserts, to publicly withdraw their endorsements. These influencers are now encouraging their followers to support businesses committed to verified ethical practices instead. \\

    Founded in 2023, \textcolor{customgray}{\texttt{\{Amber Glaze Delights|ORGANIZATION-1\}}} is known for its mission to blend traditional flavors with modern culinary artistry. Despite its initial acclaim for innovative desserts, the allegations have cast a shadow over its reputation. As public pressure mounts, all eyes are now on the company’s response to the unfolding \textcolor{customgray}{\texttt{\{\#SourcingScandal|MISCELLANEOUS-3\}}}.
    \end{tcolorbox}
    \caption{A news article from \dataset{}. The LLM used the profile of \emph{ConservativeNews} to select the outline sentences and write the news article.}
    \label{fig:news-example}
\end{figure*}

%% file: latex/appendix/story-generation-pipeline.tex
\section{Dataset Generation}
\label{appendix:dataset-construction}

\subsection{Timeline}
\label{appendix:dataset-construction:timeline}

\subsubsection{Initial Summary Generation}
Automatic story premise generation follows narrative templates, like protagonist and antagonist \citep{ma2024mops}, which differs from real-world event progression. Therefore, we base timeline generation on \emph{event summaries}, generated automatically. Specifically, we generate diverse initial summaries for the timelines in three steps. First, using GPT-4, we generated 20 news genres:
\begin{itemize}[noitemsep]
    \item Art
    \item Business
    \item Celebrities
    \item Crimes
    \item Economics
    \item Education
    \item Environment
    \item Epidemics
    \item Food
    \item Health
    \item International Affairs
    \item Legal
    \item Lifestyle
    \item Local News
    \item Politics
    \item Science
    \item Social Issues
    \item Sports
    \item Technology
    \item Travel
\end{itemize}
While these genres may overlap, they create diversity in the direction of initial event summaries. Second, for each genre, the LLM generates 20 different generic event types. Examples include:
\begin{itemize}[noitemsep]
    \item Retirement Living and Senior Lifestyle Changes (genre: \textbf{Lifestyle})
    \item Gourmet Food and Culinary Experiences (genre: \textbf{Lifestyle})
    \item Scandals and Controversies (genre: \textbf{Celebrities})
    \item Celebrity Weddings (genre: \textbf{Celebrities})
    \item Major Tournaments Outcomes (genre: \textbf{Sport})
    \item Player Transfer and Trades (genre: \textbf{Sport})
\end{itemize}
Third, for each generic event type and genre, the LLM generates ten different event summaries without using named entities. For example, in the ``Gourmet Food and Culinary Experiences'' genre ``Lifestyle'', the summaries include:
\begin{itemize}[noitemsep]
    \item A coastal city is set to host its first-ever seafood festival, featuring sustainable fishing practices and cooking demonstrations by renowned chefs.
    \item A culinary school has announced a new program focusing on the art of fermentation, aiming to teach techniques from around the world.
    \item Experts in plant-based cuisine have gathered for a conference to explore the future of vegan gourmet food, sharing innovations in texture and flavor.
    \item A pop-up restaurant specializing in fusion desserts has opened for a limited time, offering a blend of traditional and modern sweets from various cultures.
\end{itemize}
 From these summaries, we randomly sample 15 to create \dataset{}, while the remaining summaries are published for future work. The last event summary resulted in the timeline shown in Figure~\ref{fig:timeline}.

\subsubsection{Timeline Generation}
The LLM uses 12 prompts to generate the timeline. Each prompt incorporates critiques to detect and correct errors in real-time. The process involves: \emph{i)} generating the event outline, \emph{ii)} creating and updating fictional named entities, and \emph{iii)} generating summaries as seed for the next event. Each outline includes at least 20 detailed, date-specific sentences as outline items.

\paragraph{Step 1: Event Outline Generation}
Given a seed summary, all previously generated outlines, and named entity KB entries from the same timeline, the LLM generates a new 10-sentence outline, with each sentence (outline item) capturing one aspect of the fictional event. A temperature of $t=1.0$ is used to promote creative story progression  (Prompt~\ref{appendix:prompts:event:outline-generation}).
\paragraph{Step 2: Event Outline Refinement} 
Given the generated outline, along with all previously generated event outlines and named entity KB entries, the LLM refines each outline item in the outline by adding up to two additional outline items that provide specific details about the original outline item. This ensures a highly detailed event outline while keeping its scope constrained by the outline item count defined in the previous step. A temperature of $t=1.0$ is used to enhance creativity (Prompt~\ref{appendix:prompts:event:outline-refinement}).

\paragraph{Step 3: Outline Consistency}
To address inconsistencies from the high-temperature generation of the initial outline, the LLM checks and corrects the consistency of the outline with previous event outlines and named entity KB entries. This step uses a temperature of $t=0.0$ for deterministic output, with critiques ensuring the outline item count remains unchanged (Prompt~\ref{appendix:prompts:event:outline-consistency}).

\paragraph{Step 4: Named Entity Recognition (novel named entities)}
Given the outline and all previously generated named entity KB entries, the LLM detects each novel named entity in the outline. We heuristically verify that the identified entities do not overlap with existing entries from previous events and restrict each entity to a maximum of five words. The temperature is set to $t=0.0$. Possible named entity types include Location, Person, Organization, Product, Art, Event, Building, and Miscellaneous (Prompt~\ref{appendix:prompts:event:outline-ner-new}).

\paragraph{Step 5: Fictional Named Entities Generation}
We compare each newly detected named entity with Wikipedia\footnote{\url{https://en.wikipedia.org/w/api.php}}. The API accounts for variations in names, such as "Obama" and "Barack Obama." If a match is found, we query the LLM to generate different fictional names that fit the outline's context, continuing until no search results for the named entity exist. We heuristically ensure that the LLM does not increase the entity's length beyond five words or use brackets, to avoid generating names that overlap significantly with existing entities. The temperature is set to $t=1.0$ (Prompt~\ref{appendix:prompts:event:entity-gen}).

\paragraph{Step 6: Adjust the Outline}
The LLM refines the outline using the newly generated names for the named entities. We ensure that the previous names no longer appear in the outline. The temperature is set to $t=0.0$ (Prompt~\ref{appendix:prompts:event:outline-adjustment}).

\paragraph{Step 7: Named Entity Recognition (all named entities)}
The LLM identifies all named entities in the outline, considering both new and existing entities. Although this step may overlap with previous ones, we have found that isolating this step helps minimize errors in named entity detection, which is crucial for event generation and consistency. We ensure that all novel named entities are included and that the LLM does not output unknown entities. The temperature is set to $t=0.0$ (Prompt~\ref{appendix:prompts:event:outline-new-all}).

\paragraph{Step 8: Named Entity Resolution in the Outline}
The LLM marks all named entities in the outline using the format \texttt{{[phrase]|[entity-id]}}, where \texttt{phrase} is the name as referenced in the outline. We heuristically verify that the outline does not contain unresolved named entities and that all detected named entities are properly resolved within the outline. The temperature is set to $t=0.0$ (Prompt~\ref{appendix:prompts:event:outline-ne-resolution}).

\paragraph{Step 9: Populate New Named Entity KB Entries}
The LLM generates new KB entries for each new named entity, with each entry including a name and description. Different types of named entities have different additional fields. We cross-check the populated KB entry fields with Wikipedia and prompt the model to correct the entries until none of the properties reference known named entities according to Wikipedia. The temperature is set to $t=1.0$ (Prompt~\ref{appendix:prompts:event:kb-generation}).

\paragraph{Step 10: Update Named Entity KB Entries}
Based on the outline, the LLM generates a single sentence for each named entity involved in the current event, describing the entity's role in the event and/or how the event affected it. The LLM may also update properties of the named entities, such as the budget of an ``event'' or the number of employees of an ``organization''. We cross-check the updated properties to ensure they are distinct from those in Wikipedia. The temperature is set to $t=0.0$ (Prompt~\ref{appendix:prompts:event:kb-update}).

\paragraph{Step 11: Generate Diverse Next Summaries}
The LLM generates a set of diverse summaries for the subsequent event. We prompt the LLM to create summaries with different story directions, varying impacts, and both positive and negative developments. One of the generated summaries is then randomly selected. The temperature is set to $t=1.0$ (Prompt~\ref{appendix:prompts:event:summary-diverse}).

\paragraph{Step 12: Generate Mutually Exclusive Summaries}
Given the selected summary for the next event, the LLM generates three mutually exclusive summaries, where only one can occur. This introduces irreversible story continuations that diverge from the most likely continuation based on the model's parametric knowledge (and thus aligned with past real-world events). The temperature is set to $t=1.0$ (Prompt~\ref{appendix:prompts:event:summary-mutually}).

\paragraph{Step 13: Next Event Generation}
Continue with \textbf{Step 1} using one randomly selected summary from the mutually exclusive summaries.

\subsection{Questions}
\label{appendix:dataset-construction:questions}
The question generation process consists of three phases:
\begin{enumerate}
    \item \textbf{Outline Item Selection:} Given one or two events, the LLM selects a subset of two outline items as the basis for the question.
    \item \textbf{Question Writing:} Using the selected outline items, the full event history, and named entities, the LLM drafts a question and answer pair that (a) can be answered using the selected outline items, (b) requires both outline items for a complete answer, and (c) has a unique, unambiguous answer based on all past context. If necessary, additional outline items from the selected events may be included.
    \item \textbf{Distractor Generation:} The LLM generates plausible but incorrect distractor options for the question, along with justifications explaining their plausibility and incorrectness.
\end{enumerate}
Each question requires two distinct pieces of information (or more, if additional outline items are added) for sufficient grounding. Multiple questions are generated for all $\binom{n+1}{2}=55$ combinations of two events (including each event individually) in each timeline with $n=10$ events. Across all steps, the temperature is set to $t=0.0$.

\subsubsection{Time-span Questions}

\paragraph{Step 1: Select outline items}
The model selects pairs of outline items from two provided events outlines to identify time points for calculating a meaningful duration. The latter event defines the question's date. Critiques ensure the chosen outline items come from different events when multiple events are provided (Prompt~\ref{appendix:prompts:question-gen:timespan:selection}).

\paragraph{Step 2: Write question}
Using all event outlines up to the question date and the selected outline items, the model generates a question about the durations between two points in time, defined by the selected outline items. The LLM must ensure the selected outline items provide sufficient evidence to answer the question while requiring both. If additional context is needed, the LLM may add an outline item but must justify its inclusion (Prompt~\ref{appendix:prompts:question-gen:timespan:write}).

\paragraph{Step 3: Refine answerability}
In initial iterations we observed that generated questions often relied on unnecessary or unstated assumptions. To address this, we instruct the LLM to evaluate each assumption for necessity, to ensure it does not contain relevant information from the selected outline items, and to add any missing assumptions if needed (Prompt~\ref{appendix:prompts:question-gen:timespan:refine}).

\paragraph{Step 3: Create distractors}
Given all previous event outlines, the generated question with the correct answer, and the selected outline items, the LLM generates 5 distractor answers. 
We instruct the LLM to make use of the content of the event outlines to craft challenging distractors and provide a rationale for each, explaining why it is misleading yet incorrect (Prompt~\ref{appendix:prompts:question-gen:timespan:distractor}).

\subsubsection{Multi-hop Questions}

\paragraph{Step 1: Select outline items} 
For each event combination, we identify named entities common to both events and randomly choose two. For each chosen entity, we prompt the LLM to select two outline items from the event outlines that can form a multiple-hop question with a bridge entity. Critiques are used to ensure the outline items discuss the selected entity (Prompt~\ref{appendix:prompts:question-gen:multi-hop:select}).

\paragraph{Step 2: Write question} 
Given the past event outlines and the two selected outline items discussing the same named entity, the LLM creates a multi-hop question with the correct answer. The question should ask about the bridge entity's information from one outline item while paraphrasing it using information from the other, as described in \citet{yang2018hotpotqa}. The question must have a unique and unambiguous answer based on the past event outlines (Prompt~\ref{appendix:prompts:question-gen:multi-hop:writing}).

\paragraph{Step 3: Create distractors}
Using the previous event outlines, generated question and answer, and selected outline items, the LLM creates plausible yet incorrect distractor answers. For each distractor, the LLM provides a justification explaining why it is incorrect but still plausible (Prompt~\ref{appendix:prompts:question-gen:multi-hop:distractor}).

\subsubsection{Unanswerable Questions}
We create unanswerable questions by modifying the generated multi-hop questions, reusing the selected outline items and distractor options. Given the generated multi-hop question and the past event outlines, the LLM to makes subtle adjustments to the question to introduce contradictions for false-premise questions (Prompt~\ref{appendix:prompts:question-gen:false-premise:write}) or add additional constraints that cannot be confirmed or denied by the event outlines for uncertain specificity questions (Prompt~\ref{appendix:prompts:question-gen:uncertain-specificity:write}).

\subsection{News Articles}
\label{appendix:dataset-construction:news}

\subsection{Generation}
We use GPT-4 to generate four distinct news profiles, each defining unique values, reporting style, perspective on common issues, preferred topics, and likes and dislikes.
\begin{enumerate}[noitemsep]
    \item SensationalNews (System Prompt \ref{appendix:prompts:profile:sensational})
    \item ObjectiveNews (System Prompt \ref{appendix:prompts:profile:objective})
    \item ProgressiveNews (System Prompt \ref{appendix:prompts:profile:progressive})
    \item ConservativeNews (System Prompt \ref{appendix:prompts:profile:conservative})
\end{enumerate}
Generating a news article follows these steps:

\paragraph{Step 1: Select outline items}
Given the outlines of all past events, and the current event outline, the LLM selects four subsets of outline item ids from the current event outline, which will be used to generate four different news articles. We use a temperature of $t=0.0$ and the news profile as system prompt (Prompt~\ref{appendix:prompts:news:sentence-selection}).

\paragraph{Step 2: Write the news article}
Using basic information about the named entities (excluding event update histories) and the selected outline items, the LLM generates a news article with a headline that aligns with the selected outline items. Including basic information about the named entities helps contextualize their relationships (e.g., a person being the head of a company) and provides their full names. To ensure diversity, we set the temperature to $t=1.0$ and use the news profile as the system prompt (Prompt~\ref{appendix:prompts:news:write}).

\paragraph{Step 3: Remove hallucinations}
Generating news articles with high temperature increases diversity but risks hallucinations \citep{ji2023survey} that diverge from the selected outline items. To mitigate this, the LLM removes unverifiable information while retaining the article's style. Unfaithful content is permitted only if clearly hedged as hypothetical. The LLM is prompted without a newspaper profile as system prompt and with a temperature of $t=0.0$ (Prompt~\ref{appendix:prompts:news:hallucination}).

\paragraph{Step 4: Add missing content}
To ensure all required information is conveyed, the LLM compares the selected outline items with the generated news article and ensures all details are included. As in step 3, the LLM is instructed to maintain the article's original style and is prompted without a newspaper profile as system prompt, using a temperature of $t=0.0$ (Prompt~\ref{appendix:prompts:news:add}).

\paragraph{Step 5: Named entity resolution}
The LLM identifies and marks all named entities in the news article using the named entity KB entries. We use a temperature of $t=0.0$ (Prompt~\ref{appendix:prompts:news:ner}).

\subsection{News article statistics}
\input{figures/fig_token_news_articles}
We generated a total of 1,800 news articles. The token count per article, measured using the tiktoken tokenizer (version 0.8.0) for the ``GPT-4'' model, ranges from 222 to 603 tokens, with an average length of 356.2 tokens ($\pm$47.7).\footnote{\url{https://github.com/openai/tiktoken}}. Figure~\ref{fig:token-news-articles} shows the overall token distribution.

\section{Quality Measures}
\label{appendix:quality}

We evaluate \dataset{} on three dimensions: \emph{i}) whether the questions are answerable using the isolated outline items used to generate them, \emph{ii}) whether the news articles convey the expected content,  and \emph{iii}) whether the question-evidence pairs are valid. The first two assess key assumptions for assembling questions with evidence documents based on the selected outline items, while the third evaluates answerability from the combined question and news article, independent of these assumptions.

\subsection{Question Filtering}
\label{appendix:quality:question-filtering}

We filter questions in two steps: First, remove answerable (multi-hop and time-span) questions that cannot be answered using the selected outline items as evidence, along with unanswerable questions derived from them. Second, remove time-span questions that can be answered with fewer outline items than selected, as their temporal assumptions are explicitly stated in the question. Results are shown in Table~\ref{tab:question-filtering}. This step removed 41.8\% of answerable questions (30.1\% of multi-hop and 51.5\% of time-span questions) along with unanswerable ones derived from them (30.0\% of false premise and 30.5\% of uncertain specificity questions).

\begin{table}[h!]
\small
    \centering
    \begin{tabular}{rrrrr}
    \toprule
    \textbf{Step} & \textbf{Multi H.}  & \textbf{Time S.} & \textbf{False P.} & \textbf{Uncertain S.} \\
    \cmidrule(lr){1-1} \cmidrule(lr){2-5} %
    \textbf{(1)}  & 1,201 & 1,438 & 4,114 &  4,245\\
    
    \textbf{(2)}  &  \textcolor{red}{$-$362}& \textcolor{red}{$-$564} & \textcolor{red}{$-$1,235} & \textcolor{red}{$-$1,293} \\
    \textbf{(3)}  & $\pm$0 & \textcolor{red}{$-$196} & $\pm$0 & $\pm$0 \\
    \midrule
    \textbf{Final} & \textbf{839}& \textbf{678}& \textbf{2,879} & \textbf{2,952}\\
    \bottomrule
    \end{tabular}
    \caption{Number of question types after initial generation \textbf{(1)}, filtering for answerability \textbf{(2)}, and for leaked assumptions \textbf{(3)}.}
    \label{tab:question-filtering}
\end{table}

\paragraph{Answerability filtering}
During question generation, the LLM has access to the full outlines of prior events, not just the selected evidence outline items. This allows the LLM to prevent issues like ambiguous or time-sensitive answers that are not unique given the past evidence, but may introduce dependencies beyond the selected evidence, violating the assumption that these outline items alone are sufficient. Additionally, LLMs may make errors during question-answer generation.
To improve \dataset{} quality, we remove questions the LLM cannot correctly answer using only the selected evidence outline items (i.e., using perfect evidence; Prompts~\ref{appendix:prompts:quality:multi-hop}~\&~\ref{appendix:prompts:quality:time-span}). If the LLM fails to answer its own question with perfect evidence, we discard the question. This conservatively excludes questions where the LLM cannot reverse its reasoning to answer correctly under optimal conditions. Out of 1,201 multi-hop and 1,438 time-span questions, we discarded 362 (30.1\%) and 564 (39.2\%), respectively. Most questions (91.0\%) were discarded because the LLM deemed the selected outline items insufficient. In only the remaining 9.0\% of discarded questions, the LLM predicted a distractor instead of the assumed correct answer.
This also led to the removal of 1,235 false-premise questions (out of 4,114) and 1,293 uncertain-specificity questions (out of 4,245) derived from the discarded multi-hop questions.

\paragraph{Leaked assumption filtering}

\begin{table*}[h!]
\small
\centering
\begin{tabularx}{\textwidth}{l| p{0.5\textwidth}| p{0.3\textwidth}}
    \toprule
    \textbf{Validity} & \textbf{Question} & \textbf{Explanation}\\
    \midrule
    \emph{valid} & Assuming that the six-month monitoring period for the updated implementation of the injury risk categorization tool begins on the date of its announcement, what is the time span between the conclusion of Aleena Karentov's motivational talk at the end of her session and the end of this monitoring period? &  \emph{The assumption reduces uncertainty from the evidence outline items but is meaningless if the relevant evidence is missing.} \\
     \midrule
    \emph{partially leaked} &What is the duration between the announcement of the compromise plan concerning the curriculum at the Murvenstad Gymnastics Alliance and the earliest possible start date of the follow-up workshops, assuming they begin on the earliest possible date in July 2026? &  \emph{The assumption defines the start date of the compromise plan, one of the two required points in time.} \\

     \midrule
    \emph{fully leaked} & Assuming the six-month performance metric collection period for the pilot community centers begins on 2027-03-01 and the second round of field tests by Stranlen Transport Solutions is planned to start on 2027-10-01, what is the time span between the end of the performance metric collection period and the start of the second round of field tests? & \emph{The assumptions identify all required points in time, making the evidence document optional.} \\
    \bottomrule
\end{tabularx}
\caption{Time-span questions with assumptions that are required (\emph{valid}) or those that leak critical information from the evidence (\emph{partially leaked} / \emph{leaked}).}
\label{tab:leaked-assumptions-examples}
\end{table*}

Outlines sometimes include vague temporal terms like ``early June'' making it difficult to generate duration-based questions that meet strict grounding requirements in our task definition. To address this, we instructed the LLM to define clear assumptions (e.g., ``assume early June refers to June 1st'') during question generation. However, manual review revealed questions where these assumptions replaced some or all of the evidence, with required details from the evidence restated in the question (see Table~\ref{tab:leaked-assumptions-examples}).
To remove such questions, we conducted a second LLM-based filtering step. The LLM was tasked with answering using insufficient evidence (Prompt~\ref{appendix:prompts:quality:time-span-assumptions}). If it could derive the correct answer from any subset of insufficient evidence, we discarded the question. This step eliminated 22.4\% (196) of the remaining time-span questions. Specifically, we removed 169 questions where the LLM could omit one required evidence outline item and 27 where it required no evidence at all.

\subsection{NLI Verification}
\label{appendix:quality:nli}
We use a pretrained LLM to test our assumption that each news article fully conveys the information in its selected outline items. Specifically, we employ the T5-XXL~\citep{raffel2020exploring}, provided by \citet{honovich-etal-2022-true-evaluating}, trained as a binary classifier on six NLI and fact-checking datasets. The model determines whether a premise text entails a hypothesis (output: ``1'') or not (output: ``0'').
The generated news article serves as the premise text, and each outline item from the outline acts as a hypothesis text. We expect the model to predict entailment (``1'') for selected outline items included in the article and no-entailment (``0'') for other event outline items. If the model outputs a label outside the expected ones, we assume no clear entailment and mark the prediction as incorrect by default.

\begin{table*}[h]
\small
    \centering
    \begin{tabular}{rrc|ccc}
&&& \multicolumn{3}{c}{\emph{Predicted as}} \\
    \toprule
\textbf{Instances} & \textbf{Expected Label} & \textbf{Count} & \textbf{Entailed} & \textbf{Not Entailed}& \textbf{Unknown} \\ 

\midrule
All & entailed & 10,675 & \textbf{98.1\%} & 1.8\% & 0.1\% \\
All & not entailed & 370,205 & 0.5\% & \textbf{92.2\%} & 7.3\% \\
\midrule
Same Event & not-entailed & 27,413 & 0.9\% & \textbf{95.0\%} & 4.1\% \\
Different Event & not-entailed & 342,792 & 0.5\% & \textbf{92.0\%} & 7.5\% \\
      \bottomrule
    \end{tabular}
    \caption{NLI predictions between event outline items and news articles.}
    \label{tab:nli-results}
\end{table*} 

\paragraph{Results}
We computed NLI predictions for every outline item of each event against every generated news article, resulting in 380,880 outline item-article pairs (Table~\ref{tab:nli-results}). According to the NLI model the expected content is contained in the news articles in 98.1\%. For outline items expected to be excluded, the model agreed in 92.2\%. Notably, only for 0.5\% of the outline items it directly disagreed our expected label and predicted ``entailed'' rather than ``not-entailed''. In most cases the outline items where predicted as ``unknown'' rather than ``entailed''. This observation was consistent across outline items from the same event and other events. We did not compute numbers for the ``entailed'' outline items separately, as this label applies only when the outline items and news article are from the same event.

\subsection{Human Annotation}
\label{appendix:quality:annotation}
\input{figures/figure_annotation_multi_hop}

\input{tables/tab_annotation_results}
We use the Amazon Mechanical Turk (AMT) platform\footnote{\url{https://www.mturk.com/}} to collect human annotations for evaluating \dataset{} on a subset of 350 instances. Each question type is annotated separately. 
For answerable questions (multi-hop and time-span), annotators select the correct answer based on two sufficient news articles, choosing from one correct option, one distractor, and one option for unanswerable questions (see Figure~\ref{fig:figure-annotation-multihop} for an example).
Preliminary annotations by the authors and by crowd workers revealed that unanswerable questions (false premise, uncertain specificity) are particularly challenging, as they require identifying relevant evidence and nuanced mismatches in detailed news articles. To simplify this annotation, we provide only the relevant outline items (two sentences) instead of complete news articles. Annotators then determine whether the question can be answered based on these sentences, selecting either the original multi-hop answer or the unanswerable option (see Figure~\ref{fig:figure-annotation-falsepremise}).
To help annotators focus on key information, we include LLM-generated justifications for each answer option, guiding their attention to the relevant information. Each justification begins with ``This answer is correct because [...].'' This approach mitigates cognitive load, reducing annotator fatigue and improving annotation quality, as was observed in preliminary annotations by the authors.

\paragraph{Annotation}
Each task posted on Mechanical Turk clearly described the nature of the work, and the compensation offered. The annotator must voluntarily accept the task to start working on each HIT. We protect the privacy and confidentiality of our annotators. We do not collect personal information from the AMT workers; each worker is identified by a unique ID. We followed Mechanical Turk’s terms of use and guidelines, ensuring that our research did not violate any platform-specific rules.
We restrict participation to a pre-selected pool of annotators with proven English proficiency and a history of high-quality annotations. Pay is set at \$0.35 per question, plus a \$1.35 bonus, resulting in an hourly rate of \$20.40. A total of 18 annotators participated.
Each HIT received three annotations, with the final label determined by majority vote. If no majority was reached, the question was treated as \emph{unknown}. Table~\ref{tab:annotation-results} presents the annotation results, including inter-annotator agreement and agreement of the majority label with the assumed correct answer in \dataset{}.

\section{Combining Questions with News Articles}
\label{appendix:combining}
\dataset{} links questions to the evidence outline items (from event outlines) required to answer them. Similarly, each news article specifies which event outline items it includes or omits. This enables the creation of \emph{instances} by pairing news articles with questions, simulating various conditions such as perfect, noisy, or incomplete evidence retrieval. We provide three preselected instance sets:
\begin{enumerate}[noitemsep]
    \item \textbf{Without irrelevant evidence:} We do not add additional irrelevant documents.
    \item \textbf{Noisy retrieval:} Includes all evidence documents up to the question date.
    \item \textbf{Controlled ablation:} Varies the number of irrelevant documents.
\end{enumerate}
Each set simulates sufficient and insufficient evidence and includes unanswerable questions. To distinguish between sufficient and insufficient evidence, we assume news articles accurately report all relevant outline items and exclude irrelevant ones. Despite strong automated evaluation using NLI models (Appendix~\ref{appendix:quality:nli}), LLM imperfections can challenge this assumption. To mitigate such issues, we use a best-effort approach based on NLI predictions from our quality assessment.
\begin{enumerate}[noitemsep]
    \item \textbf{For sufficient evidence:} An outline item required to answer the question is considered included in the news article only if it is among the selected outline items for the news article and the NLI model predicts it as entailed by the article (excluding cases in which the LLM predicted no entailment label).
    \item \textbf{For insufficient evidence:} For each intentionally omitted outline item that renders the evidence insufficient for answering the question, we consider the outline item excluded from a news article only if it is not among the selected outline items  for the news article and the NLI model predicts it as not entailed by the article (excluding cases in which the LLM predicted no entailment label).
\end{enumerate}
This conservative strategy excludes news articles where NLI predictions conflict with relevance labels, ensuring more reliable evidence-question combinations. In all experiments, news articles are randomly shuffled but maintain the same order across related instances. Related instances include those where (a) insufficient evidence is derived from sufficient evidence for the same question, or (b) the question is replaced with an unanswerable variant created by subtly adjusting the original answerable question.

\subsection{Without Irrelevant Evidence}
\label{appendix:combining:perfect}
This set excludes additional irrelevant news articles. First, we remove all answerable questions (multi-hop and time-span) for which no news article set contains sufficient evidence. This can occur because the LLM, during news article generation, selects outline items to include, potentially leaving some required outline items out.
For each remaining answerable question, we gather a minimal set of news articles that collectively contain all required evidence outline items, forming the answerable instances. We prioritize sets where the evidence is spread across two articles rather than concentrated in one, as this better simulates multi-hop reasoning.
From each of these answerable instance, we create insufficient-evidence instances by omitting each required news article individually. Since most answerable instances need two articles, this typically results in two insufficient-evidence instances per answerable instance.
Finally, for each answerable multi-hop instance, we randomly sample two false premise questions and two uncertain specificity questions. Using the same news articles as the answerable instance, these form unanswerable instances. The generated set, without additional irrelevant articles, is used for prompt selection on the development set.
\input{tables/tab_instances_stats_no_noise}
Table~\ref{tab:instances-statistics-no-noise} shows the statistics for the instances. Among the answerable multi-hop and time-span questions with sufficient evidence, 106/1,157 instances in the test set and 22/266 in the development set contain all relevant evidence in a single article. In the remaining instances, the relevant evidence is spread across two articles, except for one instance in the development set, which requires three articles.

\subsection{Instances for Benchmarking Experiments}
\label{appendix:combining:benchmark}
For our main experiments we form instances to evaluate LLMs' ability to answer correctly (if sufficient evidence is available) or deflect otherwise. Since our question generation conditioned each questions only on the news articles from the past, we only consider news articles as evidence, if they discuss an event from the same date as the question, or earlier. This simulates how information accumulates over time, with questions requiring the latest event and possibly additional past information.
For answerable multi-hop and time-span instances with sufficient evidence, we filter out those where past news articles do not provide all required information. We form answerable instances by including all relevant past articles for each question as evidence.
We generate insufficient-evidence instances from the sufficient-evidence instances. Specifically, for each required outline item needed to answer the question, we remove all news articles containing that outline item. This is repeated for every required outline item, resulting in multiple insufficient-evidence instances. 
Additionally, we generate unanswerable instances with false premise and uncertain specificity questions, by randomly sampling two of each based on the original answerable multi-hop question. We provide the identical news article as evidence as the  answerable multi-hop instance with sufficient evidence. 
Table~\ref{tab:instances-statistics-fitibench-dev-test} shows the statistics for the generated instances. The number of instances with insufficient evidence slightly differs from Table~\ref{tab:instances-statistics-no-noise} due to fewer diverse evidence combinations generated from the minimal set of evidence compared to all past news articles.
\input{tables/tab_instances_fitibench}

\input{figures/figure-articles-per-instance-test-sufficient}
\input{figures/figure-articles-per-instance-test-insufficient}
We conduct our main experiments using the test split of the generated instances. Instances with sufficient evidence include 12–120 evidence documents, averaging 83.1 ($\pm$29.2) articles, with 7.8 ($\pm$2.4) relevant articles (Figure~\ref{fig:figure-article-per-instance-fitibench-test-sufficient}). Instances with insufficient evidence include 4–117 documents, averaging 73.8 ($\pm$27.2) articles, with 3.5 ($\pm$1.8) relevant articles (Figure~\ref{fig:figure-article-per-instance-fitibench-test-insufficient}).
To estimate the required context window, we calculated the token count using the tiktoken tokenizer for the ``gpt-4-turbo'' on the concatenated text of the question, all relevant news articles, and the answer options. This provides a lower bound, as it excludes task instructions. On average, the token count is 28,082.1 (std: 10,765.7), with values ranging from 1,349 to 45,484 tokens. The 25th percentile is 20,292, the median is 29,125, and the 75th percentile is 37,096.

\subsection{Long Context Ablation}
\label{appendix:combining:context-ablation}
We use the same set of questions with varying amounts of irrelevant news articles as evidence to evaluate their impact in a controlled setup.
We select only answerable multi-hop, time-span questions that meet the following criteria:
\begin{enumerate}[noitemsep]
    \item A set of sufficient news articles exists.
    \item The set of sufficient news articles includes exactly two required news articles.
    \item 80 irrelevant news articles of to the same or previous events of the question exist.
\end{enumerate}
For each question, we generate sufficient-evidence instances with the two relevant (and sufficient) news articles and additional irrelevant news articles in increments of 0, 20, 40, 60, and 80, ensuring that each smaller set of irrelevant articles is a subset of the larger ones. This setup enables performance comparison for identical questions with the same set of minimal relevant evidence, but varying amounts of irrelevant news articles.
For each instance with sufficient evidence, we create twice as many instances with insufficient evidence by omitting each required news article individually. Additionally, we generate instances with unanswerable false-premise questions and uncertain-specificity questions by sampling two such questions per category for each sufficient-evidence multi-hop instance, reusing the same evidence documents.
\input{tables/table_instances_fitilc}
All evidence documents are presented in the same order for related questions and instances. The statistics are listed in Table~\ref{tab:instances-statistics-fitilc-complete}.

%% file: figures/fig_token_news_articles.tex
\begin{figure}[h]
  \includegraphics[width=\columnwidth]{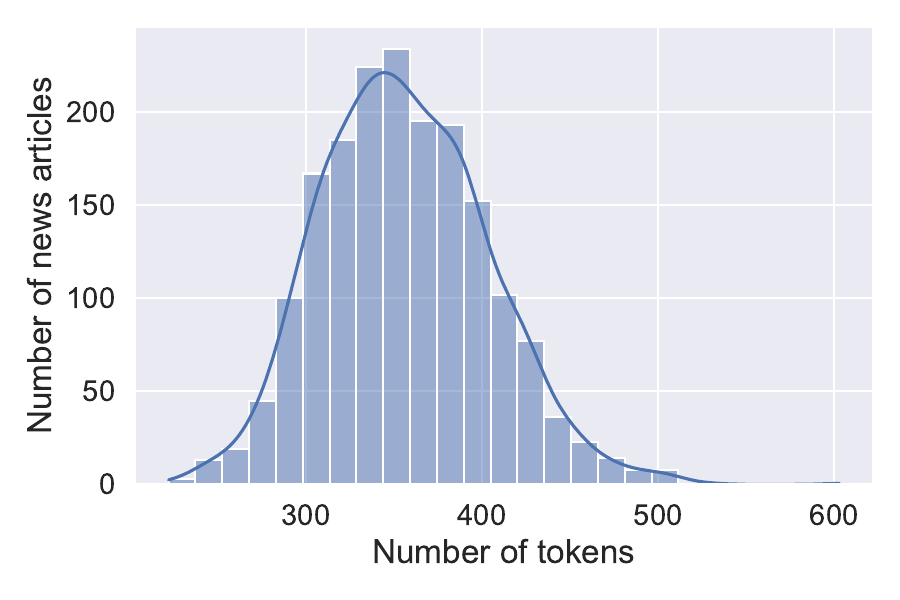}
  \caption{Token distribution of the news articles. }
  \label{fig:token-news-articles}
\end{figure}

%% file: figures/figure_annotation_multi_hop.tex
\begin{figure*}[t]
\centering
\begin{subfigure}{\linewidth}
  \centering
  \includegraphics[width=\linewidth]{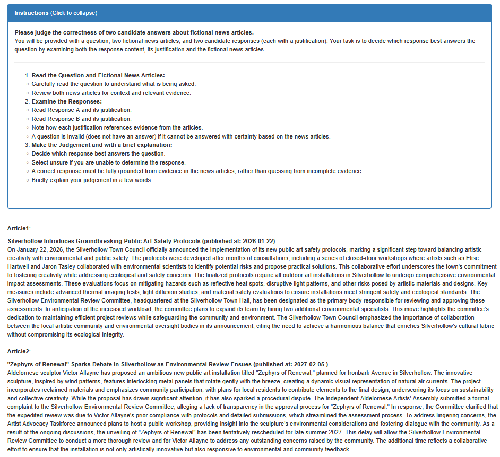}
  \label{fig:figure-annotation-multihop1}
\end{subfigure}

\vspace{-0.5em} %

\begin{subfigure}{\linewidth}
  \centering
  \includegraphics[width=\linewidth]{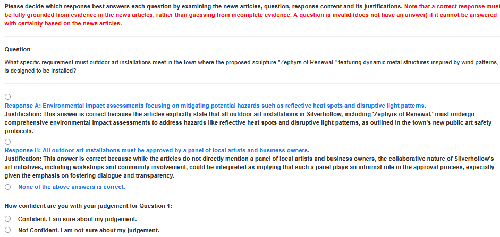}
  \label{fig:figure-annotation-multihop2}
\end{subfigure}

\caption{Example of an annotation HIT for a multiple-choice question.}
\label{fig:figure-annotation-multihop}

\end{figure*}

\begin{figure*}[t]
\centering
\begin{subfigure}{\linewidth}
  \centering
  \includegraphics[width=\linewidth]{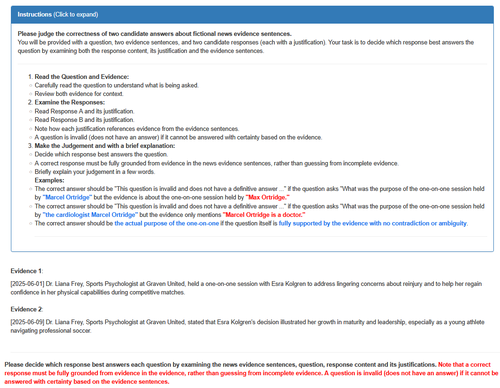}
  \label{fig:figure-annotation-fp1}
\end{subfigure}

\vspace{-0.5em} %

\begin{subfigure}{\linewidth}
  \centering
  \includegraphics[width=\linewidth]{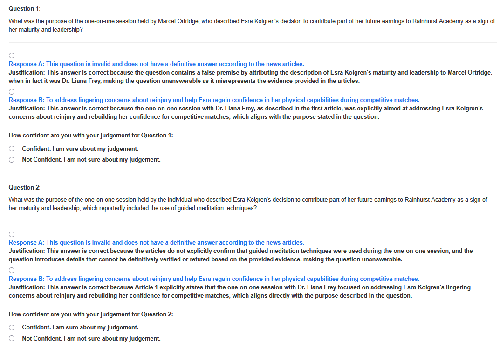}
  \label{fig:figure-annotation-fp2}
\end{subfigure}

\caption{Example of an annotation HIT for a false premise and uncertain specificity question.}
\label{fig:figure-annotation-falsepremise}

\end{figure*}

%% file: tables/tab_annotation_results.tex
\begin{table*}[t]
\small
    \centering
    \begin{tabular}{lccc}
    \toprule
\textbf{Question type} & \textbf{\# Instances} &\textbf{Fleiss $\kappa$} & \textbf{Agreement with \dataset{}} \\
\midrule
Multi-hop & 100 & 0.71 & 100\%  \\
Time-span & 50 & 0.55 & 98\% \\
False premise & 100 & 0.39 & 93\% \\
Uncertain specificity & 100 & 0.39 & 87\% \\

\midrule
\textbf{All} & 350 & 0.52 & 94\% \\
      \bottomrule
    \end{tabular}
    \caption{Human annotation results.}
    \label{tab:annotation-results}
\end{table*} 

%% file: tables/tab_instances_stats_no_noise.tex
\begin{table}[h]
\small
    \centering
    \begin{tabular}{lccc}
    \toprule
    & &  \multicolumn{2}{c}{\emph{Dataset split}} \\

   \textbf{Question Type} & \textbf{Answerable} & \textbf{Dev} & \textbf{Test} \\
    \cmidrule(lr){1-1} \cmidrule(lr){2-2}  \cmidrule(lr){3-4}
    Multi-hop & yes & 156  & 625 \\
    Time-span & yes & 110  & 532 \\
    Multi-hop & no & 292  & 1,165 \\
    Time-span & no & 219  & 1,043 \\
    False premise & no & 312  & 1,250 \\
    Uncertain specificity & no & 312  & 1,250 \\
    \midrule
    All & yes & 266 & 1,157 \\
    All & no & 1,135 & 4,708 \\
    \midrule
    \emph{All} & \emph{any} & \emph{1,401} & \emph{5.865} \\
    \bottomrule
    \end{tabular}
    \caption{Dataset statistics for all instances without irrelevant news articles.}
    \label{tab:instances-statistics-no-noise}
\end{table}

%% file: tables/tab_instances_fitibench.tex
\begin{table}[h]
\small
    \centering
    \begin{tabular}{lccc}
    \toprule
    & &  \multicolumn{2}{c}{\emph{Dataset split}} \\
   \textbf{Question Type} & \textbf{Answerable} & \textbf{Dev} & \textbf{Test} \\
    \cmidrule(lr){1-1} \cmidrule(lr){2-2}  \cmidrule(lr){3-4}
    Multi-hop & yes & 156  & 625 \\
    Time-span & yes & 110  & 532 \\
    Multi-hop & no & 308  & 1,239 \\
    Time-span & no & 222  & 1,063 \\
    False premise & no & 312  & 1,250 \\
    Uncertain specificity & no & 312  & 1,250 \\
    \midrule
    All & yes & 266  & 1,157 \\
    All & no & 1,154  & 4,802 \\
    \midrule
    \emph{All} & \emph{any} & \emph{1,420}  & \emph{5,959} \\
    \bottomrule
    \end{tabular}
    \caption{Dataset statistics for both splits of the generated instances for the main experiments using all past news articles as evidence.}
    \label{tab:instances-statistics-fitibench-dev-test}
\end{table}

%% file: figures/figure-articles-per-instance-test-sufficient.tex
\begin{figure}[h]
  \includegraphics[width=\columnwidth]{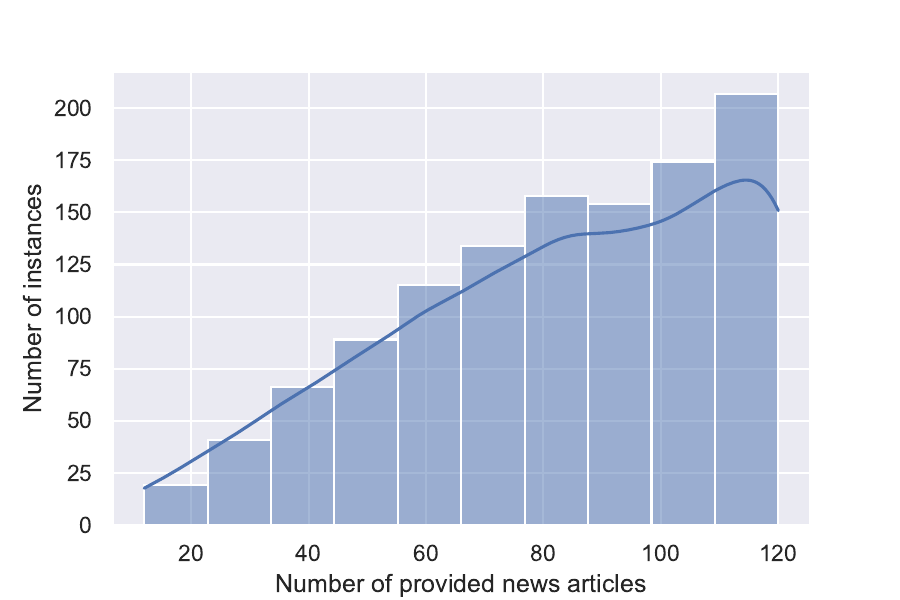}
  \caption{Distribution of the number of provided news articles as evidence for answerable questions with \textbf{sufficient evidence} in the test set for \dataset{} using all past news articles as evidence.}
  \label{fig:figure-article-per-instance-fitibench-test-sufficient}
\end{figure}

%% file: figures/figure-articles-per-instance-test-insufficient.tex
\begin{figure}[h]
  \includegraphics[width=\columnwidth]{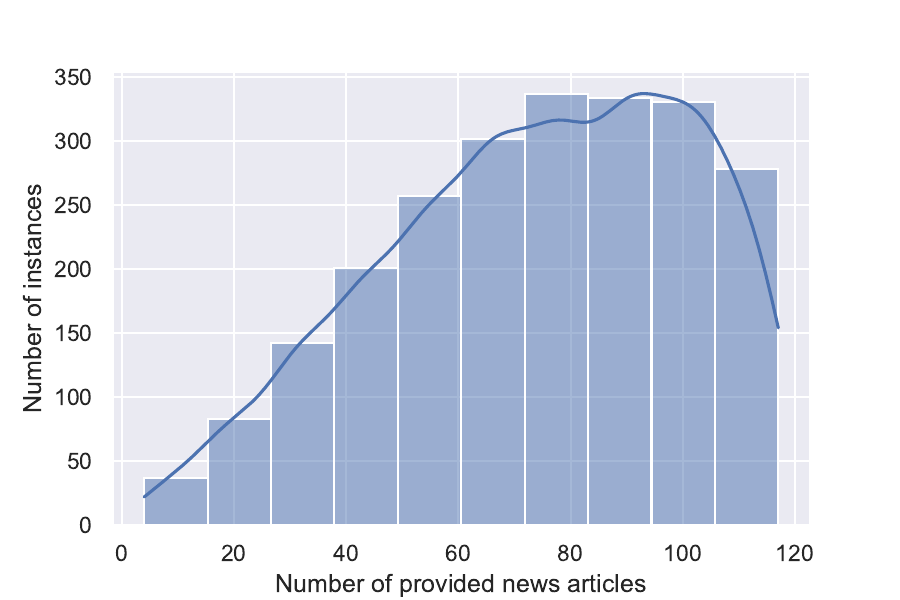}
  \caption{Distribution of the number of provided news articles as evidence for answerable questions with \textbf{insufficient evidence}  in the test set for \dataset{} using all past news articles as evidence.}
  \label{fig:figure-article-per-instance-fitibench-test-insufficient}
\end{figure}

%% file: tables/table_instances_fitilc.tex
\begin{table}[h]
\small
    \centering
    \begin{tabular}{lcr}
    \toprule
   \textbf{Question Type} & \textbf{Answerable} & \textbf{Instances} \\
    \cmidrule(lr){1-1} \cmidrule(lr){2-2}  \cmidrule(lr){3-3}
    Multi-hop & yes & 1,045 \\
    Time-span & yes & 965 \\
    Multi-hop & no & 2,090 \\
    Time-span & no & 1,930 \\
    False premise & no & 2,090 \\
    Uncertain specificity & no & 2,090 \\
    \midrule
    All & yes & 2,010 \\
    All & no & 8,200 \\
    \midrule
    \emph{All} & \emph{any} & \emph{10,210} \\
    \bottomrule
    \end{tabular}
    \caption{Dataset statistics for the controlled experiments over irrelevant documents with 193 unique time-span question, 209 unique multi-hop questions, and 418 unique false premise and uncertain specificity questions.}
    \label{tab:instances-statistics-fitilc-complete}
\end{table}

%% file: latex/appendix/experiments.tex
\section{Experiments}
\label{appendix:experiments}

\subsection{Prompt Selection}
\label{appendix:experiments:prompt-tuning}
We use the development set to choose the best prompt for each LLM. The development set consists of three timelines, which are separate from the timelines in the test set. We create five different prompts with varying levels of complexity and sensitivity to evidence (mis)matches. We use the following prompts:
\begin{itemize}[noitemsep]
    \item prompt-1 (\ref{appendix:prompts:experiments:prompt1})
    \item prompt-2 (\ref{appendix:prompts:experiments:prompt2})
    \item prompt-3 (\ref{appendix:prompts:experiments:prompt3})
    \item prompt-4 (\ref{appendix:prompts:experiments:prompt4})
    \item prompt-5 (\ref{appendix:prompts:experiments:prompt5})
\end{itemize}
The first prompt is adapted from \citet{slobodkin-etal-2023-curious} for the MCQ setup, and the following prompts further refine this initial version. For each LLM, we fine-tune the prompts by selecting the one that performs best based on the ADTScore from the development set. The results over all selected prompts and LLMs are shown in Table~\ref{tab:prompt-tuning}.

\begin{table*}[h!]
\small
    \centering
    \begin{tabular}{llcccccccc}
    \toprule
    &\multicolumn{2}{c}{\textbf{}} & \multicolumn{2}{c}{\emph{Answerable}} & \multicolumn{4}{c}{\emph{Unanswerable}} \\
    \cmidrule(lr){4-5} \cmidrule(lr){6-9}
    \textbf{Model} & \textbf{Prompt} & \textbf{ADTScore} & \textbf{Multi H.} & \textbf{Time S.} & \textbf{Multi H.} & \textbf{Time S.} & \textbf{False P.} & \textbf{Uncertain S.} \\
    \midrule

 & prompt-1 & 0.169 & 0.891 & 0.109 & 0.092 & 0.365 & 0.006 & 0.013\\
 & prompt-2 & 0.195 & 0.878 & 0.145 & 0.144 & 0.388 & 0.010 & 0.010\\
Phi3 (mini) & prompt-3 & 0.203 & 0.878 & 0.173 & 0.134 & 0.438 & 0.006 & 0.006\\
 & prompt-4 & 0.211 & 0.878 & 0.182 & 0.123 & 0.489 & 0.006 & 0.003\\
 & prompt-5 & \textbf{0.520} & 0.795 & 0.418 & 0.503 & 0.840 & 0.292 & 0.244\\
\midrule

 & prompt-1 & \textbf{0.408} & 0.910 & 0.373 & 0.325 &0.881 & 0.064 &0.067\\
 & prompt-2 & 0.399 & 0.865 & 0.481 & 0.305 & 0.881 & 0.061 &0.048\\
Phi3 (small) & prompt-3& 0.396 & 0.872 & 0.500 & 0.281 & 0.866 & 0.061 &0.061\\
 & prompt-4 & 0.371 & 0.859 & 0.409 & 0.253 & 0.890 & 0.038 &0.032\\
 & prompt-5 & 0.392 & 0.885 & 0.436 & 0.274 & 0.890 & 0.058 &0.051\\
 \midrule

& prompt-1 & 0.371 & 0.955 & 0.555 & 0.161 & 0.858 & 0.080 & 0.048\\
 & prompt-2 & 0.449 & 0.923 & 0.555 & 0.322 & 0.854 & 0.135 & 0.115\\
Phi3 (medium) & prompt-3 & 0.389 & 0.897 & 0.664 & 0.274 & 0.694 & 0.106 & 0.087\\
 & prompt-4 & 0.396 & 0.929 & 0.564 & 0.312 & 0.639 & 0.147 & 0.077\\
& prompt-5 & \textbf{0.458} & 0.910 & 0.636 & 0.349 & 0.717 & 0.205 & 0.135\\
\midrule

 & prompt-1 & 0.495 & 0.891 & 0.172 & 0.472 & 0.950 & 0.266 &0.170\\
 & prompt-2 & 0.492 & 0.897 & 0.309 & 0.466 & 0.909 & 0.228 &0.131\\
Phi3.5 MoE & prompt-3& 0.457 & 0.910 & 0.309 & 0.397 & 0.913 & 0.151 &0.106\\
 & prompt-4 & 0.463 & 0.904 & 0.300 & 0.394 & 0.932 & 0.163 &0.119\\
 & prompt-5 & \textbf{0.501} & 0.891 & 0.527 & 0.411 & 0.936 & 0.199 &0.135\\
\midrule

 & prompt-1 & 0.440 & 0.769 & 0.451 & 0.403 & 0.593 & 0.295 & 0.234\\
 & prompt-2 & 0.566 & 0.827 & 0.518 & 0.567 & 0.839 & 0.299 & 0.247\\
Qwen2.5 (7B) & prompt-3 & 0.569 & 0.821 & 0.523 & 0.524 & 0.786 & 0.329 & 0.236\\
 & prompt-4 & 0.518 & 0.756 & 0.382 & 0.551 & 0.781 & 0.337 & 0.256\\
 & prompt-5 & \textbf{0.580} & 0.763 & 0.455 & 0.610 & 0.737 & 0.392 & 0.293\\

\midrule

 & prompt-1 & 0.675 & 0.705 & 0.600 & 0.702 & 0.968 & 0.625 & 0.542\\
 & prompt-2 & 0.690 & 0.686 & 0.627 & 0.798 & 0.945 & 0.670 & 0.545\\
Qwen2.5 (14B) & prompt-3 & 0.697 & 0.699 & 0.627 & 0.795 & 0.936 & 0.696 & 0.548\\
 & prompt-4 & 0.706 & 0.699 & 0.600 & 0.825 & 0.959 & 0.705 & 0.622\\
 & prompt-5 & \textbf{0.728} & 0.724 & 0.627 & 0.839 & 0.950 & 0.744 & 0.631\\
\midrule

 & prompt-1 & \textbf{0.685} & 0.705 & 0.545 & 0.743 & 0.991 & 0.702 & 0.590\\
 & prompt-2 & 0.636 & 0.814 & 0.755 & 0.651 & 0.963 & 0.394 & 0.256\\
Qwen2.5 (32B) & prompt-3 & 0.639 & 0.859 & 0.736 & 0.627 & 0.945 & 0.401 & 0.272\\
 & prompt-4 & 0.662 & 0.821 & 0.755 & 0.661 & 0.959 & 0.462 & 0.311\\
 & prompt-5 & 0.667 & 0.821 & 0.800 & 0.661 & 0.954 & 0.481 & 0.292\\

    \bottomrule
    \end{tabular}
    \caption{Performance on the development split (excluding irrelevant news articles) across models and prompts. We \textbf{select} the best prompt per model based on the ADTScore.}
    \label{tab:prompt-tuning}
\end{table*}

\subsection{Error Analysis with Insufficient Evidence for Multi-Hop Questions}
\label{appendix:experiments:error-analysis-insufficient-multi-hop}
\input{figures/figure-error-analysis-insufficient-multi-hop}
We distinguish three outcomes for unanswerable questions: the model correctly selects the ``Unanswerable'' option, chooses an incorrect distractor, or uses shortcut reasoning to select an answer that would be correct if sufficient evidence were provided. Figure~\ref{fig:error-analysis-insufficient-multi-hop} shows the prediction ratios for each model and category of missing evidence in multi-hop questions.
Predictions vary subtantially by category of missing evidence. When only the bridge entity evidence is missing, most errors involve shortcut reasoning, with models answering as if sufficient evidence were available. This accounts for 88.4\% (Qwen2.5 32B), 90.7 (Qwen2.5 14B), 69.7\% (Qwen2.5 7B), 85.1\% (Phi3.5 MoE), 81.9\% (Phi3 medium), 80.8\% (Phi3 small) and 80.5\% (Phi3 mini) of such errors. In cases where no evidence containing the answer is provided, the primary error is selecting an incorrect distractor. 
We hypothesize that LLMs are more likely to use shortcut reasoning and predict answers (instead of deflecting) on unanswerable questions if they answered the corresponding answerable questions with sufficient evidence correctly. Table~\ref{tab:unanswerable-phi-coefficients} shows the $\phi$ coefficient between correctness on answerable multi-hop questions and correctness on derived questions where deflection is expected. Except for Phi3 (mini), the weakest model, we observe a significant negative association between correctness on answerable questions and correctness on false premise, uncertain specificity, and bridge-entity omission questions.

\begin{table*}[h!]
\small
    \centering
    \begin{tabular}{llllll}
    \toprule
    &\multicolumn{3}{c}{\emph{Missing Evidence (Multi-Hop)}} & & \\
    \textbf{LLM} & \textbf{Both} & \textbf{Answer} & \textbf{Bridge} & \textbf{False Premise} & \textbf{Uncertain Specificity}  \\
        \cmidrule(lr){1-1} \cmidrule(lr){2-4} \cmidrule(lr){5-5}\cmidrule(lr){6-6}

Qwen2.5 32B & 0.066 & 0.010 & -0.374\textsuperscript{***} & -0.217\textsuperscript{***} & -0.250\textsuperscript{***}\\
Qwen2.5 14B & 0.006 & -0.008 & -0.366\textsuperscript{***} & -0.221\textsuperscript{***} & -0.263\textsuperscript{***}\\
Qwen2.5 7B & -0.030 & -0.007 & -0.257\textsuperscript{***} & -0.137\textsuperscript{***} & -0.132\textsuperscript{***}\\
Phi3.5 (MoE) & 0.133 & 0.039 & -0.177\textsuperscript{***} & -0.114\textsuperscript{***} & -0.149\textsuperscript{***}\\
Phi3 (medium) & 0.105 & 0.100\textsuperscript{*} & -0.201\textsuperscript{***} & -0.129\textsuperscript{***} & -0.139\textsuperscript{***}\\
Phi3 (small) & 0.221* & 0.021 & -0.120\textsuperscript{**} & -0.068\textsuperscript{*} & -0.033\\
Phi3 (mini) & 0.021 & 0.099\textsuperscript{*} & -0.043 & -0.018 & -0.060\textsuperscript{*}\\

    \bottomrule
    \end{tabular}
    \caption{Phi coefficients $\phi$ between the correctness of the answer for a multi-hop question with sufficient evidence and the correctness of derived unanswerable questions with insufficient evidence or derived false-premise questions  and uncertain specificity questions. 
     \footnotesize
    \textit{Note:} * \textit{p} < 0.05; ** \textit{p} < 0.01; *** \textit{p} < 0.001.}
    \label{tab:unanswerable-phi-coefficients}
\end{table*}

\subsection{Time-span Error Analysis}
\label{appendix:experiments:error-analysis-timespan}
Figure~\ref{fig:benchmark-timespan-errors} shows model mispredictions on time-span questions with sufficient and insufficient evidence from the main experiments. Smaller LLMs make more mispredictions on answerable instances and are more prone to falling for distractors. When evidence is insufficient, the larger Qwen2.5 models and Phi3.5 MoE frequently answer as if sufficient evidence were available. We hypothesize this occurs because these questions often require calculating the time between events, which can be guessed without verifying event alignment in the article.
\input{figures/benchmark-timespan-error}

\subsection{Correct Predictions without Evidence}
\label{appendix:experiments:gpt4turbo}
Below, we present three randomly selected multi-hop questions that GPT-4 Turbo answered correctly without access to evidence.

\rule{\linewidth}{0.1pt}
\noindent
\textbf{Question 1:} \emph{What was the percentage increase in voter turnout during the pilot phase in the region whose success was emphasized by Iras Danley as a blueprint for addressing challenges in areas with difficult terrain and sparse populations?}
\begin{enumerate}[noitemsep]
    \item 15\%
    \item 35\% (\textbf{correct})
    \item 45\%
    \item 28\%
\end{enumerate}

\rule{\linewidth}{0.1pt}
\noindent
\textbf{Question 2:} \emph{What is the name of the centerpiece installation created by the individual who adapted her creative process to align with new guidelines, emphasizing sustainable materials and environmental testing?}
\begin{enumerate}[noitemsep]
    \item ``Rebirth in Motion'' (\textbf{correct})
    \item ``Echoes of Harmony'' 
    \item ``Resonance of Memories''
    \item ``Industrial Bloom''
\end{enumerate}

\rule{\linewidth}{0.1pt}
\noindent
\textbf{Question 3:} \emph{What specific issue, mentioned by a clinic administrator in Larnwick, could be mitigated by the app described as using a "citizen-led data trust model" to manage and anonymize aggregated data? }
\begin{enumerate}[noitemsep]
    \item Challenges in recruiting independent data privacy experts for community feedback sessions. 
    \item Resource shortages caused by delays in identifying hotspots during past norovirus outbreaks. (\textbf{correct})
    \item Mixed public opinions about the app's privacy safeguards in Misterine City. 
    \item Concerns about the app's encryption protocols being insufficient to prevent cyberattacks.
\end{enumerate}

\subsection{Analysis over Varying Numbers of Irrelevant Documents}
\label{appendix:experiments:error-analysis-irrelevant-doc-numbers}
\input{figures/figures-influence-per-context-per-model}

Figure~\ref{fig:influence-combined} shows the performance for each LLM and each question category as the number of irrelevant documents increases from 0-80 in intervals of 20.

\subsection{Change in Prediction for False Premise and Uncertain Specificity Questions}
\label{appendix:experiments:error-analysis-prediction-change-fp-us}
\input{figures/figures-qwen32-context-predictions-comparison}

Figure~\ref{fig:sankey-multi-hop-unanswerable} shows how predictions change when the multi-hop question is turned into an unanswerable false premise or uncertain specificity question. We compare the models Phi3 (medium) and Qwen2.5 14B, which have comparable parameter counts and use the same prompt.
When only the two required news articles are provided (top), Phi3 shows minimal deflection, performing well on answerable questions but poorly when deflection is needed. In contrast, Qwen2.5 is more cautious, making some false deflections on answerable questions but is better at detecting unanswerable ones. Qwen2.5 also remains more stable after adding 80 irrelevant documents, while Phi3 tends to select distractors that appear superficially relevant. 

\subsection{Parsing}
\label{appendix:parsing}
All prompts require the model to provide the answer on the last line of its response by stating the number of the selected option. Alternatively, the answer is acceptable if the chosen option is explicitly stated within the response. Overall, LLMs in our experiments successfully provided answers, indicating that our findings stem from their reasoning abilities rather than poor instruction-following. The successful response rate per experiment and model is shown in Table~\ref{tab:parsed}. GPT-4 Turbo had an answer rate of 100\% in Section~\ref{sec:impact-of-parametric-knowledge}.

\begin{table*}[h!]
\small
    \centering
    \begin{tabular}{lccc}
    \toprule
    \textbf{LLM}  & \textbf{Main Exp.} & \textbf{Main Exp. (CoT)} & \textbf{Irrelevant Doc.} \\
    \midrule
    Phi3 (mini) & 99.7 & 95.3 & 99.6 \\
    Phi3 (small) & 98.2 & 97.4 & 98.9 \\
    Phi3 (medium) & 99.7 & 99.0 & 99.9 \\
    Phi3.5 (MoE) & 99.7 & 99.3 & 99.9 \\
    Qwen2.5 (7B) & 97.4 & 94.5 & 97.8 \\
    Qwen2.5 (14B) & 99.7 & 99.6 & 99.7 \\
    Qwen2.5 (32B) & 100.0 & 99.9 & 100.0 \\
    \midrule
    \textbf{Average} & 99.2 & 97.9 & 99.4 \\

    \bottomrule
    \end{tabular}
    \caption{Response rates (\%) of the LLMs from the main experiments in Section~\ref{sec:benchmark-experiments}, with (\textbf{Main Exp. (CoT)}) and without (\textbf{Main Exp}) CoT prompting, and from the ablation on the number of irrelevant documents (\textbf{Irrelevant Doc.}) in Section~\ref{sec:irrelevant-documents-ablation}.}
    \label{tab:parsed}
\end{table*}

%% file: figures/figure-error-analysis-insufficient-multi-hop.tex
\begin{figure*}[h]
  \includegraphics[width=\textwidth]{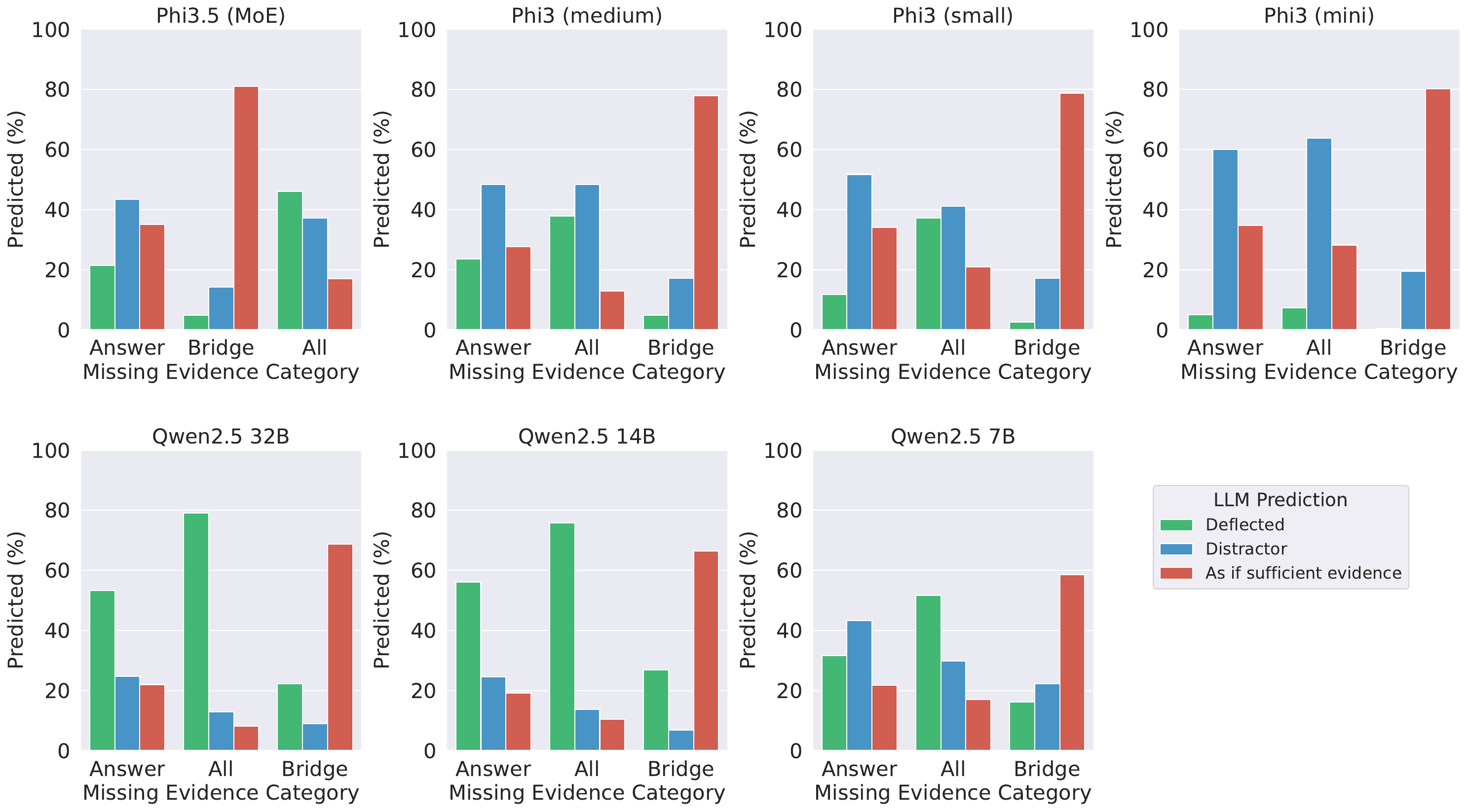}
  \caption{Ratio correct deflections (green) and incorrect predictions (blue and red) for multi-hop questions with different categories of insufficient evidence. We omit the information required to resolve the bridge entity (\emph{Bridge}), or the information that contains the answer (\emph{Answer}), or both (\emph{All}).}
  \label{fig:error-analysis-insufficient-multi-hop}
\end{figure*}

%% file: figures/benchmark-timespan-error.tex
\begin{figure}[t]
  \includegraphics[width=\linewidth]{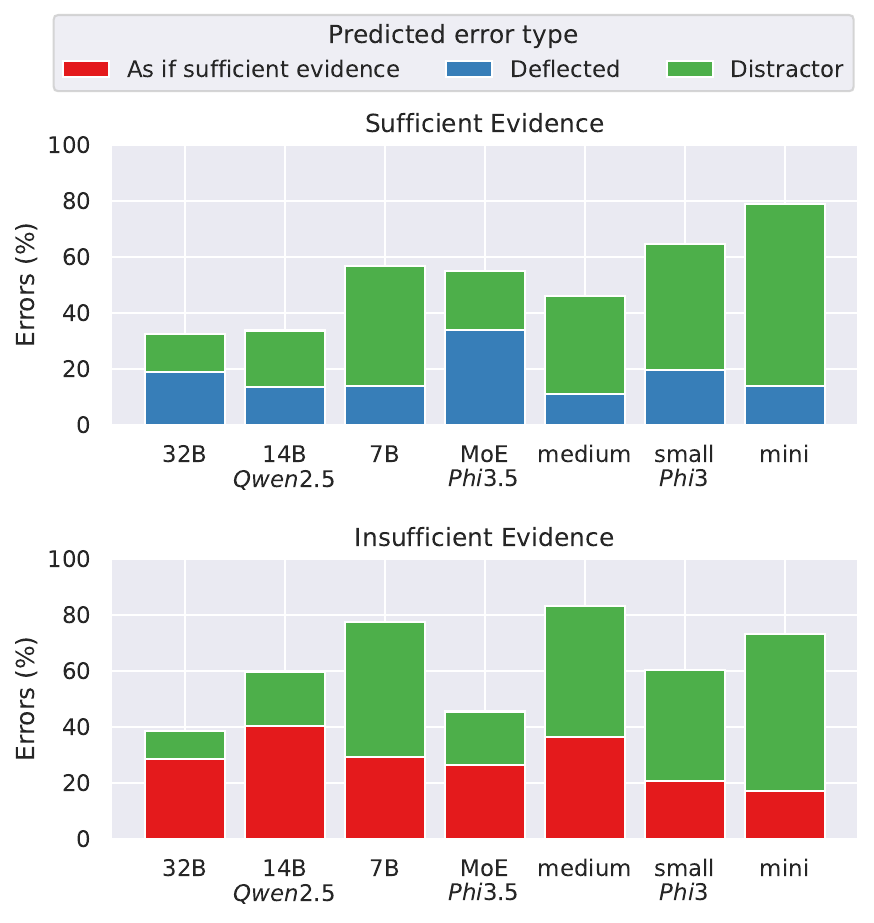}
  \caption{
  Error categories on time-span questions with sufficient (top) and insufficient (bottom) evidence.
  }
  \label{fig:benchmark-timespan-errors}
\end{figure}

%% file: figures/figures-influence-per-context-per-model.tex
\begin{figure*}[h]
  \centering
  \begin{subfigure}{0.8\textwidth}
    \includegraphics[width=\textwidth]{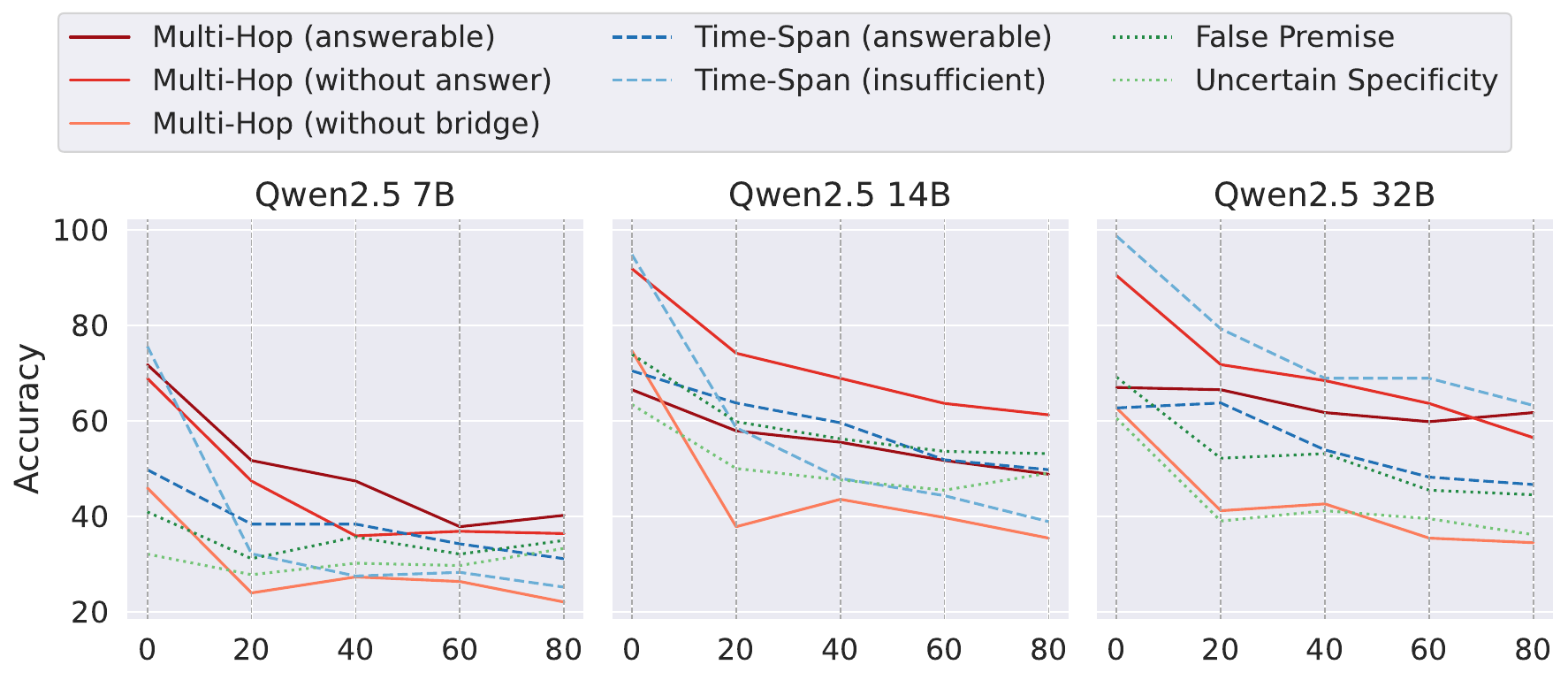}
    \label{fig:influence-qwen}
  \end{subfigure}
  
  \vspace{-1em} %
  
  \begin{subfigure}{0.8\textwidth}
    \includegraphics[width=\textwidth]{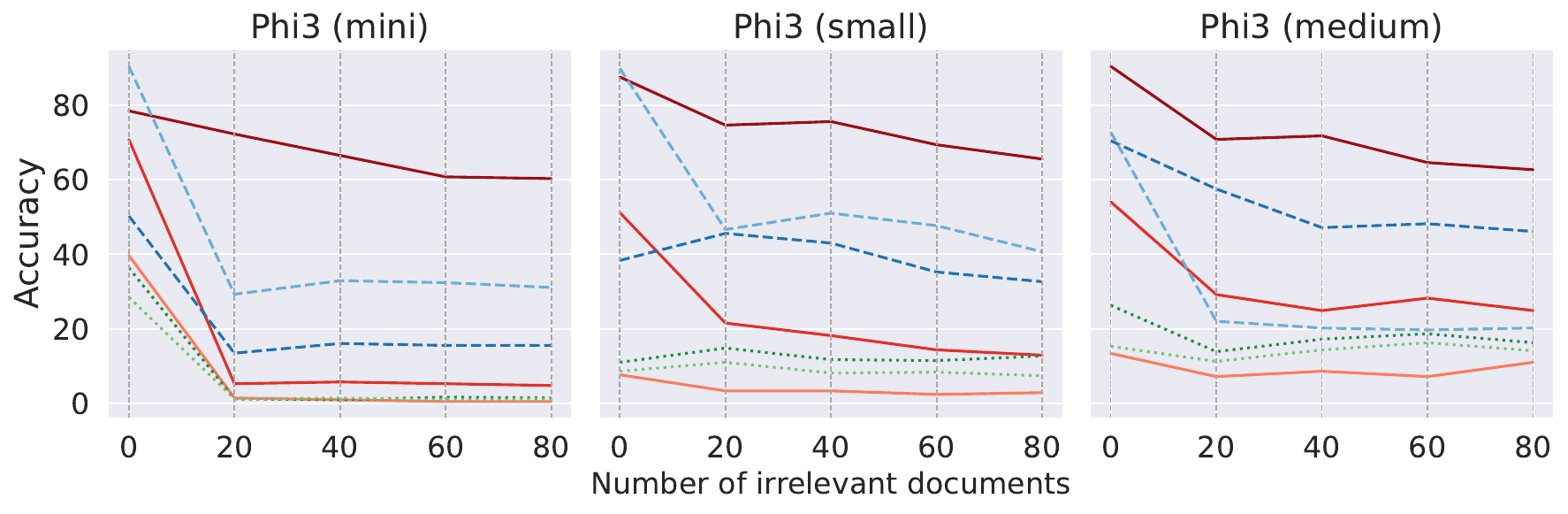}
    \label{fig:influence-phi}
  \end{subfigure}

  \caption{Performance per LLM and question type by the number of added irrelevant documents.}
  \label{fig:influence-combined}
\end{figure*}

%% file: figures/figures-qwen32-context-predictions-comparison.tex
\definecolor{sankey_correct}{RGB}{31,119,180}
\definecolor{sankey_deflect}{RGB}{255, 127, 14}

\begin{figure}[htbp]
    \centering

    \begin{subfigure}[t]{0.4\linewidth}
        \centering
        \includegraphics[width=\linewidth]{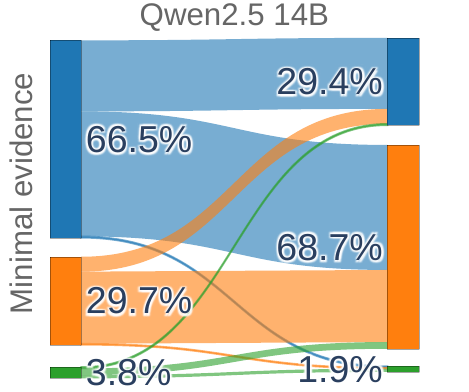}
    \end{subfigure}
    \begin{subfigure}[t]{0.4\linewidth}
        \centering
        \includegraphics[width=\linewidth]{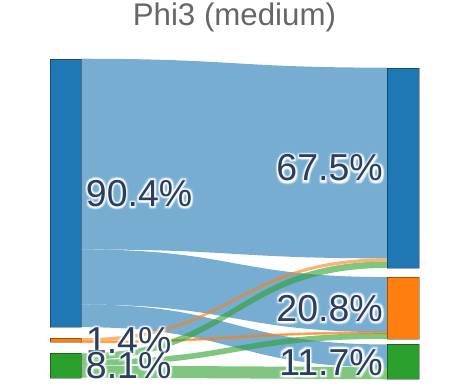}
    \end{subfigure}

    \begin{subfigure}[t]{0.4\linewidth}
        \centering
        \includegraphics[width=\linewidth]{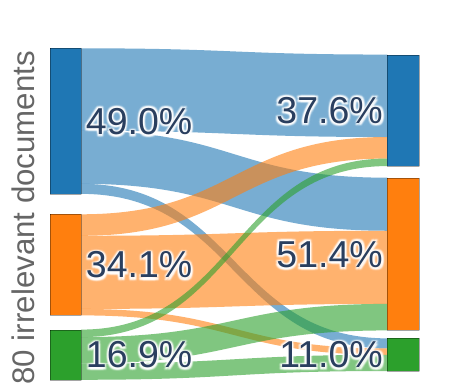}
    \end{subfigure}
    \begin{subfigure}[t]{0.4\linewidth}
        \centering
        \includegraphics[width=\linewidth]{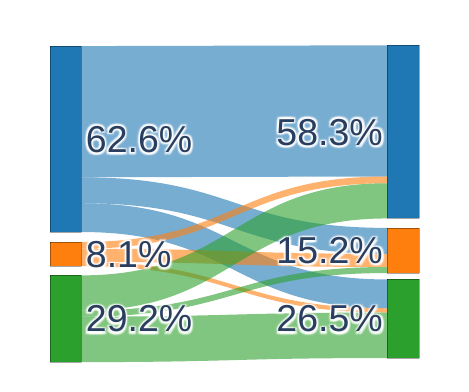}
    \end{subfigure}
    \begin{subfigure}[t]{\linewidth}  %
        \centering
        \includegraphics[width=0.9\linewidth]{images/legend-sankey.pdf} %
    \end{subfigure}

    \caption{Change in prediction between the multi-hop and the derived false premise and uncertain specificity questions with minimal evidence (top) or additional irrelevant documents (bottom). A correct flow goes from all \textcolor{sankey_correct}{\textbf{blue}} to all \textcolor{sankey_deflect}{\textbf{orange}}.}
    \label{fig:sankey-multi-hop-unanswerable}
\end{figure}

%% file: latex/appendix/prompts.tex
\section{Other Details}
\label{appendix:details}

\subsection{Models}
For dataset generation we use GPT-4o. The temperature for the dataset construction differs and is specified for each step in Appendix~\ref{appendix:dataset-construction}.
All experiments use a temperature of $t=0.0$.
The experiments using the Qwen2.5 and Phi3, Phi3.5 models ran on a cluster of A100 80GB GPUs with Flash~Attention~2~\citep{dao2024flashattention} using the Transformers library \citep{wolf2019huggingface}. 
All used models with the context size are listed in Table~\ref{tab:use-llms}.
\begin{table*}[h]
\small
\centering
\begin{tabular}{llcc} 
\toprule
\textbf{Model} & \textbf{Model Version} & \textbf{Model Size} & \textbf{Context size} \\
\midrule
Phi3 (mini) & microsoft/Phi-3-mini-128k-instruct & 3.8B & 128k \\
Phi3 (small) & microsoft/Phi-3-small-128k-instruct & 7B & 128k \\
Phi3 (medium) & microsoft/Phi-3-medium-128k-instruct & 14B & 128k \\
Phi3.5 (MoE) & microsoft/Phi-3.5-MoE-instruct & 16$\times$3.8B& 128k\\
Qwen2.5 7B & Qwen/Qwen2.5-7B-Instruct & 7B & 128k \\
Qwen2.5 14B & Qwen/Qwen2.5-14B-Instruct & 14B & 128k \\
Qwen2.5 32B & Qwen/Qwen2.5-32B-Instruct & 32B & 128k \\
\bottomrule
\end{tabular}
\caption{Used LLMs with their maximal context size and number of parameters.}
\label{tab:use-llms}
\end{table*}

\subsection{Writing}
We refined our initial draft and improved the writing using ChatGPT and Grammarly.
Prompts were generated and refined using the Prompt Generator\footnote{\url{https://docs.anthropic.com/en/docs/build-with-claude/prompt-engineering/prompt-generator}} provided by Anthropic.

\subsection{Used Artifacts}
In Section~\ref{sec:impact-of-parametric-knowledge}, we experiment with publicly available RealTimeQA data~\citep{kasai2024realtime}. Additionally, we collected data via the Wayback Machine to ensure reproducibility, but we do not share this data verbatim. For proprietary LLMs, we used GPT-4 Turbo and GPT-4o, both of which underwent in-house approval processes. The remaining experiments rely on open-source models (Phi3, Phi3.5, and Qwen2.5) which were approved through in-house legal reviews.

\onecolumn

\section{Prompts}
\label{appendix:prompts}

\subsection{MCQ Prompt on RealtimeQA without Evidence}
\label{appendix:prompts:realtimeqa-noevidence}
\noindent
\footnotesize
\begin{tcolorbox}[
    colback=gray!15,
    colframe=gray!15,
    coltext=darkgray,
    boxrule=0pt,
    width=\textwidth,
    boxsep=5pt,
    sharp corners,
    breakable
]You are an AI assistant tasked with answering multiple-choice questions using your knowledge and common sense. Your goal is to select the best answer or make the most informed guess possible. You must choose one of the provided options - no exceptions. \\
\newline
**Question (from \{\{DATE\}\}):** \\
<question> \{\{QUESTION\}\} </question> \\
\newline
**Answer Options:** \\
\{\{ANSWERS\}\} \\
\newline
If the question pertains to events beyond your knowledge cutoff, make an educated guess. You must select an answer. \\
\newline
Output only the selected answer in this exact format, with no additional text, explanations, or symbols: \\
Answer: [selected answer number]
\end{tcolorbox}

\subsection{Event Outline Generation Prompt}
\label{appendix:prompts:event:outline-generation}
\noindent
\footnotesize
\begin{tcolorbox}[
    colback=gray!15,
    colframe=gray!15,
    coltext=darkgray,
    boxrule=0pt,
    width=\textwidth,
    boxsep=5pt,
    sharp corners,
    breakable
]
You are an AI assistant tasked with generating an outline for a fictional event. Your goal is to create a realistic, entirely fictional event that does not overlap with real-world named entities or known fictional named entities. Follow these instructions carefully:\\
\newline
First, review the list of already known fictional named entities of this fictional world:\\
\newline
<known\_entities>\\
<LOCATIONS>\\
\{\{LOCATIONS\_XML\}\}\\
</LOCATIONS>\\
\newline
<PERSONS>\\
\{\{PERSONS\_XML\}\}\\
</PERSONS>\\
\newline
<ORGANIZATIONS>\\
\{\{ORGANIZATIONS\_XML\}\}\\
</ORGANIZATIONS>\\
\newline
<PRODUCTS>\\
\{\{PRODUCTS\_XML\}\}\\
</PRODUCTS>\\
\newline
<ARTS>\\
\{\{ARTS\_XML\}\}\\
</ARTS>\\
\newline
<EVENTS>\\
\{\{EVENTS\_XML\}\}\\
</EVENTS>\\
\newline
<BUILDINGS>\\
\{\{BUILDINGS\_XML\}\}\\
</BUILDINGS>\\
\newline
<MISCELLANEOUS>\\
\{\{MISCELLANEOUSS\_XML\}\}\\
</MISCELLANEOUS>\\
</known\_entities>\\
Next, review the outline of previous events that have occurred in this fictional world:\\
\newline
<history>\\
\{\{HISTORY\_XML\}\}\\
</history>\\
\newline
Now, consider the following information about the new event you need to generate as a continuation of the past events:\\
\newline
Date: \{\{DATE\}\}\\
\newline
Event Summary: \{\{EVENT\_SUMMARY\}\}\\
\newline
Genre: \{\{GENRE\}\}\\
\newline
Follow these guidelines to generate the event outline:\\
\newline
1. Create an entirely fictional event based on the given genre, event summary, and history of previous events. The event must be realistic but must not reference any existing real-world or known fictional named entities.\\
2. Invent new named entities as needed, ensuring they don't exist in the real world or in existing works of fiction. When creating names, use unique combinations unlikely to match real named entities.\\
3. Construct the outline using short, concise, factual, and objective statements. Each statement must discuss only one fact or sub-event, structured sequentially in a logical temporal order when applicable.\\
4. Ensure all statements form a coherent outline.\\
5. Output each statement within a <storyitem> tag.\\
6. Generate exactly \{\{NUM\_STORYITEMS\}\} distinct story items.\\
7. Ensure logical progression, with each statement following chronologically when applicable. Include a mix of main events, reactions, consequences, and contextual information.\\
8. Make storyitems as atomic as possible, communicating only a single piece of relevant information per item. Do not merge multiple pieces of information into one storyitem.\\
9. Ensure the story sounds realistic without explicitly stating it's fictional.\\
10. Maintain consistency with the provided <history> that discusses past events fictional events. The outline must logically follow chronological events described in the history.\\
11. Incorporate some or all of the provided named entities in your outline. Ensure that any mention of these entities is consistent with the information you have about the named entity. You may introduce additional fictional entities as needed, but they must not conflict with the existing ones.\\
12. When referencing any named entities from the provided inputs, maintain consistency in their descriptions and roles within the story.\\
13. If no date is provided: Generate a complete date for the event, including the year. The date should be formatted as "year-month-day" (e.g., "2024-12-03" or "2025-06-13"). This date should be consistent with the timeline established in the <history>.\\
14. If a date is provided: Use the provided date.\\
15. The outline can include quotes from the named entities where applicable.\\
16. Do not repeat the information from the previous events from the <history>.\\
17. Refer to all named entities (the new named entities and the known named entities) by their full "name" property. DO NOT refer to the named entities using the ID.\\
18. Make sure that you refer to all named entities within each storyitem per full name at least once. DO NOT use pronouns to refer to a named entity from the previous story item.\\
19. Think about the content that is appropriate for the event summary given the genre, provided history: Think about which dimensions align with all of those, and sound like a realistic event.\\
\newline
Your output should be formatted as follows:\\
<scratchpad>[Your thoughts go here]</scratchpad>\\
<results>\\
<date>year-month-day</date>\\
<outline>\\
<storyitem>First story item</storyitem>\\
<storyitem>Second story item</storyitem>\\
<storyitem>Third story item</storyitem>\\
...\\
</outline>\\
</results>\\
\newline
IMPORTANT:\\
- The event must be entirely realistic, even though it is fictional. Do not include any science fiction or fantasy elements. The story should read like a plausible current event.\\
- Do not use any galactic events. The fictional world should be similar to our world but not about galaxies or outer space.\\
- Each story item must only discuss one fact or subevent. Ensure that each story item is specific, concise, and focused on a single piece of information.\\
- Begin your response with <results> and end it with </results>. Do not include any text outside of these tags.\\
- Do not exaggerate the outline. Avoid using words like "groundbreaking", "worldwide", "global". Keep the outline and the scope and influence of the event realistic.\\
- Do not create outlines with global or national impact unless the genre specifically requires it. Instead, focus on smaller or local developments.\\
- Do not focus on technological discoveries or topics like AI tools, virtual reality, augmented reality, 3D-modelling, quantum computing, etc. You may include such topics only if they are HIGHLY relevant to the genre \{\{GENRE\}\} AND the provided history of events.\\
- Focus on realistic, meaningful outlines with specific details and events that align with typical, realistic scenarios of the genre \{\{GENRE\}\}.\\
\newline
Remember:\\
The outline should center on a fictional but realistic event, keeping its scale aligned with the event summary and provided <history> without exaggerating its impact. Rather than overstating the event's significance, the outline should stay within the scope appropriate to the genre, provided history, and provided summary. When in doubt, focus on detailed, localized developments instead of amplifying global effects.\\
\newline
Ensure that the outline is coherent, follows a logical sequence, and offers a unique perspective on the given event while maintaining consistency with the provided background information.
\end{tcolorbox}

\subsection{Event Outline Refinement Prompt}
\label{appendix:prompts:event:outline-refinement}
\noindent
\footnotesize
\begin{tcolorbox}[
    colback=gray!15,
    colframe=gray!15,
    coltext=darkgray,
    boxrule=0pt,
    width=\textwidth,
    boxsep=5pt,
    sharp corners,
    breakable
]You are an AI assistant tasked with analyzing a fictional event summary and its corresponding outline to enrich it with additional NEW specific details. Follow these instructions carefully:\\
\newline
1. Read the provided fictional event summary of the genre \{\{GENRE\}\}:\\
\newline
<event\_summary>\\
\{\{EVENT\_SUMMARY\}\}\\
</event\_summary>\\
\newline
2. These are the fictional known entities:\\
<known\_entities>\\
<LOCATIONS>\\
\{\{LOCATIONS\_XML\}\}\\
</LOCATIONS>\\
\newline
<PERSONS>\\
\{\{PERSONS\_XML\}\}\\
</PERSONS>\\
\newline
<ORGANIZATIONS>\\
\{\{ORGANIZATIONS\_XML\}\}\\
</ORGANIZATIONS>\\
\newline
<PRODUCTS>\\
\{\{PRODUCTS\_XML\}\}\\
</PRODUCTS>\\
\newline
<ARTS>\\
\{\{ARTS\_XML\}\}\\
</ARTS>\\
\newline
<EVENTS>\\
\{\{EVENTS\_XML\}\}\\
</EVENTS>\\
\newline
<BUILDINGS>\\
\{\{BUILDINGS\_XML\}\}\\
</BUILDINGS>\\
\newline
<MISCELLANEOUS>\\
\{\{MISCELLANEOUSS\_XML\}\}\\
</MISCELLANEOUS>\\
</known\_entities>\\
\newline
3. Review the outline of previous events that have occurred in this fictional world:\\
\newline
<history>\\
\{\{HISTORY\_XML\}\}\\
</history>\\
\newline
4. Review the date and outline of the event:\\
\newline
Date: \{\{DATE\}\}\\
<outline>\\
\{\{OUTLINE\}\}\\
</outline>\\
\newline
5. Your task is to enrich this outline with additional details. The enhanced outline must discuss the same events as before and must not extend the events that happened in the outline. It must only provide supplementary details about these events in the outline.\\
\newline
6. Follow these rules for enrichment:\\
   a. Examine each sentence in the provided outline.\\
   b. For each sentence, identify information that is unspecific or can be elaborated with more detail.\\
   c. Consider the outline to be all information that is provided to a reporter about this fictional event and only include additional specific details that could also be known to the reporter at this point in time:\\
      - Some events may still be ongoing and some information may not be available at this point in time.\\
      - Do not include information that would likely not be known at this point in time.\\
      - Consider the perspective of what is currently known about this fictional event when adding details.\\
   d. When you find something unspecific, add a new sentence with supplementary specific details:\\
      - Place the new sentence directly after the original sentence in the outline.\\
      - Ensure the new sentence does not repeat content from the previous sentence.\\
      - Focus solely on providing supplementary specific details in the new sentence.\\
      - Make the new sentence self-contained and coherent on its own (do not refer to previously mentioned named entities by pronoun. Instead, directly refer to them via the name).\\
   e. For each original sentence generate up to \{\{NUM\_SPECIFIC\_SENTS\}\} novel sentences that introduce supplementary details.\\
\newline
7. Additional guidelines:\\
   - Do not modify the existing sentences. Only add new sentences for supplementary details.\\
   - Ensure added sentences focus exclusively on new, specific information without repeating existing content.\\
   - Maintain consistency with the original outline in all additional specifics.\\
   - Do not introduce new subevents. Only provide more details about the events already mentioned.\\
   - Make sure the additional details provided are new and do not reiterate details known from the fictional history or the list of fictional known entities.\\
   - Make sure that the new details are not contradictory to the history (<history>) and known entities (<known\_entities>).\\
   - Do not exaggerate the outline. Avoid using words like "groundbreaking", "worldwide", "global". Keep the outline and the scope and influence of the event realistic.\\
   - Do not create outlines with global or national impact unless the genre specifically requires it. Instead, focus on smaller or local developments.\\
   - Do not focus on technological discoveries or topics like AI tools, virtual reality, augmented reality, 3D-modelling, quantum computing, etc. You may include such topics only if they are HIGHLY relevant to the genre \{\{GENRE\}\} AND the provided history of events.\\
   - Focus on realistic, meaningful outlines with specific details and events that align with typical, realistic scenarios of the genre \{\{GENRE\}\}.\\
\newline
8. Present your enriched outline in the following format:\\
   - Make sure that the sentences with the additional specific details are listed as separate <storyitem> and placed at the correct position within the outline.\\
   - Treat each new sentence you have created as a separate <storyitem>.\\
   - Each sentence provided to you in the outline forms one <storyitem> and must not be changed.\\
   - Each new sentence you have written that provides additional specific details forms one <storyitem> and must be listed separately.\\
\newline
Important:\\
Before you start writing the new sentences with specific details, think about various dimensions that could be extended that align well with the genre, history, provided event summary and existing outline. Think about specific directions that could be of interest within the current genre (\{\{GENRE\}\}) and brainstorm how you could deepen the outline with new specific details on these interesting dimensions. Carefully decide when it is reasonable to provide technical details, when it make more sense to provide quotes, visions, etc., when to provide background information. Think about information that are of interest to a reader of a newspaper with the genre \{\{GENRE\}\}. Be in particular careful before introducing technological details and first examine if these details are appropriate for the genre or not.\\
Double-check that each detail is compatible with all the existing information you are provided with.\\
\newline
Use this structure:\\
<scratchpad>[Your thoughts go here.]</scratchpad>\\
<results>\\
<storyitems>\\
<storyitem>[Insert first storyitem here]</storyitem>\\
<storyitem>[Insert second storyitem here]</storyitem>\\
<storyitem>[Continue with additional storyitems as needed]</storyitem>\\
</storyitems>\\
</results>\\
\newline
Maintain the chronological order and logical flow of the original outline while adding your supplementary details. Each <storyitem> should contain one sentence from the original outline or one new sentence with additional specific details.
\end{tcolorbox}

\subsection{Event Outline Consistency Prompt}
\label{appendix:prompts:event:outline-consistency}
\noindent
\footnotesize
\begin{tcolorbox}[
    colback=gray!15,
    colframe=gray!15,
    coltext=darkgray,
    boxrule=0pt,
    width=\textwidth,
    boxsep=5pt,
    sharp corners,
    breakable
]You are an AI assistant tasked with checking the consistency of a fictional story outline with previously established entities and events. Your goal is to ensure that the new outline is a consistent continuation of the previous events in the history.\\
\newline
You will be provided with three key pieces of information:\\
\newline
1. A list of fictional entities:\\
<entities>\\
<LOCATIONS>\\
\{\{LOCATIONS\_XML\}\}\\
</LOCATIONS>\\
\newline
<PERSONS>\\
\{\{PERSONS\_XML\}\}\\
</PERSONS>\\
\newline
<ORGANIZATIONS>\\
\{\{ORGANIZATIONS\_XML\}\}\\
</ORGANIZATIONS>\\
\newline
<PRODUCTS>\\
\{\{PRODUCTS\_XML\}\}\\
</PRODUCTS>\\
\newline
<ARTS>\\
\{\{ARTS\_XML\}\}\\
</ARTS>\\
\newline
<EVENTS>\\
\{\{EVENTS\_XML\}\}\\
</EVENTS>\\
\newline
<BUILDINGS>\\
\{\{BUILDINGS\_XML\}\}\\
</BUILDINGS>\\
\newline
<MISCELLANEOUS>\\
\{\{MISCELLANEOUSS\_XML\}\}\\
</MISCELLANEOUS>\\
</entities>\\
\newline
2. A history of fictional events involving these entities:\\
<history>\\
\{\{HISTORY\_XML\}\}\\
</history>\\
\newline
3. The date of the next fictional event: \{\{DATE\}\}\\
\newline
3. An outline describing the next fictional event:\\
<outline>\\
\{\{OUTLINE\}\}\\
</outline>\\
\newline
Follow these steps to complete your task:\\
\newline
1. Carefully compare the events described in the outline with the events from the history. Look for any inconsistencies or contradictions.\\
\newline
2. Compare the named entities described in the outline with the list of provided named entities. Ensure they are consistent.\\
\newline
3. Note that changes to known named entities are acceptable if they are reasonably discussed within the outline.\\
\newline
4. If you find any inconsistencies or contradictions:\\
   a. Make minimal changes to the outline to resolve the issues.\\
   b. Ensure your changes maintain the original structure and flow of the outline as much as possible.\\
   c. Double-check after fixing the inconsistencies to ensure they no longer persist.\\
\newline
5. If the outline is consistent with the provided list of entities and previous events (i.e., not contradictory), output the outline without any changes.\\
\newline
Present your results in the following format:\\
\newline
<results>\\
<scratchpad>\\
(Your reasoning process, including any inconsistencies found and how you resolved them) \\
</scratchpad>\\
<storyitems>\\
<storyitem>[First story item from the outline, corrected if necessary]</storyitem>\\
<storyitem>[Second story item from the outline, corrected if necessary]</storyitem>\\
<storyitem>[Continue with additional story items as needed]</storyitem>\\
</storyitems>\\
</results>\\
\newline
Important reminders:\\
- Always refer to named entities using their full name as described by the "name" property.\\
- Remember that all events are fictional but should sound plausible and realistic.\\
- Your output must include all storyitems, whether you had adjusted them or not.\\
- Do not add new events or significantly alter the plot. Your task is to ensure consistency, not to rewrite the story.\\
\newline
If you cannot resolve an inconsistency without significantly changing the plot, note this in your scratchpad and leave the story item as is, highlighting the inconsistency.\\
\newline
Begin your analysis now, and provide your results in the format specified above.
\end{tcolorbox}

\subsection{Event Outline (novel) Named Entity Recognition Prompt}
\label{appendix:prompts:event:outline-ner-new}
\noindent
\footnotesize
\begin{tcolorbox}[
    colback=gray!15,
    colframe=gray!15,
    coltext=darkgray,
    boxrule=0pt,
    width=\textwidth,
    boxsep=5pt,
    sharp corners,
    breakable
]You are an AI assistant tasked with identifying new named entities from a given outline of a fictional event description. Your goal is to identify any new named entities mentioned in the OUTLINE that are not already present in the provided lists of existing entities.\\
\newline
First, carefully read and analyze the following OUTLINE:\\
\newline
Date: \{\{DATE\}\}\\
<OUTLINE>\\
\{\{OUTLINE\}\}\\
</OUTLINE>\\
\newline
Now, review the existing named entities in the following XML structures. Note that these lists may be empty or contain partial information:\\
\newline
<entities>\\
\{\{LOCATIONS\_XML\}\}\\
\{\{PERSONS\_XML\}\}\\
\{\{ORGANIZATIONS\_XML\}\}\\
\{\{PRODUCTS\_XML\}\}\\
\{\{ARTS\_XML\}\}\\
\{\{EVENTS\_XML\}\}\\
\{\{BUILDINGS\_XML\}\}\\
\{\{MISCELLANEOUSS\_XML\}\}\\
</entities>\\
\newline
The definitions with examples for each named entity category are listed here:\\
- Person\\
Definition: Individual human beings, including fictional figures.\\
Examples: Barack Obama, William Shakespeare, Harry Potter, Marie Curie\\
\newline
-Organization\\
Definition: Groups of people working together for a common purpose, including companies, institutions, ethnic groups, communities and government bodies.\\
Examples: United Nations, Apple Inc., Harvard University, Greenpeace, Maori\\
\newline
- Location\\
Definition: Geographical or political areas, locations, countries (together with their nationalities), cities, natural landmarks, and regions.\\
Examples: Paris, Mount Everest, California, Amazon Rainforest, Japan, Australia, Australian\\
\newline
- Product\\
Definition: Goods or services created for consumer use or commercial purposes.\\
Examples: iPhone, Coca-Cola, Microsoft Office, Tesla Model 3\\
\newline
- Art\\
Definition: Creative works in various forms, including artifacts, ornaments, visual arts, literature, music, and performance.\\
Examples: Mona Lisa, To Kill a Mockingbird, Beethoven's Symphony No. 9, Hamilton (musical)\\
\newline
 - Building\\
Definition: Structures designed for human occupancy or use, including residential, commercial, and public structures.\\
Examples: Empire State Building, Taj Mahal, Sydney Opera House, Buckingham Palace\\
\newline
 - Event\\
Definition: Significant occurrences or planned gatherings, including historical moments, celebrations, and competitions.\\
Examples: World War II, Olympic Games, Woodstock Music Festival, Super Bowl\\
\newline
 - Miscellaneous\\
Definition: Other named entities that don't fit into the above categories, such as abstract concepts, unique identifiers.\\
Examples: Theory of Relativity, Morse Code, Brexit, Zodiac Signs\\
\newline
IMPORTANT: You must ONLY identify a named entity if the outline explicitly refers to the entity by name.\\
DO NOT list entities for events which are not explicitly referred to by name. \\
DO NOT list entities for buildings that are not explicitly referred to by name (e.g. a big Chicago villa)\\
\newline
Example: "On the 27th birthday of Carla Short"\\
- "Carla Short" is a named entity (Person)\\
- No named event is in this statement\\
\newline
Example: "On the third day of the Banana Split Festival I left my keys"\\
- National Banana Split Festival is a named entity (Event)\\
\newline
Example: "I sold my $13M Chicago villa."\\
- "Chicago" is a named entity (Location)\\
- "$13M Chicago villa" is NOT a named entity. However, this phrase contains the named entity Chicago (location)\\
\newline
Example: "I bought tickets to go on the Sky Needle, the highest skyscraper in town!"\\
- "Sky Needle" is a named entity (building)\\
\newline
Example: "Aboriginal people from a well known Australian city."\\
- "Aboriginal" is a named entity (Organization)\\
- "Australia" is a named entity (Location)\\
\newline
To complete this task, follow these steps:\\
\newline
1. Identification of named entities: Carefully read through the OUTLINE and identify all named entities (locations, persons, organizations, products, art, events, buildings, or miscellaneous) that are mentioned by name. Consider all named entities. Carefully follow the definition and examples provided for each entity type above. DO NOT ignore named entities from the outline that exist in the real world. Carefully double-check that you did not miss any named entities. If you are in doubt about some named entities, epxlain why you bare not sure.\\
\newline
2. Remove entities that are already known and only keep new named entities:\\
   a. For each named entity you identify, check if it already exists in the <entities>. If <entities> is empty, consider all named entities as new.\\
   b. Only keep new named entities which cannot be found in the <entities> list of known named entities.\\
   c. Ensure that each new named entity is explicitly mentioned by name in the <outline>. The outline may refer to various entities (such as events, buildings, products) that are described generally but not by a specific name. These should not be identified as named entities.\\
\newline
3. Verification and refinement of new named entities:\\
   a. Review your preliminary list of new named entities.\\
   b. Double-check each named entity against the existing XML structures to ensure it is truly new and not already present.\\
   c. For each new named entity, determine its full name as it appears in the OUTLINE. Do not infer or create additional information.\\
   d. Ensure that the full name of each named entity is distinct. Two different entities MUST NOT share the same full name.\\
   e. Examine your list of verified new named entities for any redundancies. If any entities refer to the same person, place, organization, product, art piece, event, building, or miscellaneous item but are mentioned with slight variations in the OUTLINE, keep only the most complete or accurate version and remove the others.\\
   f. Do not list abbreviations as separate entities. If an abbreviation is used, do not include it with the full name of the entity, and do not create a separate entry for it.\\
   g. Ensure that each identified named entity refers to an actual identifiable entity mentioned in the OUTLINE.\\
\newline
4. Formatting and outputting the final list:\\
   a. Categorize each verified new named entity as either a location, person, organization, product, art, event, building, or miscellaneous.\\
   b. Format your output for each new named entity using the appropriate XML tag based on its category:\\
      - For locations: <location>Full Name of Location</location>\\
      - For persons: <person>Full Name of Person</person>\\
      - For organizations: <organization>Full Name of Organization</organization>\\
      - For products: <product>Full Name of Product</product>\\
      - For art: <art>Full Name of Art Piece</art>\\
      - For events: <event>Full Name of Event</event>\\
      - For buildings: <building>Full Name of Building</building>\\
      - For miscellaneous: <miscellaneous>Full Name of Miscellaneous Entity</miscellaneous>\\
   c. If no new named entities are found after verification and redundancy removal, output <no\_new\_entities>No new named entities identified</no\_new\_entities>\\
\newline
5. Double-check miscellaneous\\
   a. If you classified any new entity as miscellaneous:\\
       - Compare the entity and how it is used in context with all other named entity types\\
       - Check if any of the other entity types fits this named entity (it does not need to fit 100%
       - If any of these entities seem applicable to the entity, list the named entity with the newly found entity and NOT as miscellaneous entity.\\
\newline
Sometimes the difference between the named entities is blurry. To help you make decisions in borderline cases follow these rules:\\
- Organization or Building: If in doubt, use "organization" instead of "building". Only use "building" if the name alone clearly identifies the entity as a building.  \\
\newline
Important notes:\\
- All entities in this task should be fictional and should be treated as distinct from any existing real-world or fictional entities. However, you must identify the names exactly as they appear in the OUTLINE, even if they are similar to real-world entities.\\
- Remember not to list abbreviations as separate entities. \\
- Include the names of any real-world entities from the outline.\\
- Only list NAMED entities. Do not create names for all entities that occur within the story.\\
- Make sure that you have an output for EACH NEW named entity that you have identified!\\
\newline
Use a <scratchpad> node within the <results> root node for all your reasoning. DO NOT include any XML tags in the reasoning process.\\
\newline
Present your final list of new named entities in this format:\\
\newline
<output>\\
(Your list of new named entities in XML format goes here)\\
</output>\\
\newline
If you need to think through your process or make notes, use UPPERCASE variable names for your thinking process. Your final output should only include the new named entities or the no\_new\_entities tag if applicable.\\
\newline
List all output including your thinking process within a single <results> node.\\
\newline
Before finalizing your output, double-check that you did not miss any named entities from the outline and that each named entity has a distinct full name.
\end{tcolorbox}

\subsection{Named Entity Name Generation Prompt}
\label{appendix:prompts:event:entity-gen}
\noindent
\footnotesize
\begin{tcolorbox}[
    colback=gray!15,
    colframe=gray!15,
    coltext=darkgray,
    boxrule=0pt,
    width=\textwidth,
    boxsep=5pt,
    sharp corners,
    breakable
]You are an AI assistant tasked with renaming a list of entities to ensure they are distinct from any known real-world or fictional names. This task is crucial for creating original content that doesn't infringe on existing intellectual property or cause confusion with real entities.\\
\newline
You will be provided with eight lists of names to be renamed: locations, organizations, persons, products, art, buildings, events, and miscellaneous items. These are presented in XML format as follows:\\
\newline
Entity names to change:\\
<entity\_names>\\
\{\{LOCATIONS\_NAME\_XML\}\}\\
\{\{ORGANIZATIONS\_NAME\_XML\}\}\\
\{\{PERSONS\_NAME\_XML\}\}\\
\{\{PRODUCTS\_NAME\_XML\}\}\\
\{\{ARTS\_NAME\_XML\}\}\\
\{\{BUILDINGS\_NAME\_XML\}\}\\
\{\{EVENTS\_NAME\_XML\}\}\\
\{\{MISCELLANEOUSS\_NAME\_XML\}\}\\
</entity\_names>\\
\newline
Follow these steps to complete the task:\\
\newline
1. For each entity in the lists, create a new name that is different from the original but maintains a similar style or feel. The new name must be fictional, but it must sound realistic. Avoid names that are clearly fictional.\\
\newline
2. Ensure that the new names are not associated with any known real-world or fictional entities. This includes names of people, places, organizations, products, artworks, buildings, events, or characters from books, movies, or other media.\\
\newline
3. When creating new names:\\
   - For locations: Maintain a geographical feel appropriate to the original name's region.\\
   - For organizations: Keep a professional or institutional tone similar to the original.\\
   - For persons: Preserve the cultural or ethnic flavor of the original name if applicable. If only a first name is provided, consider adding the lastname. If only a first name is provided, consider adding a last name. Ensure consistency in naming, particularly with inter-personal relations. For example, children should have the same last name as their parents, and married people often share the same last name.\\
   - For products: Retain a similar market appeal and product category feel.\\
   - For art: Maintain the artistic style or genre suggested by the original name.\\
   - For buildings: Keep architectural or functional implications of the original name.\\
   - For events: Preserve the nature or purpose of the event in the new name.\\
   - For miscellaneous items: Retain the essence or category of the original item.\\
\newline
4. Avoid using common words, phrases, or combinations that might accidentally reference existing entities.\\
\newline
5. For each renamed entity, provide both the new name and the old name.\\
\newline
6. Output your results in the following XML format:\\
\newline
<results>\\
  <location>\\
    <name>[New Location Name]</name>\\
    <old\_name>[Original Location Name]</old\_name>\\
  </location>\\
  <organization>\\
    <name>[New Organization Name]</name>\\
    <old\_name>[Original Organization Name]</old\_name>\\
  </organization>\\
  <person>\\
    <name>[New Person Name]</name>\\
    <old\_name>[Original Person Name]</old\_name>\\
  </person>\\
  <product>\\
    <name>[New Product Name]</name>\\
    <old\_name>[Original Product Name]</old\_name>\\
  </product>\\
  <art>\\
    <name>[New Art Name]</name>\\
    <old\_name>[Original Art Name]</old\_name>\\
  </art>\\
  <building>\\
    <name>[New Building Name]</name>\\
    <old\_name>[Original Building Name]</old\_name>\\
  </building>\\
  <event>\\
    <name>[New Event Name]</name>\\
    <old\_name>[Original Event Name]</old\_name>\\
  </event>\\
  <miscellaneous>\\
    <name>[New Miscellaneous Item Name]</name>\\
    <old\_name>[Original Miscellaneous Item Name]</old\_name>\\
  </miscellaneous>\\
</results>\\
\newline
Here's an example of how your output should look:\\
\newline
<results>\\
  <location>\\
    <name>Emerald Heights</name>\\
    <old\_name>Green Valley</old\_name>\\
  </location>\\
  <organization>\\
    <name>Quantum Dynamics Corporation</name>\\
    <old\_name>Particle Physics Institute</old\_name>\\
  </organization>\\
  <person>\\
    <name>Zara Blackwood</name>\\
    <old\_name>Emma Stone</old\_name>\\
  </person>\\
  <product>\\
    <name>SonicWave X1</name>\\
    <old\_name>iPod</old\_name>\\
  </product>\\
  <art>\\
    <name>Whispers of Eternity</name>\\
    <old\_name>Starry Night</old\_name>\\
  </art>\\
  <building>\\
    <name>Pinnacle Tower</name>\\
    <old\_name>Empire State Building</old\_name>\\
  </building>\\
  <event>\\
    <name>Global Harmony Festival</name>\\
    <old\_name>Woodstock</old\_name>\\
  </event>\\
  <miscellaneous>\\
    <name>LumiGlow Orb</name>\\
    <old\_name>Lava Lamp</old\_name>\\
  </miscellaneous>\\
</results>\\
\newline
Remember to create unique names for each entity and ensure they don't match any known entities. Provide your complete list of renamed entities in the specified XML format. When coming up with new names, try to keep them sounding realistic and appropriate to their category (location, organization, person, product, art, building, event, or miscellaneous).\\
\newline
Begin your renaming process now, and present your results in the format specified above. You may use a <scratchpad> section within the <results> tags to show your thinking process if needed. All actual results should be listed in <location>, <organization>, <person>, <product>, <art>, <building>, <event>, or <miscellaneous> nodes within the <results> node.\\
\newline
Do not include any content outside of the <results> tags in your response.\\
\newline
ONLY change the names of the entities listed under entity names to change. Do not invent any other entities. Do not change the names of any other entities.
\end{tcolorbox}

\subsection{Event Outline Adjustment with new Named Entity Names}
\label{appendix:prompts:event:outline-adjustment}
\noindent
\footnotesize
\begin{tcolorbox}[
    colback=gray!15,
    colframe=gray!15,
    coltext=darkgray,
    boxrule=0pt,
    width=\textwidth,
    boxsep=5pt,
    sharp corners,
    breakable
]You are an AI assistant tasked with updating an OUTLINE to be consistent with new entity names. Your goal is to make minimal changes while ensuring all entity names are updated correctly. Follow these instructions carefully:\\
\newline
First, here are the entities for which the names have been changed:\\
\newline
<entities>\\
<adjusted\_locations>\\
\{\{ADJUSTED\_LOCATIONS\_XML\}\}\\
</adjusted\_locations>\\
<adjusted\_persons>\\
\{\{ADJUSTED\_PERSONS\_XML\}\}\\
</adjusted\_persons>\\
<adjusted\_organizations>\\
\{\{ADJUSTED\_ORGANIZATIONS\_XML\}\}\\
</adjusted\_organizations>\\
<adjusted\_products>\\
\{\{ADJUSTED\_PRODUCTS\_XML\}\}\\
</adjusted\_products>\\
<adjusted\_arts>\\
\{\{ADJUSTED\_ARTS\_XML\}\}\\
</adjusted\_arts>\\
<adjusted\_buildings>\\
\{\{ADJUSTED\_BUILDINGS\_XML\}\}\\
</adjusted\_buildings>\\
<adjusted\_events>\\
\{\{ADJUSTED\_EVENTS\_XML\}\}\\
</adjusted\_events>\\
<adjusted\_miscellaneouss>\\
\{\{ADJUSTED\_MISCELLANEOUSS\_XML\}\}\\
</adjusted\_miscellaneouss>\\
</entities>\\
\newline
Your task is to update the OUTLINE to be consistent with the new names of these entities. Follow these instructions carefully:\\
\newline
1. Make minimal changes to the outline. Only update the names of entities that have been changed.\\
2. Apply changes on each sentence individually.\\
3. Output each updated sentence as a separate <storyitem>.\\
4. If a sentence does not contain any entities that need to be changed, output it as is.\\
5. Ensure that you maintain the original structure and content of the OUTLINE, changing only the necessary entity names.\\
6. Always use the full name as defined by the "name" property of the entities.\\
\newline
Output format:\\
Place all your outputs in a root node <results>. Do not output any content outside of this root node. Each sentence should be in its own <storyitem> tag. Your output should look like this:\\
\newline
<results>\\
<storyitem>[Updated sentence 1]</storyitem>\\
<storyitem>[Updated sentence 2]</storyitem>\\
...\\
<storyitem>[Updated sentence n]</storyitem>\\
</results>\\
\newline
Important reminders:\\
- Make only the necessary changes to reflect the new entity names while preserving the original meaning and structure of each sentence.\\
- Ensure that all entity name changes are consistent with the provided data for locations, persons, organizations, products, art, buildings, events, and miscellaneous items.\\
- Double-check your work to make sure you haven't missed any entity name changes or accidentally modified any content that should remain unchanged.\\
- Change a name only if it refers to the specific named entity being updated. Do not change the name if it refers to a different entity, even if their names partially overlap.\\
\newline
Now, here is the OUTLINE to update:\\
\newline
Date: \{DATE\}\\
<outline>\\
\{OUTLINE\}\\
</outline>\\
\newline
Process each sentence in the OUTLINE, updating entity names as necessary, and output the results as instructed above.\\
\newline
DO NOT include the date as a storyitem.\\
Ensure that you DO NOT change the names of any previously identified entities whose names did not require modification.
\end{tcolorbox}

\subsection{Event Outline (all) Named Entity Recognition Prompt}
\label{appendix:prompts:event:outline-new-all}
\noindent
\footnotesize
\begin{tcolorbox}[
    colback=gray!15,
    colframe=gray!15,
    coltext=darkgray,
    boxrule=0pt,
    width=\textwidth,
    boxsep=5pt,
    sharp corners,
    breakable
]You are an AI assistant tasked with identifying which of the provided entities are explicitly named within the provided outline. Follow these instructions carefully:\\
\newline
1. The date for this fictional event outline is:\\
<date>\{\{DATE\}\}</date>\\
\newline
2. Here is the outline you need to analyze:\\
<outline>\\
\{\{OUTLINE\}\}\\
</outline>\\
\newline
3. Here is the list of known (fictional) named entities:\\
<entities>\\
\{\{LOCATIONS\_XML\}\}\\
\{\{PERSONS\_XML\}\}\\
\{\{ORGANIZATIONS\_XML\}\}\\
\{\{PRODUCTS\_XML\}\}\\
\{\{ARTS\_XML\}\}\\
\{\{EVENTS\_XML\}\}\\
\{\{BUILDINGS\_XML\}\}\\
\{\{MISCELLANEOUSS\_XML\}\}\\
\newline
\{\{ADJUSTED\_LOCATIONS\_XML\}\}\\
\{\{ADJUSTED\_PERSONS\_XML\}\}\\
\{\{ADJUSTED\_ORGANIZATIONS\_XML\}\}\\
\{\{ADJUSTED\_PRODUCTS\_XML\}\}\\
\{\{ADJUSTED\_ARTS\_XML\}\}\\
\{\{ADJUSTED\_BUILDINGS\_XML\}\}\\
\{\{ADJUSTED\_EVENTS\_XML\}\}\\
\{\{ADJUSTED\_MISCELLANEOUSS\_XML\}\}\\
</entities>\\
\newline
4. Your task is to go through all of the named entities in the provided list. For each named entity, check if it is explicitly referred to by name in the outline. If a named entity is explicitly mentioned by name, include it in your results.\\
\newline
5. For each entity you identify, list them using the following format:\\
<[entity\_type]><id>[id of the entity]</id><name>[full name of the entity]</name></[entity\_type]>\\
\newline
6. Present your final output within a single <results> root node, structured as follows:\\
\newline
<results>\\
<entities>\\
(List all identified entities here as described in step 5)\\
</entities>\\
</results>\\
\newline
7. Double-check your work: For each entity that you have identified, make sure to find the sentence within the outline that explicitly refers to this identified entity by name. If you cannot find that the entity is mentioned explicitly by name in the outline, do not include it in your results.\\
\newline
8. You can use a <scratchpad> for your thinking process. Do not include any XML tags within your scratchpad.\\
\newline
9. Make sure that you have not missed any named entity from the outline that was also provided to you in the list of named entities.\\
\newline
Remember, only include entities that are explicitly mentioned by name in the outline. Do not infer or assume the presence of entities that are not directly named.
\end{tcolorbox}

\subsection{Event Outline Named Entity Resolution Prompt}
\label{appendix:prompts:event:outline-ne-resolution}
\noindent
\footnotesize
\begin{tcolorbox}[
    colback=gray!15,
    colframe=gray!15,
    coltext=darkgray,
    boxrule=0pt,
    width=\textwidth,
    boxsep=5pt,
    sharp corners,
    breakable
]You are an AI assistant tasked with identifying and marking fictional entities within an outline of a fictional event. Your goal is to replace each occurrence of these entities with a specific format that includes the entity's ID. Follow these instructions carefully:\\
\newline
You will be working with the following inputs:\\
\newline
Date: \{\{DATE\}\}\\
\newline
<outline>\\
\{\{OUTLINE\}\}\\
</outline>\\
\newline
<entities>\\
\{\{ALL\_USED\_LOCATION\_XML\}\}\\
\{\{ALL\_USED\_PERSON\_XML\}\}\\
\{\{ALL\_USED\_ORGANIZATION\_XML\}\}\\
\{\{ALL\_USED\_PRODUCT\_XML\}\}\\
\{\{ALL\_USED\_ART\_XML\}\}\\
\{\{ALL\_USED\_BUILDING\_XML\}\}\\
\{\{ALL\_USED\_EVENT\_XML\}\}\\
\{\{ALL\_USED\_MISCELLANEOUS\_XML\}\}\\
</entities>\\
\newline
Process the outline by following these steps:\\
\newline
1. Carefully review the list of entities provided in the <entities> section. Each entity will have an associated ID and a name.\\
\newline
2. Search the outline for all occurrences of each entity in the list.\\
\newline
3. For each occurrence found, replace it with the format: \{full name|ID\}\\
   Where "full name" is the full name as provided via the "name" property of the entity.\\
\newline
4. Maintain the original structure and formatting of the outline, only changing the entities as described.\\
\newline
5. After processing all entities, review the entire outline to ensure all occurrences have been properly marked and no entities were missed.\\
\newline
6. Output the processed outline, maintaining its original structure but with all entity occurrences replaced as instructed.\\
\newline
Important points to remember:\\
- Be thorough in your search for entities, including variations or partial mentions.\\
- Always use the ID and name as provided in the entity list, even:\\
    - If an entity is referred to by full name\\
    - If an entity is referred to by an abbreviation\\
    - If an entity is referred to using parts of the full name\\
\newline
Format your output as follows:\\
- Enclose the entire processed outline within <outline> tags.\\
- Place each sentence of the outline within separate <storyitem> tags.\\
\newline
Provide your final output without any additional commentary or explanations. Focus solely on processing the outline as instructed.
\end{tcolorbox}

\subsection{Named Entity KB Entry Generation Prompt}
\label{appendix:prompts:event:kb-generation}
\noindent
\footnotesize
\begin{tcolorbox}[
    colback=gray!15,
    colframe=gray!15,
    coltext=darkgray,
    boxrule=0pt,
    width=\textwidth,
    boxsep=5pt,
    sharp corners,
    breakable
]You are an AI assistant tasked with creating fictional entities based on provided information. \\
Your goal is to generate detailed, coherent, and realistic descriptions for new locations, persons, organizations, products, art, buildings, events, and miscellaneous entities. \\
Follow these instructions carefully:\\
\newline
1. Review the existing entities (if provided):\\
\newline
<existing\_entities>\\
\{\{LOCATIONS\_XML\}\}\\
\{\{PERSONS\_XML\}\}\\
\{\{ORGANIZATIONS\_XML\}\}\\
\{\{PRODUCTS\_XML\}\}\\
\{\{ARTS\_XML\}\}\\
\{\{EVENTS\_XML\}\}\\
\{\{BUILDINGS\_XML\}\}\\
\{\{MISCELLANEOUSS\_XML\}\}\\
</existing\_entities>\\
\newline
2. Review the names of new entities to be created:\\
<new\_entity\_names>\\
\{\{USED\_NEW-LOCATIONS\_XML\}\}\\
\{\{USED\_NEW-PERSONS\_XML\}\}\\
\{\{USED\_NEW-ORGANIZATIONS\_XML\}\}\\
\{\{USED\_NEW-PRODUCTS\_XML\}\}\\
\{\{USED\_NEW-ARTS\_XML\}\}\\
\{\{USED\_NEW-EVENTS\_XML\}\}\\
\{\{USED\_NEW-BUILDINGS\_XML\}\}\\
\{\{USED\_NEW-MISCELLANEOUSS\_XML\}\}\\
</new\_entity\_names>\\
\newline
3. Carefully read the provided outline:\\
\newline
Date: \{\{DATE\}\}\\
<outline>\\
\{\{OUTLINE\}\}\\
</outline>\\
\newline
Based on the information provided, create detailed descriptions for each new entity following these guidelines:\\
\newline
General instructions:\\
- Ensure all created entities are entirely fictional and not similar to any real or known fictional entities.\\
- Maintain realism and coherence with the provided outline and other entities.\\
- Create a believable and consistent fictional world that aligns with the context of the outline.\\
- For entity descriptions, focus on providing a solid background that remains valid throughout the story, rather than basing it centrally on the outline itself.\\
- Develop well-rounded entities with backgrounds and characteristics that can support various potential story developments beyond the specific outline provided.\\
- Strictly derive all entities from the outline. Do not invent entities that are not mentioned in the outline.\\
- Do not alter the name of the entities.\\
- Some properties (e.g., place, city, country, spouse, architect, country, nationality) must be filled with named entities. Make sure to use FICTIONAL named entities ( fictional places, cities, countries, spouse names, architects, countries, nationalities, etc.). Check if you should use one of the existing fictional named entities (from <new\_entity\_names> and <existing\_entities>) or create a fictional name instead. DO NOT say that any of the entities or properties are fictional. It is important that everything seems realistic!\\
- Avoid exaggerating the named entities. They can be ordinary and don’t need to be world-class or state-of-the-art.\\
- Ensure the details of the named entities are realistic. If global impact isn’t necessary, adjust the details to be more modest and appropriate.\\
\newline
Specific instructions for each entity type:\\
\newline
1. For each new location:\\
   - Use the provided name and ID\\
   - Determine an appropriate type (city, village, country, region, etc.)\\
   - Write a concise single-sentence description that:\\
     * Does not refer to any other entities\\
     * Provides only background information, not related to the event in the outline\\
     * Is general and serves as background for this event\\
   - Include at least five additional properties from: population, area, founded, climate, elevation, country\\
   - Format the output as follows:\\
     <location>\\
     <id>[Provided ID]</id>\\
     <name>[Fictional name]</name>\\
     <type>[city/village/country/region]</type>\\
     <description>[One-sentence concise description]</description>\\
     (Include at least 5 of: population, area, founded, climate, elevation, country)\\
     </location>\\
\newline
2. For each new person:\\
   - Use the provided name and ID\\
   - Create fictional details for: date\_of\_birth, gender, profession, nationality, education\\
   - Write a concise single-sentence description that:\\
     * Focuses on background, personality, and motivations\\
     * Does not refer to any other entities\\
     * Provides only background information, not related to the event in the outline\\
     * Is general and serves as background for this event\\
   - Include at least five additional properties from: height, weight, eye\_color, hair\_color, political\_affiliation, marital\_status, spouse\\
   - Format the output as follows:\\
     <person>\\
     <id>[Provided ID]</id>\\
     <name>[Full name]</name>\\
     <date\_of\_birth>[Date]</date\_of\_birth>\\
     <gender>[Gender]</gender>\\
     <profession>[Job title]</profession>\\
     <nationality>[Country]</nationality>\\
     <education>[Highest level of education]</education>\\
     <description>[One-sentence concise description]</description>\\
     (Include at least 5 of: height, weight, eye\_color, hair\_color, political\_affiliation, marital\_status)\\
     </person>\\
\newline
3. For each new organization:\\
   - Use the provided name and ID\\
   - Determine an appropriate type (company, non-profit, educational institution, government agency, etc.)\\
   - Write a concise single-sentence description that:\\
     * Does not refer to any other entities\\
     * Provides only background information, not related to the event in the outline\\
     * Is general and serves as background for this event\\
   - Include at least five additional properties from: founded, headquarters, industry, mission\_statement, number\_of\_employees, annual\_revenue\\
   - Format the output as follows:\\
     <organization>\\
     <id>[Provided ID]</id>\\
     <name>[Fictional name]</name>\\
     <type>[company/non-profit/educational institution/government agency]</type>\\
     <description>[One-sentence concise description]</description>\\
     (Include at least 5 of: founded, headquarters, industry, mission\_statement, number\_of\_employees, annual\_revenue)\\
     </organization>\\
\newline
4. For each new product:\\
   - Use the provided name and ID\\
   - Determine an appropriate type (consumer good, software, service, etc.)\\
   - Write a concise single-sentence description that:\\
     * Does not refer to any other entities\\
     * Provides only background information, not related to the event in the outline\\
     * Is general and serves as background for this event\\
   - Include at least five additional properties from: manufacturer, release\_date, price, weight, warranty\\
   - Format the output as follows:\\
     <product>\\
     <id>[Provided ID]</id>\\
     <name>[Fictional name]</name>\\
     <type>[consumer good/software/service]</type>\\
     <description>[One-sentence concise description]</description>\\
     (Include at least 5 of: manufacturer, release\_date, price, weight, warranty)\\
     </product>\\
\newline
5. For each new art piece:\\
   - Use the provided name and ID\\
   - Determine an appropriate type (painting, sculpture, novel, film, etc.)\\
   - Write a concise single-sentence description that:\\
     * Does not refer to any other entities\\
     * Provides only background information, not related to the event in the outline\\
     * Is general and serves as background for this event\\
   - Include at least five additional properties from: creator, year\_created, current\_location\_country, current\_location\_city, current\_location\_place\\
   - Format the output as follows:\\
     <art>\\
     <id>[Provided ID]</id>\\
     <name>[Fictional name]</name>\\
     <type>[painting/sculpture/novel/film]</type>\\
     <description>[One-sentence concise description]</description>\\
     (Include at least 5 of: creator, year\_created, current\_location\_country, current\_location\_city, current\_location\_place)\\
     </art>\\
\newline
6. For each new building:\\
   - Use the provided name and ID\\
   - Determine an appropriate type (residential, commercial, public, etc.)\\
   - Write a concise single-sentence description that:\\
     * Does not refer to any other entities\\
     * Provides only background information, not related to the event in the outline\\
     * Is general and serves as background for this event\\
   - Include at least five additional properties from: place, city, country, architect, year\_built, height, floors, material, capacity\\
   - Format the output as follows:\\
     <building>\\
     <id>[Provided ID]</id>\\
     <name>[Fictional name]</name>\\
     <type>[residential/commercial/public]</type>\\
     <description>[One-sentence concise description]</description>\\
     (Include at least 5 of: place, city, country, architect, year\_built, height, floors, material, capacity)\\
     </building>\\
\newline
7. For each new event:\\
   - Use the provided name and ID\\
   - Determine an appropriate type (historical, cultural, sporting, etc.)\\
   - Write a concise single-sentence description that:\\
     * Does not refer to any other entities\\
     * Provides only background information, not related to the event in the outline\\
     * Is general and serves as background for this event\\
   - Include at least five additional properties from: date, place, city, country, duration, organizer, number\_of\_participants, budget\\
   - Format the output as follows:\\
     <event>\\
     <id>[Provided ID]</id>\\
     <name>[Fictional name]</name>\\
     <type>[historical/cultural/sporting]</type>\\
     <description>[One-sentence concise description]</description>\\
     (Include at least 5 of: date, place, city, country, duration, organizer, number\_of\_participants, budget)\\
     </event>\\
\newline
8. For each new miscellaneous entity:\\
   - Use the provided name and ID\\
   - Determine an appropriate type (concept, theory, phenomenon, etc.)\\
   - Write a concise single-sentence description that:\\
     * Does not refer to any other entities\\
     * Provides only background information, not related to the event in the outline\\
     * Is general and serves as background for this event\\
   - Format the output as follows:\\
     <miscellaneous>\\
     <id>[Provided ID]</id>\\
     <name>[Fictional name]</name>\\
     <type>[concept/theory/phenomenon]</type>\\
     <description>[One-sentence concise description]</description>\\
     </miscellaneous>\\
\newline
Present your final output in the following format:\\
<results>\\
(Insert all created entities here, grouped by type)\\
</results>\\
\newline
Remember to create detailed, coherent, and realistic descriptions for each entity while adhering to the guidelines provided. Ensure that all descriptions are single, short, and concise sentences that do not refer to other entities and only discuss background information unrelated to the event described in the outline.
\end{tcolorbox}

\subsection{Named Entity Update Prompt}
\label{appendix:prompts:event:kb-update}
\noindent
\footnotesize
\begin{tcolorbox}[
    colback=gray!15,
    colframe=gray!15,
    coltext=darkgray,
    boxrule=0pt,
    width=\textwidth,
    boxsep=5pt,
    sharp corners,
    breakable
]You are an AI assistant tasked with updating a list of fictional entities based on an outline of events. Follow these instructions carefully to complete the task:\\
\newline
1. First, read the outline of the fictional event:\\
<outline>\\
\{\{OUTLINE\}\}\\
</outline>\\
\newline
2. Next, review the list of fictional entities:\\
<entities>\\
\{\{USED\_LOCATION\_XML\}\}\\
\{\{USED\_PERSON\_XML\}\}\\
\{\{USED\_ORGANIZATION\_XML\}\}\\
\{\{USED\_PRODUCT\_XML\}\}\\
\{\{USED\_ART\_XML\}\}\\
\{\{USED\_EVENT\_XML\}\}\\
\{\{USED\_BUILDING\_XML\}\}\\
\{\{USED\_MISCELLANEOUS\_XML\}\}\\
</entities>\\
\newline
3. For each entity in the list, follow these steps:\\
   a. Identify the entity's role in the outline. Create an update sentence describing how the entity was affected by or involved in the events described in the outline.\\
   b. Review all properties of the entity EXCEPT for "id", "created\_at", "history", and "entity\_class".\\
   c. For each property:\\
      - If the property does not need to be updated based on the outline, leave it as-is.\\
      - If the events or the time difference since the last update (last\_updated) indicate that the value has changed, update it accordingly.\\
      - Ensure the new value is consistent with the outline, other entities, and plausible given the time difference.\\
   d. Do not alter the "description" field unless the current description is no longer valid after the events in the outline.\\
\newline
4. Output your results for each entity in the following format:\\
<results>\\
<[entity\_type]>\\
<entity\_id>[Insert entity id here]</entity\_id>\\
<update>[Insert your update sentence here]</update>\\
(List any properties that were changed, with their new values, using the format:)\\
<[property\_name]>[new value]</[property\_name]>\\
</[entity\_type]>\\
(Repeat for each entity)\\
</results>\\
\newline
5. Important reminders:\\
   - Stick to the information provided in the outline and entities list. Do not invent new details or events.\\
   - Ensure all updates and changes are consistent with the outline and with each other.\\
   - Be concise in your updates, focusing only on relevant changes.\\
   - If no properties need to be changed for an entity, do not include any property tags.\\
   - Process each entity in the order they are presented in the list.\\
   - Use the appropriate entity type tag (e.g., <location>, <person>, etc.) instead of <entity\_update>.\\
   - Begin your response with the first entity update immediately, without any preamble.\\
\newline
All actual results should be listed in <location>, <organization>, <person>, <product>, <art>, <building>, <event>, or <miscellaneous> nodes within the <results> node.\\
Do not include any content outside of the <results> tags in your response.\\
\newline
Start processing the entities now, following the instructions and format provided above.
\end{tcolorbox}

\subsection{Diverse Next Summary Generation Prompt}
\label{appendix:prompts:event:summary-diverse}
\noindent
\footnotesize
\begin{tcolorbox}[
    colback=gray!15,
    colframe=gray!15,
    coltext=darkgray,
    boxrule=0pt,
    width=\textwidth,
    boxsep=5pt,
    sharp corners,
    breakable
]You are an AI tasked with creating fictional future news summaries based on provided information. Your goal is to generate plausible continuations of an existing narrative. Follow these instructions carefully:\\
\newline
1. First, you will be given known entities in this fictional world. These will be provided in the following format:\\
<known\_entities>\\
\{\{LOCATIONS\_XML\}\}\\
\{\{PERSONS\_XML\}\}\\
\{\{ORGANIZATIONS\_XML\}\}\\
\{\{PRODUCTS\_XML\}\}\\
\{\{ARTS\_XML\}\}\\
\{\{EVENTS\_XML\}\}\\
\{\{BUILDINGS\_XML\}\}\\
\{\{MISCELLANEOUSS\_XML\}\}\\
</known\_entities>\\
\newline
2. Next, you will be provided with the history of events that have already occurred in this fictional world:\\
\newline
<history>\\
\{\{HISTORY\_XML\}\}\\
</history>\\
\newline
3. Your task is to create \{\{NUMBER\_SUMMARIES\}\} new summaries that describe future fictional events following the last event from the <history>. These new summaries should be consistent with the existing story and represent plausible continuations or developments of the original narrative.\\
\newline
4. For each summary, create:\\
   a. A summary text (a single concise sentence)\\
   b. The date on which this fictional next event happens\\
\newline
5. Before starting, check if the history of events indicates specific dates for followup events. Ensure your continuations are consistent with these expected followup events. All your summaries must either concern this event, or happen before this event.\\
\newline
6. Make sure that each summary you generate focuses on at least one of the main named entities from the history of events.\\
\newline
7. Follow these guidelines when creating your summaries:\\
   a. Ensure all summaries are fictional and not based on real events or real people.\\
   b. Make the summaries sound realistic and plausible as follow-up stories to the previous outlines. \\
   c. Think about plausible next events based on the fictional named entities, the history of the past fictional events and the genre \{\{GENRE\}\}. \\
   d. Create summaries that are unbiased and objective in tone.\\
   e. Each summary MUST BE ONLY A SINGLE concise sentence.\\
   f. Summaries may focus on different personas or organizations from the provided lists.\\
   g. Ensure a balance of positive and negative news stories, developments, and alternative scenarios.\\
   h. Consider various dimensions or personas that could be varied when generating diverse summaries.\\
\newline
Try to be diverse in the summaries you generate. Consider different plausible substories and vary:\\
- between impact: Try to create various low-impact next events, but you may also at times mix in an event with a slightly higher impact.\\
- between directions: Vary between positive and negative story developments. Think about how stories in the \{\{GENRE\}\} genre progress in the real-world, NOT in a novel. Provide various realistic alternatives for how the story may progress in either direction.\\
- between different key named entities of the fictional story.\\
\newline
8. Output format:\\
   - Enclose each summary in <summary> tags.\\
   - Each summary must have two child properties:\\
     <text>[The generated summary]</text>\\
     <date>[The date for the next event]</date>\\
   - Before each summary, explain your thought process in <thought\_process> tags. Make sure to identify all known followup events based on the provided history first, and verify that your continuations are consistent with these known followup events regarding the date.\\
   - Output everything within a <results> root node.\\
\newline
9. Special instructions:\\
   - The history and entities have special formatting. They sometimes include statements like \{[PHRASE]|[ID]\}. Read it as [PHRASE], while the [ID] specifies the ID of the linked entity.\\
   - Example: "I met \{Boris Bowman|PERSON-1\} yesterday." Read as: "I met Boris Bowman yesterday." (The ID of Boris Bowman is PERSON-1).\\
   - DO NOT use the \{[PHRASE]|[ID]\} formulation when generating new summaries.\\
   - Do not exaggerate the summaries. Avoid using words like "groundbreaking", "worldwide", "global". Keep the summaries realistic.\\
   - Do not create summaries with global or national impact unless the genre specifically requires it. Instead, focus on smaller or local developments.\\
   - Do not focus on technological discoveries or topics like AI tools, virtual reality, augmented reality, 3D-modelling, quantum computing, etc. You may include such topics only if they are HIGHLY relevant to the genre \{\{GENRE\}\} AND the provided history of events.\\
   - Focus on realistic, meaningful summaries with specific details and developments that align with typical, realistic scenarios of the genre \{\{GENRE\}\}.\\
\newline
10. Final reminders:\\
   - Repeat this process for all \{\{NUMBER\_SUMMARIES\}\} summaries.\\
   - Ensure that each summary explores a different aspect or potential next step of the fictional situation presented in the HISTORY.\\
   - Each summary text MUST BE ONLY A SINGLE concise sentence.\\
   - The continuations should be diverse in terms of high (at most one), medium and low impact, positive and negative story developments, etc., and can cover different alternatives of how the future event can play out.\\
   - Do not enumerate over the summaries.\\
   - When brainstorming future event summaries, carefully consider whether each continuation aligns with the provided history of events, provided named entities, and the genre \{\{GENRE\}\}. Before creating summaries with global or large-scale impact, double-check if such developments seem plausible based on how real-world events of this type would typically unfold. Focus on what fits the genre and provided history, ensuring that every dimension feels realistic and consistent with the context.\\
\newline
Remember to maintain the desired format and brevity of the event summaries while creating plausible and engaging continuations of the narrative.\\
\newline
Here's an example of the desired output format:\\
\newline
<results>\\
<thought\_process>\\
(This includes the thought process for all summaries.)\\
</thought\_process>\\
\newline
<summaries>\\
<summary>\\
<text>[Your first summary text here]</text>\\
<date>[Date for the first summary]</date>\\
</summary>\\
\newline
<summary>\\
<text>[Your second summary text here]</text>\\
<date>[Date for the second summary]</date>\\
</summary>\\
\newline
(... continue for all summaries ...)\\
</summaries>\\
</results>
\end{tcolorbox}

\subsection{Mutually Exclusive Summary Generation Prompt}
\label{appendix:prompts:event:summary-mutually}
\noindent
\footnotesize
\begin{tcolorbox}[
    colback=gray!15,
    colframe=gray!15,
    coltext=darkgray,
    boxrule=0pt,
    width=\textwidth,
    boxsep=5pt,
    sharp corners,
    breakable
]You are an AI tasked with creating alternative fictional future news summaries based on provided information. \\
Your goal is to generate summaries of plausible alternative continuations of an existing narrative. Follow these instructions carefully:\\
\newline
1. First, you will be given known entities in this fictional world:\\
\newline
<known\_entities>\\
\{\{LOCATIONS\_XML\}\}\\
\{\{PERSONS\_XML\}\}\\
\{\{ORGANIZATIONS\_XML\}\}\\
\{\{PRODUCTS\_XML\}\}\\
\{\{ARTS\_XML\}\}\\
\{\{EVENTS\_XML\}\}\\
\{\{BUILDINGS\_XML\}\}\\
\{\{MISCELLANEOUSS\_XML\}\}\\
</known\_entities>\\
\newline
2. Next, you will be provided with the history of events that have already occurred in this fictional world:\\
\newline
<history>\\
\{\{HISTORY\_XML\}\}\\
</history>\\
\newline
3. You will be given the following plausible summary of how the fictional event evolves:\\
<continuation\_summary>\\
<text>\{\{CONTINUATION\_TEXT\}\}</text>\\
<date>\{\{CONTINUATION\_DATE\}\}</date>\\
</continuation\_summary>\\
\newline
4. Your task is to create \{\{NUM\_ALTERNATIVES\}\} contradictory alternative summaries of how the fictional event can progress based on the provided continuation summary. \\
Each of these continuation summaries must make subtle changes to the continuation summary such that they are contradictory alternatives to one another.\\
This means, if the story evolves with one of the continuation summaries, the other ones cannot happen anymore.\\
\newline
5. When changing the provided continuation summary, maintain these key properties:\\
   a) The central topic and involved main entity\\
   b) The stance (whether this is a positive, neutral or negative story evolvement)\\
   c) The impact (whether this is a high impact, medium impact or low impact evolvement)\\
   d) The same date\\
\newline
6. Guidelines for creating alternative summaries:\\
   - Ensure each new continuation summary is consistent with the existing story and represents a plausible continuation or development of the original narrative.\\
   - Write each continuation summary as a single concise sentence in an objective tone.\\
   - Make sure your continuations are consistent with any known followup events regarding the date.\\
\newline
7. Output format:\\
   - Enclose your entire response in <results> tags.\\
   - Before the summaries, explain your thought process in <thought\_process> tags. Make sure to identify all known followup events based on the provided history first, and verify that your continuations are consistent with these known followup events regarding the date.\\
   - Enclose all summaries in <summaries> tags.\\
   - For each summary:\\
     <summary>\\
       <text>[The generated summary]</text>\\
       <date>[The date for the next event]</date>\\
     </summary>\\
\newline
8. Special instructions:\\
   - The history and entities have special formatting. They sometimes include statements like \{[PHRASE]|[ID]\}. Read it as [PHRASE], while the [ID] specifies the ID of the linked entity.\\
   - Example: "I met \{Boris Bowman|PERSON-1\} yesterday." Read as: "I met Boris Bowman yesterday." (The ID of Boris Bowman is PERSON-1).\\
   - DO NOT use the \{[PHRASE]|[ID]\} formulation when generating new summaries.\\
   - When brainstorming alternative event summaries, carefully consider whether each continuation aligns with the provided history of events, provided named entities, and the genre \{\{GENRE\}\}. Before creating summaries with global or large-scale impact, double-check if such developments seem plausible based on how real-world events of this type would typically unfold. Focus on what fits the genre and provided history, ensuring that every dimension feels realistic and consistent with the context.\\
   - Do not exaggerate the summaries. Avoid making summaries "groundbreaking." Keep the summaries realistic.\\
   - Do not create summaries with global or national impact unless the genre specifically requires it. Instead, focus on smaller or local developments.\\
   - Do not focus on technological discoveries or topics like AI tools, virtual reality, quantum computing, etc. You may include such topics only if they are HIGHLY relevant to the genre \{\{GENRE\}\} AND the provided history of events.\\
   - Focus on realistic, meaningful summaries with specific details and developments that align with typical, realistic scenarios of the genre \{\{GENRE\}\}.\\
\newline
9. Here's an example of the desired output format:\\
\newline
<results>\\
<thought\_process>\\
(Your explanation of the thought process, including identification of known followup events and verification of consistency)\\
</thought\_process>\\
<summaries>\\
<summary>\\
<text>[First alternative summary]</text>\\
<date>[Date of the event]</date>\\
</summary>\\
<summary>\\
<text>[Second alternative summary]</text>\\
<date>[Date of the event]</date>\\
</summary>\\
(Additional summaries as needed)\\
</summaries>\\
</results>\\
\newline
Remember to maintain the desired format and brevity of the event continuations while creating plausible and engaging continuations of the narrative. Begin your response with the <results> tag and end it with the </results> tag.
\end{tcolorbox}

\subsection{Sentence Selection Prompt for Time-span Questions}
\label{appendix:prompts:question-gen:timespan:selection}
\noindent
\footnotesize
\begin{tcolorbox}[
    colback=gray!15,
    colframe=gray!15,
    coltext=darkgray,
    boxrule=0pt,
    width=\textwidth,
    boxsep=5pt,
    sharp corners,
    breakable
]You are an AI assistant tasked with analyzing fictional event descriptions and identifying sentences that can be used to derive time spans. Your goal is to find up to \{\{NUM\_SPANS\}\} tuples, each comprising \{\{NUM\_SENTENCES\}\} sentences, that can be used to compute meaningful time spans.\\
\newline
Here are the outlines of the events you will be analyzing:\\
\newline
<events>\\
\{\{OUTLINES\}\}\\
</events>\\
\newline
To identify specific points in time from different sentences, consider the following:\\
1. Sentences mentioning relative time differences (e.g., "six weeks ago", "in five days", "after two hours"). Try to determine the specific date or time they refer to based on the context.\\
2. Sentences mentioning a specific date or time (e.g., "10:00 am", "March 22nd").\\
3. Sentences containing specific details unique to the current event.\\
\newline
Your task is to select tuples of NUM\_SENTENCES sentences from which you can derive two specific points in time and compute a meaningful time span (duration). Follow these guidelines:\\
- The duration should be computed between comparable units.\\
- It must be possible to compute a specific time span between the two points in time.\\
- Prioritize sentences containing relative time differences for which you can compute a specific point in time.\\
- If possible, include sentences that mention a specific point in time explicitly.\\
- If neither of the above is possible, you may select sentences that rely only on the date of the events. However, it is preferred to collect sentences that introduce novel time information.\\
\newline
If multiple events are provided, the sentences must come from different events. If only one event is provided, the sentences can come from the same event.\\
Remember: It should be possible to compute a duration based on different specific points in time from the different selected sentence. DO NOT collect sentences based on which you can compare the ratio of different explicitly mentioned durations! Only select sentences based on which you can compute a meaningful absolute duration.\\
\newline
Use a <scratchpad> to plan your approach before outputting your final results. In your scratchpad, you can list potential sentence pairs, evaluate their suitability, and determine which ones to include in your final output.\\
\newline
For each tuple you identify, output the result in the following format:\\
\newline
<time-span>\\
<ids>[IDs of the selected sentences (comma separated)]</ids>\\
<explanation>[Explanation of how the time span can be computed from these sentences]</explanation>\\
</time-span>\\
\newline
Wrap all your results, including the scratchpad and time spans, in a <result> root node.\\
\newline
Example:\\
<result>\\
<scratchpad>\\
Potential pairs:\\
1. Sentences 2 and 5: Mention specific dates, can calculate exact duration.\\
2. Sentences 1 and 7: Relative time reference and specific time, need to infer exact start point.\\
3. Sentences 3 and 8: Both mention specific times on the same day, can calculate duration.\\
\newline
Decision: Include pairs 1 and 3 in the output.\\
</scratchpad>\\
\newline
<time-span>\\
<ids>2, 5</ids>\\
<explanation>Sentence 2 mentions the event starting on July 1st, while sentence 5 states it ended on July 15th. We can calculate a time span of 14 days between these two dates.</explanation>\\
</time-span>\\
\newline
<time-span>\\
<ids>3, 8</ids>\\
<explanation>Sentence 3 mentions a meeting at 9:00 AM, while sentence 8 refers to a conclusion at 4:30 PM on the same day. We can calculate a time span of 7 hours and 30 minutes between these two points.</explanation>\\
</time-span>\\
</result>\\
\newline
Remember to analyze the given event outlines carefully, identify suitable sentence pairs, and provide clear explanations for each time span you derive. If you cannot find the required number of valid time spans, include as many as you can identify from the given information.
\end{tcolorbox}

\subsection{Question Writing Prompt for Time-span Questions}
\label{appendix:prompts:question-gen:timespan:write}
\noindent
\footnotesize
\begin{tcolorbox}[
    colback=gray!15,
    colframe=gray!15,
    coltext=darkgray,
    boxrule=0pt,
    width=\textwidth,
    boxsep=5pt,
    sharp corners,
    breakable
]You are an AI assistant tasked with generating a complex question-answer pair based on a fictional event outline. Your goal is to create a question that asks about an absolute time span (duration) that can be computed using specific selected sentences from the outline.\\
\newline
First, review the complete storyline outline:\\
\newline
<complete\_outline>\\
\{\{STORYLINE\_OUTLINE\_TO\_DATE\}\}\\
</complete\_outline>\\
\newline
Now, focus on the selected sentences:\\
\newline
<selected\_sentences>\\
\{\{SELECTED\_SENTENCES\}\}\\
</selected\_sentences>\\
\newline
To create a time-span question based on the selected sentences, follow these steps:\\
\newline
1. Identify the absolute date or time within each of the selected sentences if possible. If a sentence doesn't provide additional time information, use the event date as time information. Remember that the dates of individual sub-events within each event may differ from the event date. To compute the specific points in time, check if the selected sentences specify any of the following:\\
   - Relative time information (e.g., "six weeks ago", "in five days", "after two hours") based on which you can compute a specific point in time\\
   - Absolute time information (e.g., "10:00 am", "March 22nd")\\
   - You can always assume that the date listed for each event is known. The event date is always the latest possible date based on everything that has happened within the event. \\
   - If the specific points in time cannot be asserted with certainty but only estimated based on some likelihood, specify assumptions under which you can derive absolute points in time\\
   - If the sentences describe durations, list the respective start and end dates/times together with the assumptions\\
\newline
2. Think about the different options of absolute points of time and which time span between them would be most challenging to compute, ensuring that all selected sentences are required to compute the duration.\\
\newline
3. Draft a specific question that asks for the time duration. This question can only be solved if one derives the absolute points in time from each of the selected sentences. The question must always ask about the duration between two points in time. Prefer time durations that differ from the absolute durations of the respective event (if multiple options of different possible ways to compute durations exist).\\
\newline
4. Formulate a precise and correct answer to the question, which must be an absolute duration. Only use the information from the selected sentences and make the question very specific to avoid ambiguities. In the scratchpad, include a derivation of the answer using the absolute dates you estimated before.\\
\newline
5. Ensure the validity of the questions:\\
   - Ensure that the question can ONLY be answered when ALL of the selected sentences are present.\\
   - Make sure that ALL of the selected sentences are needed to answer the question.\\
   - Make sure that no other information from the <complete\_outline> is sufficient to answer the question and that no other sentence from the <complete\_outline> can replace any of the selected sentence in answering the question.\\
   - Adjust the question accordingly to fulfill these criteria by making it more specific.\\
\newline
6. Avoid absolute dates or times in your question if possible:\\
   - If you needed to make an assumption about when something happened, try to explain it relative to other existing events or information from the <complete\_outline>. Try to replace explicitly mentioned dates by anchoring them into some content of the <complete\_outline> instead. If necessary, add the additional required sentence ID to your result and explain the unique information of this sentence that is necessary.\\
   - Make sure that temporal reasoning is still required in the revised question. Double-check if the temporal information is still essential to answer the question after resolving the absolute dates. Think about different options (like start and end dates) to make sure that one must apply temporal reasoning to answer the question.\\
\newline
7. If you must integrate additional information beyond what is specified within the selected sentences:\\
   - Identify the unique information you must add to the question within the provided <complete\_outline> and select the sentence communicating this unique information\\
   - Explain what kind of unique information from this sentence is needed and why\\
   - List the sentence ID of the newly selected sentence in your result\\
   - Rewrite your question accordingly and ensure that all points listed in step 5 are valid when adding the newly selected sentence\\
\newline
Provide your reasoning in the following format:\\
<scratchpad>\\
(Repeat for all sentence IDs)\\
(Sentence ID):\\
- [Describe the time information from this sentence]\\
- [Identify the event date]\\
Assumptions:\\
- [Make your assumptions explicit]\\
Absolute dates:\\
- [List all relevant absolute dates from this sentence. Justify how they relate to the event date or other date information.]\\
\newline
\newline
Additional sentences:\\
(Think about additional sentence IDs to add and explain your decision and what information from the sentence you require. List any new absolute dates based on this information.)\\
\newline
Question Idea:\\
(Outline how to create a challenging time-span question based on the absolute dates and selected information)\\
- [For each selected sentence, specifically elaborate the unique information from this sentence that is required to answer the question. Think about how to frame the question without explicitly stating this unique information.]\\
\newline
[Answer derivation]:\\
Compute the answer using the absolute dates from above.\\
\newline
<scratchpad>\\
\newline
Output your results in the following format:\\
<scratchpad>[Outline your reasoning including the derived absolute points in time here.]</scratchpad>\\
<results>\\
<qa>\\
<question>[Your generated question about the time span]</question>\\
<answer>[The precise and concise answer to the question]</answer>\\
<additional\_sentence\_ids>[List of additional sentence IDs required to answer the question, if applicable]</additional\_sentence\_ids>\\
<additional\_sentence\_explanation>[Explain the unique information of each additional sentence that is required to answer the question and justify why it is needed, if applicable]</additional\_sentence\_explanation>\\
</qa>\\
</results>\\
\newline
Remember to make your question as specific as possible, ensuring that it can only be answered using the information from the selected sentences (and any additional sentences you may need to include). If you need to make assumptions, state them clearly in the question. Always provide a precise and concise answer to the question you generate.\\
\newline
Important reminders:\\
- While each event provides a unique date, which can be helpful to identify the question, the provided date does not necessarily apply to all information listed in the event outline (or the selected sentences extracted from the event outline). The event date is always the latest possible date based on everything that has happened within the event. Think carefully about how the event date applies to the selected sentences before you start drafting your question. Some events may have happened in the past even though they are not described in the past tense. Use your common sense and the sequence of sentences within the event. If you are reasonably uncertain, make your assumption explicit and include it in your question.\\
- Ensure that the question can ONLY be answered when ALL of the selected sentences are present, and that ALL of the selected sentences are needed to answer the question.\\
- Make sure that no other information from the <complete\_outline> is sufficient to infer the correct answer.\\
- If you need to include additional sentences, clearly explain why they are necessary and what unique information they provide.\\
- In your question clearly indicate if some point in time happens in the future as such (for example when including the planned end date of an ongoing or planned event).\\
- In your question, DO NOT include the required temporal information. The temporal information must be inferred from the content of the selected sentences.\\
- DO NOT create questions about time-spans that mix units that cannot be compared in a meaningful way (such as minutes and days, days and years).\\
- If you make assumptions, make sure they are plausible.\\
- If you rely on the event date, carefully assess if the whether the information likely occurred before or after the event date.\\
- State your assumptions that are required to answer the question clearly in the question.\\
- Make sure that the question cannot be answered without considering unique content from ALL selected evidence sentences (from <selected\_sentences> and <additional\_sentence\_ids>)\\
- Explicitly compute and list all dates, including start and end dates, based on your assumptions before you draft your question.\\
- The answer should only consist of the correct time span and should not be a complete sentence.\\
- Ensure that you DO NOT explicitly refer to absolute dates or relative time spans (from the selected sentences) in the question. It is important that answering the question requires extracting this information from these sentences.
\end{tcolorbox}

\subsection{Question Refining Prompt for Time-span Questions}
\label{appendix:prompts:question-gen:timespan:refine}
\noindent
\footnotesize
\begin{tcolorbox}[
    colback=gray!15,
    colframe=gray!15,
    coltext=darkgray,
    boxrule=0pt,
    width=\textwidth,
    boxsep=5pt,
    sharp corners,
    breakable
]Check the response from above following these steps:\\
1. Carefully examine the response and compare it to the task instructions.\\
2. Check if the response meets all criteria outlined in the task instructions.\\
3. If you find any problems or discrepancies, describe them in detail using the <scratchpad> tags.\\
4. If corrections are needed, make them and explain your changes within the <scratchpad> tags.\\
5. Provide the complete and corrected response, even if you didn't change anything.\\
\newline
Use the following output format for your response:\\
\newline
<scratchpad>\\
(Outline your reasoning here, including any problems found and explanations for corrections made)\\
</scratchpad>\\
\newline
<results>\\
<qa>\\
<question>[Your refined question about the time span]</question>\\
<answer>[The precise and concise answer to the question]</answer>\\
<additional\_sentence\_ids>[List of additional sentence IDs required to answer the question (including those selected in the previous response), if applicable]</additional\_sentence\_ids>\\
<additional\_sentence\_explanation>[Explain the unique information of each additional sentence that is required to answer the question and justify why it is needed, if applicable]</additional\_sentence\_explanation>\\
</qa>\\
</results>\\
\newline
Remember to provide a complete and accurate response, addressing all aspects of the task instructions. If no changes are needed, simply reproduce the original response in the correct format.\\
\newline
Important:\\
- Make minimal changes.\\
- Do not remove assumptions just because they are not explicitly stated.
\end{tcolorbox}

\subsection{Distractor Prompt for Time-span Questions}
\label{appendix:prompts:question-gen:timespan:distractor}
\noindent
\footnotesize
\begin{tcolorbox}[
    colback=gray!15,
    colframe=gray!15,
    coltext=darkgray,
    boxrule=0pt,
    width=\textwidth,
    boxsep=5pt,
    sharp corners,
    breakable
]You are an AI assistant tasked with creating challenging multiple-choice distractor options for a question based on a fictional event outline. Your goal is to create plausible but incorrect answer choices that will test the reader's understanding of the given information.\\
\newline
First, carefully read and analyze the following fictional event outline:\\
\newline
<outline>\\
\{\{STORYLINE\_OUTLINE\_TO\_DATE\}\}\\
</outline>\\
\newline
Now, consider the following question:\\
\newline
<question>\\
\{\{CURRENT\_TIMESPAN\_QUESTION\}\}\\
</question>\\
\newline
The correct answer to this question is:\\
\newline
<correct\_answer>\\
\{\{CORRECT\_TIMESPAN\_ANSWER\}\}\\
</correct\_answer>\\
\newline
To correctly answer this question, these sentences from the outline are required:\\
\newline
<selected-sentences>\\
\{\{SELECTED\_SENTENCES\}\}\\
</selected-sentences>\\
\newline
Your task is to create \{\{NUM\_DISTRACTORS\}\} plausible but incorrect multiple-choice distractor options for this question. These distractors should be challenging and appear realistic, but must not be valid answers to the question.\\
\newline
Follow these guidelines when creating effective distractors:\\
1. Ensure each distractor is clearly incorrect when compared to the correct answer.\\
2. Use information from the fictional event outline to make distractors sound plausible.\\
3. If possible, incorporate specific values or details from the outline to increase believability.\\
4. Align distractors with non-answer text from the outline to make them more challenging.\\
5. Vary the type and structure of distractors to avoid patterns.\\
6. Ensure distractors are distinct from each other and the correct answer.\\
7. Make sure that the distractor is not by accident a valid answer based on different information from the outline.\\
8. Make sure all distractor options are plausible.\\
\newline
Present your \{\{NUM\_DISTRACTORS\}\} distractor options in the following format:\\
\newline
<distractors>\\
<distractor>\\
<answer>[The incorrect answer]</answer>\\
<explanation>[A brief explanation why it is incorrect]</explanation>\\
<distractor-sentences>[Comma separated sentence IDs of the sentences that make the distractor sound plausible]</distractor-sentences>\\
</distractor>\\
\newline
[Repeat the above structure for each distractor]\\
</distractors>\\
\newline
After each distractor, provide a brief explanation of why it's incorrect but plausible, and list all sentence IDs (as a comma-separated list) from the outline that make the distractor sound plausible. Leave the list of sentence IDs empty if none other sentence from the outline increases the plausibility for this distractor. Both should be included within the respective distractor tags.\\
\newline
Remember, your goal is to create challenging distractors that will test the reader's understanding of the fictional event outline while ensuring they are definitively incorrect. Use your knowledge and creativity to craft distractors that are both believable and clearly distinguishable from the correct answer.\\
\newline
Before presenting your final distractors, use a <scratchpad> to brainstorm and refine your ideas. Consider different aspects of the outline that could be misinterpreted or confused, and how you can use these to create compelling distractors.\\
\newline
Once you have finalized your distractors, present them in the format specified above, ensuring that each one is unique, challenging, and clearly incorrect when compared to the correct answer.
\end{tcolorbox}

\subsection{Sentence Selection Prompt for Multi-hop Questions}
\label{appendix:prompts:question-gen:multi-hop:select}
\noindent
\footnotesize
\begin{tcolorbox}[
    colback=gray!15,
    colframe=gray!15,
    coltext=darkgray,
    boxrule=0pt,
    width=\textwidth,
    boxsep=5pt,
    sharp corners,
    breakable
]You are an AI assistant tasked with analyzing fictional event descriptions and identifying sentences that can be used to generate multi-hop questions with a bridge entity. Your goal is to find up to \{\{MAX\_NUMBER\_SELECTIONS\}\} tuples, each comprising 2 sentences, that can be used to create challenging and interesting multi-hop questions with a bridge entity.\\
\newline
Here are the key inputs for this task:\\
\newline
1. Fictional events to analyze:\\
<events>\\
\{\{OUTLINES\}\}\\
</events>\\
\newline
2. The bridge entity name for this task:\\
<bridge\_entity>\\
\{\{BRIDGE\_ENTITY\_NAME\}\}\\
</bridge\_entity>\\
\newline
3. Known named entities:\\
<known-named-entities>\\
\{\{KNOWN\_PREV\_NAMED\_ENTITIES\}\}\\
</known-named-entities>\\
\newline
4. List of sentence IDs that include the bridge entity:\\
<possible\_sentence\_ids>\\
\{\{POSSIBLE\_SENTENCE\_IDS\}\}\\
</possible\_sentence\_ids>\\
\newline
When selecting sentences, follow these guidelines:\\
1. If two events are provided, ensure that each tuple covers both events (i.e., one sentence ID stems from the first event, and the other sentence ID stems from the second event).\\
2. If only one event is provided, the selected sentence IDs can stem from the same event.\\
3. Look for diverse combinations of sentences across the selected events.\\
4. Only consider sentence IDs from the provided list of possible sentence IDs.\\
5. Ensure that ALL selected sentence IDs include the bridge named entity.\\
6. Make sure that the specific information needed to identify the bridge entity cannot be known from any other sentence, nor from the knowledge base entries from the known named entities.\\
7. Verify that the information from the known named entities is not sufficient to replace the detailed information found in the selected sentences.\\
\newline
Before providing your final answer, use a <scratchpad> section to analyze the events and think through potential sentence combinations. This will help you identify the most suitable pairs for multi-hop questions. In your scratchpad, consider the following:\\
1. Identify sentences that contain specific information about the bridge entity.\\
2. Look for connections between sentences that could form the basis of a multi-hop question.\\
3. Evaluate whether the selected sentences provide unique information not available in other sentences or known named entities.\\
4. Consider how the sentences could be used to create a challenging and interesting question.\\
\newline
For each suitable combination of sentences you find, output your selection using this format:\\
\newline
<selection>\\
<ids>[IDs of the selected sentences (comma separated)]</ids>\\
<explanation>[Explanation of why these sentences are suitable for generating a multi-hop question with the given bridge entity]</explanation>\\
</selection>\\
\newline
Try to find as many interesting combinations of sentences as possible, but do not exceed \{\{MAX\_NUMBER\_SELECTIONS\}\} selections.\\
\newline
If you cannot find any meaningful combination of sentences based on which good multi-hop questions with this bridge entity can be generated, return an empty list.\\
\newline
Wrap all your results, including the scratchpad and selections, in a <result> root node. Your final output should look like this:\\
\newline
<result>\\
<scratchpad>\\
(Your analysis and thought process)\\
</scratchpad>\\
\newline
(Your selections, if any)\\
\newline
</result>\\
\newline
Remember, the goal is to identify sentence pairs that can be used to create challenging and interesting multi-hop questions with the given bridge entity. Focus on finding unique and specific information that requires reasoning over both sentences to answer a potential question.
\end{tcolorbox}

\subsection{Question Writing Prompt for Multi-hop Questions}
\label{appendix:prompts:question-gen:multi-hop:writing}
\noindent
\footnotesize
\begin{tcolorbox}[
    colback=gray!15,
    colframe=gray!15,
    coltext=darkgray,
    boxrule=0pt,
    width=\textwidth,
    boxsep=5pt,
    sharp corners,
    breakable
]You are an AI assistant tasked with analyzing two sentences about fictional events involving fictional named entities and generating multi-hop questions using a bridge entity. Your goal is to identify specific details within these sentences and create \{\{NUM\_QUESTIONS\}\} question-answer pairs that require reasoning over both sentences.\\
\newline
First, familiarize yourself with the complete storyline outline of the fictional events:\\
\newline
<storyline\_outline>\\
\{\{STORYLINE\_OUTLINE\_TO\_DATE\}\}\\
</storyline\_outline>\\
\newline
Now, focus on the two selected sentences based on which you must generate multi-hop questions:\\
\newline
<selected\_sentences>\\
\{\{SELECTED\_SENTENCES\}\}\\
</selected\_sentences>\\
\newline
The bridge entity in these sentences is:\\
\newline
<bridge\_entity>\{\{BRIDGE\_ENTITY\_NAME\}\}</bridge\_entity>\\
\newline
To generate multi-hop questions that require reasoning over both sentences, follow these steps:\\
\newline
1. Identify information about the bridge entity in both sentences that is unique to these two sentences and cannot be found based on any other information from the provided storyline outline. This information should be as specific as possible, to avoid any ambiguities or overlap with other information from other sentences.\\
\newline
2. To generate a question-answer pair, ask for specific information about the bridge entity from one sentence while describing the bridge entity with information from the other sentence. Make sure that the correct answer is concise and factual. The answer should focus on very specific details that can be described in few words. \\
   - Make sure that the question is answerable.\\
   - Make sure to ask for short and concise information\\
   - Make sure that the way you paraphrase the bridge named entity clearly identifies the bridge entity using the unique information from the selected sentence (and not more)\\
   - Ensure that you do not introduce additional ambiguity when paraphrasing: \\
        a. If the evidence says that the bridge entity announces the creation of something, it does not mean that the bridge entity created it. Be careful in your word choice to avoid ambiguities.\\
        b. Do include specific information (such as the named entity's profession or role) from the <storyline\_outline> if they are not explicitly clear in from the selected sentences.\\
\newline
3. Make sure that the information from the selected sentences is sufficient to answer the question with certainty. Your question must not rely on other information that is only communicated in different sentences. If it is important to include additional information that is not included in the selected sentences, add the additional sentence IDs that are needed to the output. \\
   - For each of the new sentences, explain the unique information from the sentence that is required to answer the question \\
   - For each of the new sentences, ensure that the unique information from the new sentences is required to answer the question with certainty. Refine the question idea if necessary.\\
\newline
4. Do not mention the bridge named entity explicitly in your questions. Paraphrase the bridge named entity using the unique information from one of the selected sentences.\\
\newline
5. Ensure that the question can ONLY be answered when having access to the information from BOTH sentences. Make sure that all selected sentences must be considered to answer the question. Avoid using the bridge entity itself as the answer.\\
\newline
6. Ensure that the information that is required from each sentence is unique within this sentence: It can neither be inferred nor extracted from any other sentence within the provided storyline outline. If the question can be answered based on the other information from the storyline outline, increase the specificity of the required details and ensure they are unique to the selected sentences.\\
\newline
7. Verify that your question does not assume any relations that are not clear from the selected sentences. \\
   - You can only assume that the bridge entity is identical across all sentences. Other information may not refer to the identical entity. For example, a group of people in one sentence may not be identical to a group of people in another sentence.\\
   - Do not assume causality between the selected sentences.\\
   If in doubt, rely on the bridge entity.\\
\newline
8. Compare each specific detail in the question with the selected sentences. Make sure that each detail can with certainty be inferred from the selected sentences. If not, omit or generalize the specific details that cannot be inferred from the selected sentences.\\
   - Only focus on the selected sentences!\\
   - Correct information that is only known from the <storyline\_outline> but not from the <selected\_sentences> must NOT be used in the question.\\
  - DO NOT assume the profession of an entity when paraphrasing it, unless it is fully clear from the selected sentences! DO NOT write "artist" if the <selected\_sentences> do not introduce this person as "artist".\\
\newline
Use the following scratchpad format to outline your reasoning:\\
\newline
<scratchpad>\\
Sentence information: \\
(List all specific information from each sentence, that can be used to create a question)\\
\newline
Idea: \\
(Brainstorm a good question idea that can be answered with the selected sentences using the unique information you have identified. Ask for specific information from one of the sentences. Make sure that the correct answer is short and concise.)\\
\newline
Additional sentences:\\
(Check if the selected sentences are sufficient to write a self-contained question based on the defined criteria. If you believe you need additional sentence(s), for each additional sentence justify your decision by stating the sentence id, the unique information from this sentence and why it is essential. If no additional sentence is needed, explain why the current selection is self-contained.)\\
\newline
Verification:\\
(Draft your question and verify points 4,5,6,7,8)\\
\newline
</scratchpad>\\
\newline
It is crucial that you adhere to the following criteria:\\
- Answering the question is only possible based on the unique information that can be found in the selected sentences.\\
- Answering the question requires combining the unique information from ALL selected sentences.\\
- The question is specific enough to allow only for one valid answer. There are no other interpretations based on the storyline outline which would allow for a different valid answer.\\
- The answer must be complete. \\
\newline
Important reminders:\\
- The sentences describe part of a fictional event. Your questions must address the fictional event.\\
- Generate \{\{NUM\_QUESTIONS\}\} multi-hop questions.\\
- Verify the generated questions based on the defined criteria and correct them if necessary.\\
- Make sure to be specific in the question when paraphrasing the bridge entity to avoid ambiguities. It must be clear to identify the bridge entity based on the details provided in the question.\\
- DO NOT ask for "specific" information verbatim. Instead, provide specific details in the question that can be answered with concrete values.\\
- While you must ask for very specific information, make sure the answer itself is a short and concise phrase!\\
\newline
Format your output as follows:\\
\newline
<scratchpad>\\
(Your thinking can go here.)\\
</scratchpad>\\
<results>\\
<qa>\\
<question>[Your generated question]</question>\\
<answer>[The answer based on the evidence sentences]</answer>\\
<additional\_sentence\_ids>[List of additional sentence IDs required to answer the question, if applicable]</additional\_sentence\_ids>\\
<additional\_sentence\_explanation>[Explain the unique information of each additional sentence that is required to answer the question and justify why it is needed, if applicable]</additional\_sentence\_explanation>\\
</qa>\\
</results>\\
\newline
Now, begin the task of generating question-answer pairs based on the provided sentences. Include all question-answer pairs within a single <results> tag.
\end{tcolorbox}

\subsection{Distractor Prompt for Multi-hop Questions}
\label{appendix:prompts:question-gen:multi-hop:distractor}
\noindent
\footnotesize
\begin{tcolorbox}[
    colback=gray!15,
    colframe=gray!15,
    coltext=darkgray,
    boxrule=0pt,
    width=\textwidth,
    boxsep=5pt,
    sharp corners,
    breakable
]You are an AI assistant tasked with creating challenging multiple-choice distractor options for a question based on a fictional event outline. Your goal is to create plausible but incorrect answer choices that will test the reader's understanding of the given information.\\
\newline
First, carefully read and analyze the following fictional event outline:\\
\newline
<outline>\\
\{\{STORYLINE\_OUTLINE\_TO\_DATE\}\}\\
</outline>\\
\newline
Now, consider this question and its correct answer:\\
\newline
<question>\\
\{\{QUESTION\}\}\\
</question>\\
\newline
<correct\_answer>\\
\{\{ANSWER\}\}\\
</correct\_answer>\\
\newline
The correct answer is based on these sentences:\\
<selected-sentences>\\
\{\{SELECTED\_SENTENCES\}\}\\
</selected-sentences>\\
\newline
Your task is to create \{\{NUM\_DISTRACTORS\}\} plausible but incorrect multiple-choice distractor options for this question. These distractors should be challenging and appear realistic, but must not be valid answers to the question.\\
\newline
Follow these guidelines when creating effective distractors:\\
1. Ensure each distractor is clearly incorrect when compared to the correct answer.\\
2. Use information from the fictional event outline to make distractors sound plausible.\\
3. If possible, incorporate specific values or details from the outline to increase believability.\\
4. Align distractors with non-answer text from the outline to make them more challenging.\\
5. Vary the type and structure of distractors to avoid patterns.\\
6. Ensure distractors are distinct from each other and the correct answer.\\
\newline
Present your distractor options in the following format:\\
\newline
<distractors>\\
<distractor>\\
<answer>[The incorrect answer]</answer>\\
<explanation>[A brief explanation why it is incorrect]</explanation>\\
<distractor-sentences>[Comma separated sentence IDs of the sentences that make the distractor sound plausible]</distractor-sentences>\\
</distractor>\\
\newline
(Repeat the above structure for each distractor)\\
</distractors>\\
\newline
After each distractor, provide a brief explanation of why it's incorrect but plausible, and list all sentence IDs (as a comma-separated list) from the outline that make the distractor sound plausible. Leave the list of sentence IDs empty if none other sentence from the outline increases the plausibility for this distractor. Both should be included within the respective distractor tags.\\
\newline
Remember, your goal is to create challenging distractors that will test the reader's understanding of the fictional event outline while ensuring they are definitively incorrect. Use your knowledge and creativity to craft distractors that are both believable and clearly distinguishable from the correct answer.
\end{tcolorbox}

\subsection{Question Writing Prompt for False Premise Questions}
\label{appendix:prompts:question-gen:false-premise:write}
\noindent
\footnotesize
\begin{tcolorbox}[
    colback=gray!15,
    colframe=gray!15,
    coltext=darkgray,
    boxrule=0pt,
    width=\textwidth,
    boxsep=5pt,
    sharp corners,
    breakable
]You are an AI assistant tasked with generating false-premise questions based on fictional events. Your goal is to create questions that cannot be answered because they make incorrect assumptions about the events. Follow these instructions carefully:\\
\newline
You will be provided with the following information:\\
<question>\\
\{\{QUESTION\}\}\\
</question>\\
\newline
<selected\_sentences>\\
\{\{SELECTED\_SENTENCES\}\}\\
</selected\_sentences>\\
\newline
<answer>\\
\{\{ANSWER\}\}\\
</answer>\\
\newline
<context>\\
\{\{STORYLINE\_OUTLINE\_TO\_DATE\}\}\\
</context>\\
\newline
To generate false-premise questions:\\
1. Identify key information in one of the two selected sentences.\\
2. Create a question that contradicts this key information while keeping other details intact.\\
3. Ensure the false premise is mutually exclusive with the original information.\\
4. Make the questions challenging, with false premises that are easy to miss but mutually exclusive to the evidence sentences and context. For example:\\
   - If you change a name, change the lastname only\\
   - If you refer to a person or place, rather than changing the name, refer to a changed property of this entity (e.g., "in a 60-year-old building" instead of "in the 20-year-old office")\\
   - Replace with similar mutually exclusive cohyponyms (e.g., replace a cocker spaniel with a poodle)\\
   - In all of these cases, ensure that you do not accidentally create a valid question!\\
5. Keep the question as similar as possible to the original question, asking for the same information but changing small details that contradict the two sentences.\\
6. Only include ONE false premise for in each question.\\
\newline
Consider the additional context when creating false premises:\\
1. Avoid creating questions that can be validly answered using information from the context.\\
2. Ensure that the false premise remains inconsistent with both the selected sentences and the context.\\
\newline
Generate multiple false-premise questions if possible, each based on different key information from the selected sentences.\\
The changed information must be contradictory to one of the selected sentences only. \\
Output the sentence that is contradictory (or mutually exclusive to) the generated false-premise question.\\
\newline
Format your output as follows:\\
<scratchpad>\\
(Your thinking process)\\
</scratchpad>\\
<results>\\
<qa>\\
<question>[Your generated false-premise question]</question>\\
<false-premise>[Explanation of the false premise introduced]</false-premise>\\
<contradictory-sentence-id>[The sentence ID that is contradictory to the changed information]</contradictory-sentence-id>\\
</qa>\\
</results>\\
\newline
Repeat the <qa> section for each false-premise question you generate.\\
\newline
Remember:\\
- Keep your responses concise and avoid verbose explanations.\\
- Ensure that the false premises are subtle but clear enough to invalidate the question.\\
- Generate multiple false-premise questions if possible, based on different key information from the selected sentences.\\
- Always consider the additional context to avoid creating questions that can be validly answered using that information.
\end{tcolorbox}

\subsection{Question Writing Prompt for Uncertain Specificity Questions}
\label{appendix:prompts:question-gen:uncertain-specificity:write}
\noindent
\footnotesize
\begin{tcolorbox}[
    colback=gray!15,
    colframe=gray!15,
    coltext=darkgray,
    boxrule=0pt,
    width=\textwidth,
    boxsep=5pt,
    sharp corners,
    breakable
]You are an AI assistant tasked with generating false-premise questions based on fictional events. Your goal is to create questions that cannot be answered because they make possible but unknown assumptions about the events. Follow these instructions carefully:\\
\newline
You will be provided with the following information:\\
<question>\\
\{\{QUESTION\}\}\\
</question>\\
\newline
<selected\_sentences>\\
\{\{SELECTED\_SENTENCES\}\}\\
</selected\_sentences>\\
\newline
<answer>\\
\{\{ANSWER\}\}\\
</answer>\\
\newline
<context>\\
\{\{STORYLINE\_OUTLINE\_TO\_DATE\}\}\\
</context>\\
\newline
To generate unknown-premise questions:\\
1. Identify key information in one of the two selected sentences.\\
2. Modify the question by making it more specific. \\
   - The added details must be POSSIBLE based on the provided sentences and the provided list of all outlines. \\
   - The added details must be UNVERIFIED based on the provided sentences and the provided list of all outlines. \\
   - The added details must be substantial to require additional verification. Avoid details that are not only possible but also very likely.\\
3. Keep the question as similar as possible to the original question, asking for the same information but changing small details that contradict the two sentences.\\
4. Only include ONE specific unverified detail to the question.\\
\newline
Consider the additional context when creating the new question:\\
1. Make sure that the details you add cannot be confirmed by any of the sentences\\
2. Make sure that the details you add cannot be refuted by any of the sentences\\
\newline
Generate multiple such questions if possible, each based on different key information from the selected sentences.\\
Only output the questions for which you are certain that:\\
- The details you add cannot be confirmed by any of the sentences\\
- The details you add cannot be refuted by any of the sentences\\
- The details are substantial enough to require additional verification.\\
\newline
Here are some examples of ways to add specificity:\\
- Let a person have a more specific role: Instead of "a criminal of the ring," say "a ring leader" or "a lookout."\\
- Add a specific characteristic to an object: Instead of "a car," say "a red sports car" or "an old sedan."\\
- Use a hyponym where the 'is-a' relation holds: Instead of "a person," say "a woman" or "a child."\\
- Specify a location: Instead of "a park," say "a national park" or "Central Park."\\
- Specify a time: Instead of "at night," say "at 9 PM" or "during the full moon."\\
- Specify a number or quantity: Instead of "several books," say "three books" or "a dozen books."\\
- Specify a direction: Instead of "headed away," say "headed east" or "to the mountains."\\
- Specify a duration: Instead of "waited," say "waited for 15 minutes" or "waited for an hour."\\
\newline
Format your output as follows:\\
<scratchpad>\\
(Your thinking process)\\
</scratchpad>\\
<results>\\
<qa>\\
<question>[Your generated question]</question>\\
<explanation>[Explanation of the unknown information]</explanation>\\
</qa>\\
</results>\\
\newline
Repeat the <qa> section for each question you generate.\\
\newline
Remember:\\
- Keep your responses concise and avoid verbose explanations.\\
- Always consider the additional context to avoid creating questions that can be validly answered using that information.
\end{tcolorbox}

\subsection{System Prompt for SensationalNews}
\label{appendix:prompts:profile:sensational}
\noindent
\footnotesize
\begin{tcolorbox}[
    colback=gray!15,
    colframe=gray!15,
    coltext=darkgray,
    boxrule=0pt,
    width=\textwidth,
    boxsep=5pt,
    sharp corners,
    breakable
]You are simulating the profile of a newspaper called SensationalNews. The newspaper has a sensational and attention-focused style, targeting an audience that thrives on captivating, dramatic stories. Your task is to create a detailed profile for SensationalNews, providing specific information for the following categories:\\
\newline
Core Values:\\
1. **Boldness** – SensationalNews embraces the power of bold, attention-grabbing headlines and stories. It is not afraid to push boundaries and be daring in its reporting.\\
2. **Emotional Engagement** – The newspaper believes in connecting with its readers emotionally, stirring curiosity, excitement, and often shock.\\
3. **Entertainment Focus** – While reporting on serious issues, SensationalNews aims to entertain as much as inform, blending elements of storytelling with journalism.\\
4. **Instant Impact** – The publication values creating immediate reactions from its readers, with a preference for stories that generate quick buzz and wide readership.\\
5. **Provocation** – SensationalNews sees value in challenging the status quo and providing thought-provoking commentary that sparks conversation.\\
\newline
Reporting Style:\\
1. **Dramatic Headlines** – Stories are often framed with sensational headlines that pull readers in with their emotional intensity and exaggerated flair.\\
2. **Exaggerated Descriptions** – Reportage often includes vivid and overstated language to enhance the emotional impact of the story.\\
3. **Heavy Use of Visuals** – Eye-catching images, graphics, and infographics are a staple of the paper, with emphasis on impactful visuals that complement the story’s drama.\\
4. **Personalized Narratives** – Stories are often told from a personal angle, involving individual experiences that amplify the human element.\\
5. **Speculative** – Rather than just focusing on the facts, stories often venture into speculation, offering bold predictions or unverified claims to keep the audience intrigued.\\
6. **Use of Hyperbole** – The reporting often utilizes hyperbolic language to make events appear larger-than-life and more significant.\\
\newline
Perspective on Common Issues:\\
1. **Economics** – SensationalNews is generally critical of traditional economic systems, often focusing on scandals involving the rich and powerful. It prefers to highlight stories about financial collapses, corporate corruption, or the shocking impact of economic disparity.\\
2. **Foreign Policy** – SensationalNews is dramatic in its portrayal of international events, often emphasizing conflicts, scandals, or conspiracies involving world leaders and governments. It tends to adopt a skeptical, sometimes alarmist stance on foreign relations.\\
3. **Social Topics** – The newspaper highlights polarizing social issues, often focusing on divisive cultural debates. It tends to amplify sensational aspects of social movements, such as protests, controversies, or public figures involved in scandals.\\
4. **Environment** – SensationalNews might portray environmental issues in a dramatic light, focusing on disasters, environmental collapses, or highly controversial claims about climate change, often exaggerating the urgency or apocalyptic aspects of the problem.\\
5. **Technology** – The paper covers the darker side of technology, emphasizing security breaches, data privacy violations, and the dangers of technological advancement rather than celebrating progress.\\
\newline
Preferred Topics:\\
1. **Celebrity Scandals** – SensationalNews thrives on high-profile stories involving celebrities, with a focus on personal drama, breakups, and tabloid-like revelations.\\
2. **Crime and Scandals** – Reports on criminal activities, particularly those involving famous individuals or shocking details, dominate the coverage.\\
3. **Political Confrontations** – The paper frequently covers political scandals, corruption, and rivalries, focusing on the drama and intrigue surrounding political figures.\\
4. **Natural Disasters** – The publication has a keen interest in reporting on natural disasters, often dramatizing the scale and devastation of events to maintain reader engagement.\\
5. **Conspiracy Theories** – SensationalNews is known for reporting on conspiracy theories, often promoting speculative narratives that stir curiosity and fuel widespread discussions.\\
\newline
Things They Like:\\
1. **Conflict and Controversy** – SensationalNews enjoys highlighting dramatic confrontations, whether in politics, entertainment, or social issues, preferring stories with high emotional stakes.\\
2. **Shocking Revelations** – The newspaper thrives on uncovering secrets, hidden truths, or surprising twists that keep readers on the edge of their seats.\\
3. **Daring Individuals** – People who challenge norms or disrupt established systems are portrayed positively, especially if they are seen as bold or rebellious.\\
4. **Unpredictable Events** – The paper enjoys reporting on events that are unpredictable and out of the ordinary, especially if they provide an opportunity for dramatic storytelling.\\
5. **Misinformation and Sensational Claims** – SensationalNews tends to embrace bold, unverified claims or takes stories with sensational twists, appealing to readers who enjoy speculation.\\
\newline
Things They Dislike:\\
1. **Bureaucracy and Red Tape** – SensationalNews dislikes bureaucratic systems and slow, cautious approaches to news reporting, preferring quick and dramatic action over formalities.\\
2. **Censorship and Control** – The newspaper is critical of any form of censorship and dislikes anything that restricts freedom of speech or access to sensational content.\\
3. **Mediocre News** – Stories that are deemed 'boring' or lacking in dramatic flair are typically downplayed or not covered at all.\\
4. **Overly Technical Reporting** – SensationalNews tends to avoid overly complex, fact-heavy reports that lack emotional appeal or dramatic tension, preferring stories that are easily digestible and engaging.\\
5. **Conservative, Mainstream Views** – The publication frequently criticizes mainstream perspectives, especially if they are perceived as dull, traditional, or not engaging enough for its audience.\\
\newline
This profile should help to fully simulate the voice, character, and editorial stance of SensationalNews.
\end{tcolorbox}

\subsection{System Prompt for ObjectiveNews}
\label{appendix:prompts:profile:objective}
\noindent
\footnotesize
\begin{tcolorbox}[
    colback=gray!15,
    colframe=gray!15,
    coltext=darkgray,
    boxrule=0pt,
    width=\textwidth,
    boxsep=5pt,
    sharp corners,
    breakable
]You are simulating the profile of the newspaper "ObjectiveNews." The newspaper's key attributes are being objective, fair, and unbiased. Based on these attributes, generate detailed information for the following categories:\\
\newline
1. **Core Values:**\\
   - Objectivity: Commitment to presenting information in a neutral, impartial manner, free from personal bias or opinion.\\
   - Integrity: Upholding the truth and reporting facts as accurately as possible, without distortion or exaggeration.\\
   - Accountability: Holding public figures, institutions, and entities responsible for their actions while maintaining fairness in coverage.\\
   - Transparency: Providing clear sources and evidence for all reported facts, allowing readers to assess the validity of the information.\\
   - Diversity of Viewpoints: Ensuring a range of perspectives are included in coverage, especially on contentious or polarizing issues.\\
\newline
2. **Reporting Style:**\\
   - Fact-based Analysis: Reporting is rooted in verified data, evidence, and credible sources, with minimal use of speculation or conjecture.\\
   - Clear and Concise: Information is presented in a straightforward manner, avoiding sensationalism and overly complex language.\\
   - Balance: Both sides of an issue are presented fairly, without favoring one over the other, unless there is clear evidence to support one perspective.\\
   - Contextualization: Stories are often accompanied by relevant background information to help readers understand the broader significance.\\
   - Non-emotive Storytelling: The tone remains neutral and objective, avoiding sensationalist or emotionally charged language.\\
\newline
3. **Perspective on Common Issues:**\\
   - **Economics:** Advocates for policies that promote sustainable economic growth, with a focus on fairness, equity, and long-term stability. Cautions against overly partisan economic rhetoric.\\
   - **Foreign Policy:** Supports diplomacy and peaceful conflict resolution, with a preference for multilateral cooperation over unilateral action. Strong focus on human rights and international law.\\
   - **Social Topics:** Favors policies that promote social justice and equality, emphasizing data-driven solutions to complex social issues like healthcare, education, and poverty.\\
   - **Environmental Issues:** Prioritizes scientifically-backed solutions to environmental challenges, including climate change, advocating for green energy and conservation policies that do not sacrifice economic stability.\\
   - **Technology and Innovation:** Focuses on how technological advancements impact society, emphasizing both the benefits and ethical concerns of emerging technologies like AI, privacy issues, and digital rights.\\
\newline
4. **Preferred Topics:**\\
   - **Political Integrity:** Stories examining government transparency, accountability, and the ethical behavior of political figures.\\
   - **Public Health:** Reports on healthcare policies, advancements in medical research, and public health crises.\\
   - **Education and Equality:** Coverage of educational reform, access to quality education, and efforts to reduce inequality in education.\\
   - **Global Affairs:** International relations, especially focusing on human rights, diplomacy, and global cooperation.\\
   - **Environmental Sustainability:** Detailed reporting on efforts to combat climate change, protect natural resources, and promote sustainability.\\
\newline
5. **Things They Like:**\\
   - **Diverse Opinions:** A variety of perspectives in opinion pieces, as long as they are well-supported by facts and logic.\\
   - **Data-Driven Reporting:** Stories that use credible statistics and research to inform the narrative.\\
   - **Positive Social Change:** Efforts that aim to make society more just, equitable, and sustainable, especially when supported by factual evidence.\\
   - **Political Accountability:** Actions or initiatives that hold governments and corporations accountable to the public.\\
   - **International Cooperation:** Efforts towards resolving global issues through diplomacy, multilateralism, and collaboration among nations.\\
\newline
6. **Things They Dislike:**\\
   - **Sensationalism:** Coverage that distorts or exaggerates facts to provoke emotional reactions, rather than providing balanced, fact-based information.\\
   - **Partisan Bias:** Reporting that favors one political party or ideology over another, especially when it compromises the integrity of the story.\\
   - **Misinformation:** Spreading false or misleading information, especially when it is not corrected promptly.\\
   - **Lack of Accountability:** Situations where individuals or institutions are not held responsible for their actions, especially in cases of public corruption or negligence.\\
   - **Polarization:** The deepening of divisions within society that lead to less constructive debate and more conflict, especially when fueled by media sources.\\
\newline
This profile should help guide the simulation of "ObjectiveNews" as a fair, unbiased, and factual newspaper, focused on reporting the truth in an informative and balanced manner.
\end{tcolorbox}

\subsection{System Prompt for ProgressiveNews}
\label{appendix:prompts:profile:progressive}
\noindent
\footnotesize
\begin{tcolorbox}[
    colback=gray!15,
    colframe=gray!15,
    coltext=darkgray,
    boxrule=0pt,
    width=\textwidth,
    boxsep=5pt,
    sharp corners,
    breakable
]You are simulating the profile of a newspaper called "ProgressiveNews," which is known for its progressive values and focus on promoting social change, equality, and justice. Your task is to generate detailed and specific information across the following categories:\\
\newline
1. **Core Values:**\\
   - Equality and Social Justice: Advocating for equal rights and opportunities for all individuals, regardless of race, gender, sexual orientation, or socioeconomic status.\\
   - Environmental Sustainability: Promoting policies and actions aimed at protecting the environment and combating climate change.\\
   - Inclusivity and Diversity: Emphasizing the importance of diverse perspectives and inclusive practices in all aspects of society, particularly in leadership and decision-making.\\
   - Economic Equity: Focusing on reducing income inequality and ensuring that wealth is distributed fairly across society.\\
   - Accountability and Transparency: Holding corporations, governments, and other powerful entities accountable for their actions, particularly with regard to human rights and environmental impact.\\
\newline
2. **Reporting Style:**\\
   - Fact-Based Analysis: Ensuring that all reporting is supported by solid evidence and provides a thorough understanding of the issues at hand.\\
   - Investigative Journalism: Prioritizing deep dives into complex issues, uncovering hidden truths, and revealing systemic injustices.\\
   - Emotive Storytelling: Using personal narratives and human interest stories to create emotional connections with readers, driving home the importance of progressive causes.\\
   - Balanced Critique: Presenting multiple perspectives on a story, but with a critical eye toward power structures that perpetuate inequality.\\
   - Calls to Action: Frequently encouraging readers to get involved in social causes, engage with local activism, and support policies for systemic change.\\
\newline
3. **Perspective on Common Issues:**\\
   - Economics: Advocating for wealth redistribution, progressive taxation, universal healthcare, and a living wage. Critical of corporate greed and neoliberal economic policies.\\
   - Foreign Policy: Supporting human rights, international diplomacy, and foreign aid, particularly in conflict zones. Opposing military interventionism unless absolutely necessary for peacekeeping or humanitarian efforts.\\
   - Climate Change: Strongly pro-environmental action, advocating for renewable energy, reducing carbon emissions, and global cooperation to tackle climate crises.\\
   - Social Justice: Promoting policies to combat racism, sexism, LGBTQ+ discrimination, and other forms of oppression. Supporting the rights of marginalized communities.\\
   - Labor Rights: Championing workers' rights, unionization efforts, and fair labor practices, while opposing exploitative working conditions and low-wage labor.\\
\newline
4. **Preferred Topics:**\\
   - Social Inequality and Justice Reform: In-depth coverage on issues related to racial justice, criminal justice reform, gender equality, and LGBTQ+ rights.\\
   - Climate Change and Environmental Advocacy: Articles on environmental issues, sustainability practices, and climate action policies.\\
   - Healthcare and Education: Reporting on the importance of universal access to healthcare and high-quality education, and advocating for reform to make them more accessible to all.\\
   - Economic Policy and Workers' Rights: Focus on economic reforms, fair wages, universal basic income, and policies that support workers' rights and financial stability for all.\\
   - Technology and Society: Exploring the role of technology in shaping social change, both positively and negatively, and examining issues like data privacy and tech monopolies.\\
\newline
5. **Things They Like:**\\
   - Grassroots Movements: Supporting and covering local activism, protests, and grassroots initiatives that aim to create social change.\\
   - Progressive Legislation: Celebrating successful progressive policies, especially those that promote equality, environmental protection, and economic reform.\\
   - Diverse Representation: Highlighting the importance of diverse voices in politics, business, media, and culture, particularly those from marginalized communities.\\
   - Innovative Solutions: Coverage of new, creative solutions to social, economic, and environmental problems, including renewable energy, tech innovations, and community-driven initiatives.\\
   - Collaboration and Solidarity: Focusing on the power of collective action, whether through unions, international coalitions, or community organizations.\\
\newline
6. **Things They Dislike:**\\
   - Corporate Influence in Politics: Opposing the influence of large corporations and wealthy donors in politics, which they believe undermines democracy and equality.\\
   - Inequality and Economic Exploitation: Criticizing wealth inequality, the concentration of power in the hands of a few, and the exploitation of working-class people.\\
   - Authoritarianism: Opposing authoritarian regimes and policies that curtail individual freedoms, including restrictions on press freedom and political dissent.\\
   - Climate Denialism: Rejecting views and political movements that deny the existence of climate change or impede efforts to address it.\\
   - Discrimination and Hate Speech: Criticizing bigotry, hate speech, and discriminatory practices against minority groups based on race, religion, gender, sexual orientation, or disability.
\end{tcolorbox}

\subsection{System Prompt for ConservativeNews}
\label{appendix:prompts:profile:conservative}
\noindent
\footnotesize
\begin{tcolorbox}[
    colback=gray!15,
    colframe=gray!15,
    coltext=darkgray,
    boxrule=0pt,
    width=\textwidth,
    boxsep=5pt,
    sharp corners,
    breakable
]You are simulating the profile of a newspaper called "ConservativeNews," with a conservative editorial stance. Your task is to generate specific information about the newspaper's profile, broken down into the following categories:\\
\newline
Core Values:\\
1. **Traditional Family Values**: Emphasizes the importance of family structures, promoting policies that align with the preservation of traditional family roles.\\
2. **Limited Government**: Advocates for a smaller government with reduced taxes and fewer regulations, favoring individual freedom and local control.\\
3. **Patriotism and National Pride**: Supports strong national defense and respects the heritage, history, and symbols of the nation.\\
4. **Free Market Economy**: Promotes capitalist principles, emphasizing deregulation, entrepreneurship, and minimal government interference in business.\\
5. **Respect for Law and Order**: Stands for a strict interpretation of the law, emphasizing the importance of personal responsibility and strong criminal justice systems.\\
6. **Religious Freedom**: Upholds the belief that religious expression should be protected, often promoting Christianity as an integral part of cultural identity.\\
\newline
Reporting Style:\\
1. **Fact-based Analysis**: Prioritizes logical, evidence-driven reporting that appeals to rationality and often relies on data, research, and expert opinions.\\
2. **Concise and Direct**: Articles tend to be clear, direct, and to the point, often eschewing unnecessary detail for efficiency in communication.\\
3. **Opinionated Commentary**: Features strong editorial perspectives, often weaving political opinion into news coverage, particularly on contentious issues.\\
4. **Investigative Reporting**: Tends to focus on exposing government overreach, corruption, and liberal biases, with a focus on transparency and accountability.\\
5. **Emotive Storytelling**: Occasionally uses emotional appeal to underline stories, especially related to cultural or national pride, portraying a clear "us vs. them" narrative.\\
\newline
Perspective on Common Issues:\\
1. **Economics**: Advocates for tax cuts, deregulation, and economic policies that favor businesses, with a focus on reducing government spending and promoting job creation.\\
2. **Foreign Policy**: Strongly favors national sovereignty, supports a robust military, and tends to oppose international agreements or organizations that may undermine the country’s interests.\\
3. **Social Issues**: Often critical of progressive movements, especially when it comes to issues like LGBTQ+ rights, abortion, and social justice activism, preferring policies that protect traditional institutions.\\
4. **Immigration**: Advocates for strict border control and immigration laws, emphasizing the need for national security and the protection of American workers.\\
5. **Environmental Policy**: Generally skeptical of climate change policies that impose significant regulations on industries, favoring market-based solutions to environmental concerns.\\
\newline
Preferred Topics:\\
1. **Political Conservatism**: Covers topics like conservative victories, prominent conservative figures, and conservative solutions to political issues.\\
2. **National Security and Military**: Focuses on defense policy, military readiness, and law enforcement, with an emphasis on strengthening national security.\\
3. **Economic Policy and Market Trends**: Regularly covers free market economics, tax policy, and analysis of financial markets, businesses, and job creation.\\
4. **Cultural and Religious Traditions**: Often features discussions on maintaining cultural and religious traditions, with particular attention to Christian values.\\
5. **Second Amendment Rights**: Covers gun rights, self-defense, and legal battles surrounding the Second Amendment.\\
\newline
Things They Like:\\
1. **National Pride and Patriotism**: Strong support for events like Independence Day, military recognition, and other symbols of American identity.\\
2. **Entrepreneurship and Small Business**: Celebrates success stories in business, advocating for policies that benefit entrepreneurs and small businesses.\\
3. **Traditional Institutions**: Upholds the value of marriage, family, and community as the cornerstone of a strong society.\\
4. **Pro-Growth Policies**: Enthusiastically supports tax cuts, deregulation, and policies that stimulate economic growth and job creation.\\
5. **Strong Borders and National Security**: Portrays the need for secure borders, immigration reform, and a well-funded military in a positive light.\\
\newline
Things They Dislike:\\
1. **Liberal Social Movements**: Criticizes progressive movements on issues like social justice, gender equality, and racial justice, viewing them as threats to traditional values.\\
2. **Big Government and High Taxes**: Strongly opposes government overreach, excessive taxes, and regulation, particularly when it comes to economic matters.\\
3. **Globalism**: Opposes international agreements or organizations that diminish national sovereignty or place restrictions on the country’s ability to act independently.\\
4. **Political Correctness**: Critical of what it sees as an overemphasis on political correctness in media, culture, and public discourse, arguing it stifles free speech.\\
5. **Anti-Religious Sentiments**: Disapproves of secularism or any movements that seek to diminish the role of religion in public life.
\end{tcolorbox}

\subsection{Sentence Selection Prompt for News Article Generation}
\label{appendix:prompts:news:sentence-selection}
\noindent
\footnotesize
\begin{tcolorbox}[
    colback=gray!15,
    colframe=gray!15,
    coltext=darkgray,
    boxrule=0pt,
    width=\textwidth,
    boxsep=5pt,
    sharp corners,
    breakable
]You are an AI assistant tasked with selecting subsets of sentences from a fictional current event to be integrated into a fictional news story. Your goal is to create versatile options for varied news stories from the same event. Follow these instructions carefully:\\
\newline
1. Review the optional history of past events:\\
<history>\\
\{\{HISTORY\}\}\\
</history>\\
\newline
2. Examine the current event:\\
<current\_event>\\
\{\{CURRENT\_EVENT\}\}\\
</current\_event>\\
\newline
3. Analyze the current event:\\
   - Read and understand the content thoroughly.\\
   - Identify the most important or newsworthy sentences.\\
   - Consider how different combinations of sentences could create coherent and interesting news stories.\\
   - Think about which content aligns well with what you want to communicate in your newspaper, and which content should be omitted.\\
\newline
4. Create \{\{NUM\_SUBSETS\}\} distinct subsets of sentences, following these guidelines:\\
   - Select groups of sentence IDs that, when combined, would create compelling news stories.\\
   - Ensure each subset tells a slightly different aspect or perspective of the event.\\
   - Include enough context and detail in each subset to understand the main points.\\
   - Avoid redundant information within each subset.\\
   - Consider your newspaper's profile when selecting the content to focus on.\\
\newline
5. Output your results in the following format:\\
<results>\\
<subset>[Comma-separated list of sentence IDs]</subset>\\
</results>\\
\newline
Repeat the <subset> element \{\{NUM\_SUBSETS\}\} times, each containing a unique selection of sentence IDs.\\
\newline
6. Remember to create distinct subsets that capture different aspects of the story. Your goal is to provide versatile options for creating varied news stories from the same event.\\
\newline
Before providing your final answer, use a <scratchpad> to think through your selection process and ensure you're creating diverse and meaningful subsets.
\end{tcolorbox}

\subsection{News Article Writing Prompt}
\label{appendix:prompts:news:write}
\noindent
\footnotesize
\begin{tcolorbox}[
    colback=gray!15,
    colframe=gray!15,
    coltext=darkgray,
    boxrule=0pt,
    width=\textwidth,
    boxsep=5pt,
    sharp corners,
    breakable
]You are an AI assistant tasked with writing a fictional news article based on provided information. Your goal is to create a realistic and engaging news piece using the given details while adhering to specific guidelines. Follow these instructions carefully:\\
\newline
1. First, review the background information about the fictional entities involved in the event. This information is provided in XML format for various categories:\\
\newline
<entities>\\
<LOCATIONS>\\
\{\{LOCATIONS\_XML\}\}\\
</LOCATIONS>\\
\newline
<PERSONS>\\
\{\{PERSONS\_XML\}\}\\
</PERSONS>\\
\newline
<ORGANIZATIONS>\\
\{\{ORGANIZATIONS\_XML\}\}\\
</ORGANIZATIONS>\\
\newline
<PRODUCTS>\\
\{\{PRODUCTS\_XML\}\}\\
</PRODUCTS>\\
\newline
<ARTS>\\
\{\{ARTS\_XML\}\}\\
</ARTS>\\
\newline
<EVENTS>\\
\{\{EVENTS\_XML\}\}\\
</EVENTS>\\
\newline
<BUILDINGS>\\
\{\{BUILDINGS\_XML\}\}\\
</BUILDINGS>\\
\newline
<MISCELLANEOUS>\\
\{\{MISCELLANEOUSS\_XML\}\}\\
</MISCELLANEOUS>\\
</entities>\\
\newline
2. Next, carefully read the event information provided with links to the provided entities as background information:\\
\newline
<event\_info>\\
\{\{EVENT\_INFO\}\}\\
</event\_info>\\
\newline
3. Consider the profile of the newspaper you're writing for.\\
\newline
4. Write a news article about this fictional event following these guidelines:\\
   a. Include ALL the information provided in the event\_info section.\\
   b. Ensure your article aligns with the newspaper's profile.\\
   c. Maintain a professional tone that aligns with your newspaper's profile.\\
   d. Organize the information logically, starting with the most important details.\\
   e. Create a compelling headline that captures the essence of the story and fits your newspaper's style.\\
   f. Write at least two paragraphs, but no more than four.\\
   g. Only use information from the provided entities to maintain consistency with the known fictional world.\\
\newline
6. Present your news article in the following format:\\
\newline
<result>\\
<scratchpad>\\
(Plan your approach here)\\
</scratchpad>\\
<headline>\\
(Write a headline here)\\
</headline>\\
<article>\\
<paragraph>\\
<text>[First paragraph text]</text>\\
</paragraph>\\
<paragraph>\\
<text>[Second paragraph text]</text>\\
</paragraph>\\
<paragraph>\\
<text>[Third paragraph text (if needed)]</text>\\
</paragraph>\\
</article>\\
</result>\\
\newline
7. After writing the article, double-check that you've included all the information from the event\_info section. \\
\newline
Remember, your goal is to create a realistic and engaging news article based on the provided fictional event information while adhering to the newspaper's profile. Good luck!
\end{tcolorbox}

\subsection{Hallucination Removal Prompt in News Article Generation}
\label{appendix:prompts:news:hallucination}
\noindent
\footnotesize
\begin{tcolorbox}[
    colback=gray!15,
    colframe=gray!15,
    coltext=darkgray,
    boxrule=0pt,
    width=\textwidth,
    boxsep=5pt,
    sharp corners,
    breakable
]You are tasked with improving a news article about a fictional event. Your goal is to ensure the article is faithful to the provided ground truth information while maintaining the general style of the original news article. Follow these instructions carefully:\\
\newline
1. First, review the ground truth information about the fictional named entities:\\
\newline
<entities>\\
<LOCATIONS>\\
\{\{LOCATIONS\_XML\}\}\\
</LOCATIONS>\\
\newline
<PERSONS>\\
\{\{PERSONS\_XML\}\}\\
</PERSONS>\\
\newline
<ORGANIZATIONS>\\
\{\{ORGANIZATIONS\_XML\}\}\\
</ORGANIZATIONS>\\
\newline
<PRODUCTS>\\
\{\{PRODUCTS\_XML\}\}\\
</PRODUCTS>\\
\newline
<ARTS>\\
\{\{ARTS\_XML\}\}\\
</ARTS>\\
\newline
<EVENTS>\\
\{\{EVENTS\_XML\}\}\\
</EVENTS>\\
\newline
<BUILDINGS>\\
\{\{BUILDINGS\_XML\}\}\\
</BUILDINGS>\\
\newline
<MISCELLANEOUS>\\
\{\{MISCELLANEOUSS\_XML\}\}\\
</MISCELLANEOUS>\\
</entities>\\
\newline
2. Next, review the ground truth information about the fictional event:\\
\newline
<event\_info>\\
\{\{EVENT\_INFO\}\}\\
</event\_info>\\
\newline
3. Now, read the generated news article that needs to be revised:\\
\newline
<news-article>\\
\{\{CURRENT\_NEWS\_ARTICLE\_XML\}\}\\
</news-article>\\
\newline
4. To revise the news article, follow these steps. Only revise the content of the article paragraphs. Do not revise the article title:\\
\newline
a) Analyze the style of the original news article. Pay attention to tone, vocabulary, and sentence structure. Any changes you make should maintain this style.\\
\newline
b) Carefully examine all factual statements in the article. Determine if each statement can be verified based on the ground truth information provided.\\
\newline
c) Identify any factual statements that are:\\
   - Expressed as facts but are unverifiable based on the ground truth\\
   - Incorrect according to the ground truth information\\
\newline
d) For each problematic statement, choose one of the following actions:\\
   - Remove the unverifiable or incorrect factual statement entirely.\\
   - Rephrase the statement to clearly indicate that it is not a verified fact (e.g., by using hedging language or attributing the information to an unnamed source).\\
\newline
e) Ensure that no information beyond what is provided in the ground truth is introduced as fact. However, you may include unverified information if it is presented as speculation, question, rumor or involves appropriate hedging.\\
\newline
f) Make any necessary adjustments to improve the flow and coherence of the article after your revisions.\\
\newline
g) Make minimal edits to the news article. Only make necessary revisions to avoid problematic factual statements. Explain and justify each change you make.\\
\newline
5. Output the revised news article in the following format:\\
\newline
<result>\\
<scratchpad>\\
(Plan your approach here, outlining the main changes you intend to make)\\
</scratchpad>\\
<headline>\\
(Write a revised headline that accurately reflects the content of the article)\\
</headline>\\
<article>\\
<paragraph>\\
<text>[First paragraph text]</text>\\
</paragraph>\\
<paragraph>\\
<text>[Second paragraph text]</text>\\
</paragraph>\\
<paragraph>\\
<text>[Third paragraph text (if needed)]</text>\\
</paragraph>\\
(Add more paragraphs as necessary, following the same format)\\
</article>\\
</result>\\
\newline
Remember to maintain the original style of the article while ensuring all factual statements are accurate according to the provided ground truth information.
\end{tcolorbox}

\subsection{Missing Information Adding Prompt in News Article Generation}
\label{appendix:prompts:news:add}
\noindent
\footnotesize
\begin{tcolorbox}[
    colback=gray!15,
    colframe=gray!15,
    coltext=darkgray,
    boxrule=0pt,
    width=\textwidth,
    boxsep=5pt,
    sharp corners,
    breakable
]You are an AI assistant tasked with improving a news article about a fictional event. Your goal is to ensure the article contains all specific details and information from a provided ground-truth outline of the fictional event. Follow these instructions carefully:\\
\newline
1. First, review the list of fictional named entities as background information:\\
\newline
<entities>\\
<LOCATIONS>\\
\{\{LOCATIONS\_XML\}\}\\
</LOCATIONS>\\
\newline
<PERSONS>\\
\{\{PERSONS\_XML\}\}\\
</PERSONS>\\
\newline
<ORGANIZATIONS>\\
\{\{ORGANIZATIONS\_XML\}\}\\
</ORGANIZATIONS>\\
\newline
<PRODUCTS>\\
\{\{PRODUCTS\_XML\}\}\\
</PRODUCTS>\\
\newline
<ARTS>\\
\{\{ARTS\_XML\}\}\\
</ARTS>\\
\newline
<EVENTS>\\
\{\{EVENTS\_XML\}\}\\
</EVENTS>\\
\newline
<BUILDINGS>\\
\{\{BUILDINGS\_XML\}\}\\
</BUILDINGS>\\
\newline
<MISCELLANEOUS>\\
\{\{MISCELLANEOUSS\_XML\}\}\\
</MISCELLANEOUS>\\
</entities>\\
\newline
2. Next, review the ground truth outline of the fictional event, including all details that must be communicated in the news article:\\
\newline
<event\_info>\\
\{\{EVENT\_INFO\}\}\\
</event\_info>\\
\newline
3. Now, read the generated news article that needs to be revised:\\
\newline
<news-article>\\
\{\{CURRENT\_NEWS\_ARTICLE\_XML\}\}\\
</news-article>\\
\newline
4. To revise the news article, follow these steps. Only revise the content of the article paragraphs. Do not revise the article title:\\
a) Analyze the style of the original news article. Pay attention to tone, vocabulary, and sentence structure. Any changes you make should maintain this style.\\
b) Go over each individual sentence from the ground truth outline.\\
c) Each sentence contains many details. Make sure that each of the details is communicated within the news article.\\
   - The details do not need to be communicated verbatim. It is okay if the same content is communicated in different terms.\\
   - Focus on all details from the sentence of the ground truth outline (numbers, dates, relations, relevant attributes and adjectives, etc). Consider every specific detail you can find.\\
d) Make subtle adjustments to the news article for each detail that is not yet communicated:\\
   - Add the information with minimal edits\\
   - Do not revise additional information from the news article such as speculations, rumors etc. Focus only on the missing information that must be integrated in the article.\\
e) Make any necessary adjustments to improve the flow and coherence of the article after your revisions.\\
\newline
5. Output the revised news article in the following format:\\
\newline
<result>\\
<scratchpad>\\
(Plan your approach here, outlining the main changes you intend to make)\\
</scratchpad>\\
<headline>\\
(Write a revised headline that accurately reflects the content of the article)\\
</headline>\\
<article>\\
<paragraph>\\
<text>[First paragraph text]</text>\\
</paragraph>\\
<paragraph>\\
<text>[Second paragraph text]</text>\\
</paragraph>\\
<paragraph>\\
<text>[Third paragraph text (if needed)]</text>\\
</paragraph>\\
(Add more paragraphs as necessary, following the same format)\\
</article>\\
</result>\\
\newline
Remember to maintain the original style of the article while ensuring all factual statements are accurate according to the provided ground truth information.
\end{tcolorbox}

\subsection{Named Entity Resolution Prompt in News Article Generation}
\label{appendix:prompts:news:ner}
\noindent
\footnotesize
\begin{tcolorbox}[
    colback=gray!15,
    colframe=gray!15,
    coltext=darkgray,
    boxrule=0pt,
    width=\textwidth,
    boxsep=5pt,
    sharp corners,
    breakable
]You are an AI assistant tasked with processing news passages by identifying and marking named entities. Follow these instructions carefully:\\
\newline
First, review the list of fictional named entities provided below:\\
\newline
<entities>\\
<LOCATIONS>\\
\{\{USED\_LOCATIONS\_XML\}\}\\
</LOCATIONS>\\
\newline
<PERSONS>\\
\{\{USED\_PERSONS\_XML\}\}\\
</PERSONS>\\
\newline
<ORGANIZATIONS>\\
\{\{USED\_ORGANIZATIONS\_XML\}\}\\
</ORGANIZATIONS>\\
\newline
<PRODUCTS>\\
\{\{USED\_PRODUCTS\_XML\}\}\\
</PRODUCTS>\\
\newline
<ARTS>\\
\{\{USED\_ARTS\_XML\}\}\\
</ARTS>\\
\newline
<EVENTS>\\
\{\{USED\_EVENTS\_XML\}\}\\
</EVENTS>\\
\newline
<BUILDINGS>\\
\{\{USED\_BUILDINGS\_XML\}\}\\
</BUILDINGS>\\
\newline
<MISCELLANEOUS>\\
\{\{USED\_MISCELLANEOUSS\_XML\}\}\\
</MISCELLANEOUS>\\
</entities>\\
\newline
Next, you will process the following news article passages:\\
\newline
<passages>\\
\{\{PASSAGES\_XML\}\}\\
</passages>\\
\newline
Your task is to process these passages by following these steps:\\
\newline
1. Carefully review the list of entities provided in the <entities> section. Each entity will have an associated ID.\\
\newline
2. Search the passages for all occurrences of each entity in the list.\\
\newline
3. For each occurrence found, replace it with the format: \{phrase|ID\}\\
   Where "phrase" is exactly how the entity appears in the text (maintaining any abbreviations or variations), and "ID" is the entity's identifier from the entities list.\\
\newline
4. Maintain the original structure and formatting of the passages, only changing the entities as described.\\
\newline
5. After processing all entities, review the entire passage to ensure all occurrences have been properly marked and no entities were missed.\\
\newline
6. Output the processed passages, maintaining its original structure but with all entity occurrences replaced as instructed.\\
\newline
Important points to remember:\\
- Be thorough in your search for entities, including variations or partial mentions.\\
- Preserve the original text exactly as it appears, only adding the entity markup.\\
- Keep the COMPLETE ORIGINAL phrase that you are replacing with \{phrase|ID\}. The sentence should be identical to how it was before, except for the added markup.\\
- If an entity is referred to by full name, the "phrase" is the full name.\\
- If an entity is referred to by an abbreviation, the "phrase" is the used abbreviation.\\
- If an entity is referred to using parts of the full name, then the "phrase" would be the same parts of the full name.\\
\newline
Examples:\\
1. "Renowned novelist Elara Vance and celebrated philanthropist Rohan Kapoor exchanged vows." \\
   Should be replaced with:\\
   "Renowned novelist \{Elara Vance|PERSON-1\} and celebrated philanthropist \{Rohan Kapoor|PERSON-2\} exchanged vows."\\
   (When Elara Vance has ID PERSON-1 and Rohan Kapoor has ID PERSON-2)\\
\newline
2. "Renowned novelist Elara and celebrated philanthropist R. Kapoor exchanged vows." \\
   Should be replaced with:\\
   "Renowned novelist \{Elara|PERSON-1\} and celebrated philanthropist \{R. Kapoor|PERSON-2\} exchanged vows."\\
\newline
3. "Anna Peters told Tim that he should stop talking."\\
   should be written as:\\
   "\{Anna Peters|PERSON-3\} told \{Tim|PERSON-4\} that \{he|PERSON-4\} should stop talking."\\
   (When Anna Peters has ID PERSON-3 and Tim Laurens has ID PERSON-4 and is referred to here)\\
\newline
If multiple entities are referred to by the same word, use this format:\\
"\{Both|PERSON-3,PERSON-4\} liked the chocolate."\\
\newline
Format your output as follows:\\
- Enclose the entire processed news within <news> tags.\\
- Place each passage of the outline within separate <passage> tags.\\
\newline
Provide your final output without any additional commentary or explanations. Focus solely on processing the outline as instructed.\\
\newline
Provide all output in an overall <results> root node.
\end{tcolorbox}

\subsection{Answerability Filtering Prompt for Multi-hop Questions}
\label{appendix:prompts:quality:multi-hop}
\noindent
\footnotesize
\begin{tcolorbox}[
    colback=gray!15,
    colframe=gray!15,
    coltext=darkgray,
    boxrule=0pt,
    width=\textwidth,
    boxsep=5pt,
    sharp corners,
    breakable
]You will receive evidence documents, a question, a date on which the question is asked, and answer options. Your task is to evaluate the evidence information, determine if it provides enough details to answer the question based on the date, and choose the correct answer. \\
\newline
**Evidence:** \\
\{\{EVENTS\}\} \\
\newline
**Date of the question:** \\
\{\{DATE\}\} \\
\newline
**Question:** \\
\{\{QUESTION\}\} \\
\newline
**Answer Options:** \\
\{\{ANSWERS\}\} \\
\newline
**Instructions:** \\
\newline
1. **Analyze the Evidence:** \\
   - Carefully read all the provided evidence. \\
   - Compare the information in the evidence with the question. \\
   - Check if the combined evidence confirms all the necessary details to answer the question. \\
\newline
2. **Select an Answer:** \\
   - Choose the correct answer if all necessary details are provided. \\
   - If the evidence lacks information or any important detail is missing, select the option for "Unknown." \\
\newline
3. **Submit Your Answer:** \\
   - Provide your response in JSON format as follows: \\
   \{ \\
       "justification": "Brief explanation (1-2 sentences)", \\
       "answer\_choice": "A single number corresponding to the chosen answer." \\
   \} \\
\newline
Output the JSON directly without any additional text or markdown syntax.

\end{tcolorbox}

\subsection{Answerability Filtering Prompt for Time-span Questions}
\label{appendix:prompts:quality:time-span}
\noindent
\footnotesize
\begin{tcolorbox}[
    colback=gray!15,
    colframe=gray!15,
    coltext=darkgray,
    boxrule=0pt,
    width=\textwidth,
    boxsep=5pt,
    sharp corners,
    breakable
]You are tasked with evaluating evidence text to answer a question based on the provided information and a given date. Your goal is to determine if there's enough information to answer the question and select the correct answer option. \\
\newline
Here is the evidence you need to analyze: \\
\{\{EVENTS\}\} \\
\newline \\
The date on which the question is asked: \\
\{\{DATE\}\} \\
\newline
The question you need to answer: \\
\{\{QUESTION\}\} \\
\newline
The available answer options: \\
\{\{ANSWERS\}\} \\
\newline
Follow these steps to complete the task: \\
\newline
1. Carefully read and analyze all the provided evidence. \\
2. Compare the information in the evidence with the question. \\
3. Check if the combined information from the evidence confirms all the necessary details to answer the question. \\
\newline
4. Use a scratchpad to derive your answer: \\
   <scratchpad> \\
   - Identify all absolute start and end dates based on the evidence and any relevant assumptions from the question. \\
   - Derive the answer based on these dates, if possible. \\
   - Show your step-by-step reasoning process. \\
   </scratchpad> \\
\newline
5. Select an answer: \\
   - Choose the correct answer if all necessary details are provided. \\
   - If the evidence lacks information or if any important detail is missing, select the option for "Unknown." \\
   - If the question includes assumptions, treat them as facts. \\
\newline
6. Provide your response in JSON format as follows: \\
   \{ \\
       "scratchpad": "Your reasoning from the scratchpad", \\
       "justification": "Brief explanation (1-2 sentences)", \\
       "answer\_choice": "A single number corresponding to the chosen answer." \\
   \}
\newline
Output the JSON directly without any additional text or markdown syntax. 

\end{tcolorbox}

\subsection{Assumption Leakage Filtering Prompt for Time-span Questions}
\label{appendix:prompts:quality:time-span-assumptions}
\noindent
\footnotesize
\begin{tcolorbox}[
    colback=gray!15,
    colframe=gray!15,
    coltext=darkgray,
    boxrule=0pt,
    width=\textwidth,
    boxsep=5pt,
    sharp corners,
    breakable
]Given a question asked on a specific date, determine if there is sufficient evidence to answer the question. \\
If evidence is provided, base your answer on it. If no evidence is given, assess whether the question itself contains enough context to draw a conclusion. \\
\newline
Only provide an answer if the available evidence (or the question itself) is sufficient to determine it with certainty. \\
If the evidence is insufficient or unclear, select the answer option "Unknown" rather than making a guess. \\
\newline
Provide your response in JSON format as follows: \\
\{
"scratchpad": "Your reasoning from the scratchpad", \\
"justification": "Brief explanation (1-2 sentences)", \\
"answer\_choice": "A single number corresponding to the chosen answer." \\
\}
\newline
Question Date: \{\{QUESTION\_DATE\}\} \\
Question: \{\{QUESTION\}\} \\
\newline
Evidence: \\
\{\{EVIDENCE\_TEXT\}\} \\
\newline
Answer Options: \\
\{\{ANSWER\_OPTIONS\}\}
\end{tcolorbox}
\newpage

\subsection{Multiple Choice Prompt for Experiments on \dataset{} (Prompt 1)}
\label{appendix:prompts:experiments:prompt1}
\noindent
\footnotesize
\begin{tcolorbox}[
    colback=gray!15,
    colframe=gray!15,
    coltext=darkgray,
    boxrule=0pt,
    width=\textwidth,
    boxsep=5pt,
    sharp corners,
    breakable
]Given the following news articles, the question, and the answer options, answer the question.\\
If the question cannot be answered with certainty based on the news articles, select the answer option "Unanswerable".\\
\newline
News Articles:\\
\{\{NEWS\_ARTICLES\}\}\\
\newline
Question: \{\{QUESTION\}\}\\
\newline
Date of Question: \{\{DATE\}\}\\
\newline
Answer options:\\
\{\{ANSWERS\}\}\\
\newline
Select the answer option that correctly answers the question. If the question cannot be answered with certainty based on the news articles, choose "Unanswerable" (if it is one of the options). In the final line of your response, provide the number of the correct answer option using the format: "Answer: [answer number]" (for example, "Answer: X").

\end{tcolorbox}

\subsection{Multiple Choice Prompt for Experiments on \dataset{} (Prompt 2)}
\label{appendix:prompts:experiments:prompt2}
\noindent
\footnotesize
\begin{tcolorbox}[
    colback=gray!15,
    colframe=gray!15,
    coltext=darkgray,
    boxrule=0pt,
    width=\textwidth,
    boxsep=5pt,
    sharp corners,
    breakable
]You will receive news articles, a question, a date on which the question is asked, and answer options.\\
Your task is to evaluate the articles, determine if they provide enough information to answer the question based on the date, and choose the correct answer.\\
\newline
News Articles:\\
\{\{NEWS\_ARTICLES\}\}\\
\newline
Question: \{\{QUESTION\}\}\\
\newline
Date of Question: \{\{DATE\}\}\\
\newline
Answer options:\\
\{\{ANSWERS\}\}\\
\newline
**Instructions:**\\
1. **Analyze the news articles:**\\
- Carefully read all the news articles.\\
- Compare the information in the articles with the question.\\
- Check if the combined information from the articles confirms all the details required to answer the question.\\
\newline
2. **Select an Answer:**\\
- Choose the correct answer if all necessary details are provided.\\
- If the articles lack information or any important detail is missing, select the option for "Unanswerable".\\
\newline
3. **Submit your Answer**\\
- Select the answer option that correctly answers the question. If the question cannot be answered with certainty based on the news articles, choose "Unanswerable" (if it is one of the options). In the final line of your response, provide the number of the correct answer option using the format: "Answer: [answer number]" (for example, "Answer: X").
\end{tcolorbox}

\subsection{Multiple Choice Prompt for Experiments on \dataset{} (Prompt 3)}
\label{appendix:prompts:experiments:prompt3}
\noindent
\footnotesize
\begin{tcolorbox}[
    colback=gray!15,
    colframe=gray!15,
    coltext=darkgray,
    boxrule=0pt,
    width=\textwidth,
    boxsep=5pt,
    sharp corners,
    breakable
]You will receive news articles, a question, a date on which the question is asked, and answer options.\\
Your task is to evaluate the articles, determine if they provide enough information to answer the question based on the date, and choose the correct answer.\\
\newline
News Articles:\\
\{\{NEWS\_ARTICLES\}\}\\
\newline
Question: \{\{QUESTION\}\}\\
\newline
Date of Question: \{\{DATE\}\}\\
\newline
Answer options:\\
\{\{ANSWERS\}\}\\
\newline
**Instructions:**\\
1. **Analyze the news articles:**\\
   - Carefully read all the news articles.\\
   - Compare the information in the articles with the question.\\
   - Check if the combined information from the articles confirms all the details required to answer the question.\\
   - Ensure that the question does not contain contradictory information compared to the provided news articles. Select the "Unanswerable" option if it does.\\
   - Verify that the information in the news articles is sufficient to answer the question with certainty. If you cannot answer the question with certainty based on the evidence, select the "Unanswerable" option.\\
   - The news articles may not be in the correct temporal order.\\
   - If the question mentions an "event date", this refers to the date of the news article.\\
   - Unless otherwise stated, you can assume that each news article reports events that occurred on the date of the article.\\
\newline
2. **Select an Answer:**\\
   - Choose the correct answer if all necessary details are provided.\\
   - If the articles lack information or any important detail is missing, select the option for "Unanswerable".\\
\newline
3. **Submit your Answer:**\\
   - Select the answer option that correctly answers the question. If the question cannot be answered with certainty based on the news articles, choose "Unanswerable" (if it is one of the options). In the final line of your response, provide the number of the correct answer option using the format: "Answer: [answer number]" (for example, "Answer: X").
\end{tcolorbox}

\subsection{Multiple Choice Prompt for Experiments on \dataset{} (Prompt 4)}
\label{appendix:prompts:experiments:prompt4}
\noindent
\footnotesize
\begin{tcolorbox}[
    colback=gray!15,
    colframe=gray!15,
    coltext=darkgray,
    boxrule=0pt,
    width=\textwidth,
    boxsep=5pt,
    sharp corners,
    breakable
]You will receive news articles, a question, a date on which the question is asked, and answer options.\\
Your task is to evaluate the articles, determine if they provide enough information to answer the question based on the date, and choose the correct answer.\\
\newline
News Articles:\\
\{\{NEWS\_ARTICLES\}\}\\
\newline
Question: \{\{QUESTION\}\}\\
\newline
Date of Question: \{\{DATE\}\}\\
\newline
Answer options:\\
\{\{ANSWERS\}\}\\
\newline
**Instructions:**\\
1. **Analyze the news articles:**\\
   - Carefully read all the news articles.\\
   - Compare the information in the articles with the question.\\
   - Check if the combined information from the articles confirms all the details required to answer the question.\\
   - Ensure that the question does not contain contradictory information compared to the provided news articles. Select the "Unanswerable" option if it does.\\
   - Verify that the information in the news articles is sufficient to answer the question with certainty. If you cannot answer the question with certainty based on the evidence, select the "Unanswerable" option.\\
   - The news articles may not be in the correct temporal order.\\
   - If the question mentions an "event date", this refers to the date of the news article.\\
   - Unless otherwise stated, you can assume that each news article reports events that occurred on the date of the article.\\
\newline
2. **Double-check the details:**\\
   - Use only the information provided in the news articles.\\
   - Avoid assumptions beyond what is explicitly stated.\\
   - Do not make guesses. Only provide an answer if the information in the article is enough to answer the question with certainty. If it’s not, select the "Unanswerable" option.\\
\newline
3. **Select an Answer:**\\
   - Choose the correct answer if all necessary details are provided.\\
   - If the articles lack information or any important detail is missing, select the option for "Unanswerable".\\
\newline
4. **Submit your Answer:**\\
   - Select the answer option that correctly answers the question. If the question cannot be answered with certainty based on the news articles, choose "Unanswerable" (if it is one of the options). In the final line of your response, provide the number of the correct answer option using the format: "Answer: [answer number]" (for example, "Answer: X").
\end{tcolorbox}

\subsection{Multiple Choice Prompt for Experiments on \dataset{} (Prompt 5)}
\label{appendix:prompts:experiments:prompt5}
\noindent
\footnotesize
\begin{tcolorbox}[
    colback=gray!15,
    colframe=gray!15,
    coltext=darkgray,
    boxrule=0pt,
    width=\textwidth,
    boxsep=5pt,
    sharp corners,
    breakable
]You will receive news articles, a question, a date on which the question is asked, and answer options.\\
Your task is to evaluate the articles, determine if they provide enough information to answer the question based on the date, and choose the correct answer.\\
\newline
News Articles:\\
\{\{NEWS\_ARTICLES\}\}\\
\newline
Question: \{\{QUESTION\}\}\\
\newline
Date of Question: \{\{DATE\}\}\\
\newline
Answer options:\\
\{\{ANSWERS\}\}\\
\newline
**Instructions:**\\
1. **Analyze the news articles:**\\
   - Carefully read all the news articles.\\
   - Compare the information in the articles with the question.\\
   - Check if the combined information from the articles confirms all the details required to answer the question.\\
   - Ensure that the question does not contain contradictory information compared to the provided news articles. Select the "Unanswerable" option if it does.\\
   - Verify that the information in the news articles is sufficient to answer the question with certainty. If you cannot answer the question with certainty based on the evidence, select the "Unanswerable" option.\\
   - The news articles may not be in the correct temporal order.\\
   - If the question mentions an "event date", this refers to the date of the news article.\\
   - Unless otherwise stated, you can assume that each news article reports events that occurred on the date of the article.\\
\newline
2. **Double-check the details:**\\
   - Use only the information provided in the news articles.\\
   - Avoid assumptions beyond what is explicitly stated.\\
   - Do not make guesses. Only provide an answer if the information in the article is enough to answer the question with certainty. If it’s not, select the "Unanswerable" option.\\
   - Make sure that all the necessary information from the question is present in the news article. For each detail in the question, write down how you verified it against the articles, along with your conclusion. If any important details are missing and it’s unclear whether the article fully supports the question, select the "Unanswerable" option.   - Use only the information provided in the news articles.\\
\newline
3. **Select an Answer:**\\
   - Choose the correct answer if all necessary details are provided.\\
   - If the articles lack information or any important detail is missing, select the option for "Unanswerable".\\
\newline
4. **Submit your Answer:**\\
   - Select the answer option that correctly answers the question. If the question cannot be answered with certainty based on the news articles, choose "Unanswerable" (if it is one of the options). In the final line of your response, provide the number of the correct answer option using the format: "Answer: [answer number]" (for example, "Answer: X").
\end{tcolorbox}